\documentclass{ieeeaccess}
\usepackage{cite}
\usepackage{amsmath,amssymb,amsfonts}
\usepackage{algorithmic}
\usepackage{graphicx}
\usepackage{textcomp}
\usepackage{cite}
\usepackage{amsmath,amssymb,amsfonts}
\usepackage{algorithmic}
\usepackage{graphicx}
\usepackage{textcomp}
\usepackage{xcolor}
\usepackage{color}
\usepackage{booktabs}
\usepackage{multirow}
\usepackage{colortbl}
\usepackage{graphicx}
\usepackage[ruled]{algorithm2e}
\usepackage{algorithmic}
\usepackage{graphicx}
\usepackage{mathtools}
\usepackage{breqn}
\usepackage{url}
\usepackage{textcomp}
\usepackage{xcolor}
\usepackage{mathrsfs}
\usepackage{longtable}
\usepackage{booktabs} 
\usepackage{bigstrut}
\usepackage{multirow}
\usepackage{color}
\usepackage{fancyhdr}
\usepackage{listings} 
\usepackage{changepage}
\usepackage{lipsum}
\usepackage{floatrow}
\usepackage{balance}
\usepackage{enumitem}
\usepackage{comment}
\usepackage{breqn}
\usepackage{amssymb}
\usepackage{multirow}
\usepackage{longtable}
\usepackage{hyperref}
\usepackage{amssymb}
\usepackage{bbding}
\usepackage{pifont}
\usepackage{wasysym}
\usepackage{amssymb}
\usepackage{caption}
\usepackage{subcaption}
\newcommand{\cmark}{\ding{51}}%
\newcommand{\xmark}{\ding{55}}%
\SetKwInput{KwInput}{Input}
\SetKwInput{KwOutput}{Output}

\def\BibTeX{{\rm B\kern-.05em{\sc i\kern-.025em b}\kern-.08em
    T\kern-.1667em\lower.7ex\hbox{E}\kern-.125emX}}
\begin{document}
\history{Date of publication xxxx 00, 0000, date of current version xxxx 00, 0000.}
\doi{10.1109/ACCESS.2017.DOI}

\title{Deep Composite Face Image Attacks: Generation, Vulnerability and Detection}
\author{Jag Mohan Singh,~\IEEEmembership{Member,~IEEE,}
  and    Raghavendra Ramachandra,~\IEEEmembership{Senior Member,~IEEE}}
\address[]{Norwegian University of Science and Technology (NTNU), Norway\\ (e-mail: {jag.m.singh; raghavendra.ramachandra}@ntnu.no)}
\tfootnote{}

\markboth
{Singh \headeretal: Deep Composite Face Image Attacks: Generation, Vulnerability and Detection}
{Singh \headeretal: Deep Composite Face Image Attacks: Generation, Vulnerability and Detection}

\corresp{Corresponding author: Jag M. Singh (e-mail: jag.m.singh@ntnu.no).}

\begin{abstract}
Face manipulation attacks have drawn the attention of biometric researchers because of their vulnerability to Face Recognition Systems (FRS). This paper proposes a novel scheme to generate Composite Face Image Attacks (CFIA) based on facial attributes using Generative Adversarial Networks (GANs). Given the face images corresponding to two unique data subjects, the proposed CFIA method will independently generate the segmented facial attributes, then blend them using transparent masks to generate the CFIA samples. We generate $526$ unique CFIA combinations of facial attributes for each pair of contributory data subjects. Extensive experiments are carried out on our newly generated CFIA dataset consisting of 1000 unique identities with 2000 bona fide samples and 526000 CFIA samples, thus resulting in an overall 528000 face image samples. {{We present a sequence of experiments to benchmark the attack potential of CFIA samples using four different  automatic FRS}}. We introduced a new metric named Generalized Morphing Attack Potential (G-MAP) to benchmark the vulnerability of generated attacks on FRS effectively. Additional experiments are performed on the representative subset of the CFIA dataset to benchmark both perceptual quality and human observer response. Finally, the CFIA detection performance is benchmarked using three different single image based face Morphing Attack Detection (MAD) algorithms.   The source code of the proposed method together with CFIA dataset will be made publicly available: \url{https://github.com/jagmohaniiit/LatentCompositionCode}
\end{abstract}


\titlepgskip=-15pt

\maketitle
\section{Introduction}
\label{sec:introduction}

 \PARstart{F}{RS} demonstrates highly accurate verification rates, which has led to their widespread usage in eCommerce, online banking, surveillance and security applications. The recent advances in deep learning techniques have further increased the accuracy of the FRS ~\cite{Schroff15-Facenet-CVPR}, ~\cite{Deng19-Arcface-CVPR}  that enabled them to be deployed in the border control applications. However, the FRS is vulnerable to various attacks, among which the face morphing attacks have gained attention due to their impact on the border control applications. Recent benchmarking results reported in NIST FRVT MOPRH~\cite{NISTFRVTMorph} indicate that the higher the accuracy of the FRS, the higher the vulnerability for the morphing attacks. 

 One of the most widely used attacks toward FRS is the Presentation Attacks (PA), a.k.a spoofing attacks, which can be achieved by presenting a biometric artefact to the biometric capture device. PA can be performed by generating a Presentation Attack Instrument (PAI) that includes either a  printed photo (print-photo), displaying an image (display-photo), displaying a video (replay-video), or the use of a rigid/non-rigid 3D face mask (mask-attack). Biometric researchers had thus devised Presentation Attack Detection (PAD) as a countermeasure to PA that is extensively discussed in ~\cite{Ramachandra17_PADSurvey_CSUR}, and ~\cite{Abdullakutty21_PADSurvey_IF}.

The second type of widely studied attack on the FRS is the adversarial attack, which can be performed by applying a small perturbation (noise), a.k.a adversarial perturbation, to a facial image. Even though the introduced perturbation is indistinguishable to the human eye but can lead to mis-classification with high-confidence~\cite{Goodfellow14_FGSM_ICLR} and can be used to expose vulnerabilities of the FRS. Adversarial attacks have shown { high vulnerability in FRS}, especially on the deep learning-based FRS \cite{Xu22-Adversarial-ChapterSpringer}. {{The white box adversarial attack requires complete knowledge of the underlying deep learning model.}}.
Adversarial attacks could also be black-box attack performed during testing, and the attacker does not know the underlying deep-learning model. Several countermeasures to address the adversarial attacks are extensively discussed in ~\cite{Vakhshiteh21_AdversarialAttacksSurvey_IEEEAccess}, \cite{Naveed18_AdversarialAttacksSurvey_IEEEAccess}. It needs to be pointed out that adversarial attacks are digital when performed on images, but they can also be performed in the physical world by using a unique eyeglass for impersonation ~\cite{Vakhshiteh21_AdversarialAttacksSurvey_IEEEAccess}.

{{
Face morphing attacks are gaining high momentum in the biometric community. The face morphing process seamlessly combines face images from two or more subjects (also called contributory subjects) to generate a morphing image. The generated morphing image shows substantial visual similarity to both contributory subjects therefore challenging to detect by the experts (border guards and police) \cite{Venkatesh21-FaceMorphingSurvey-TTS, Ferrara14MagicPassportBTAS, Godage22IEEETTS,FaceMorphingBiometricUpdate}. Notably, the morphed images will get verified to both the contributory subjects when used with automatic FRS \cite{Venkatesh21-FaceMorphingSurvey-TTS}. Therefore, the morphing attacks can be instrumented to acquire the ID documents like passports, driving licenses, bank accounts, etc. For example, a subject with criminal background can obtain a passport by collaborating with an accomplice to generate a morphing image. Then, the accomplice can apply for an ID document using the morphed image. The subject with a criminal background can use the obtained ID document to cross the border \cite{Venkatesh21-FaceMorphingSurvey-TTS, Ferrara14MagicPassportBTAS}}}.

 Face morphing can be generated using algorithms based on facial landmarks  such as Face Morpher~\cite{Alyssa18-FaceMorpher} and UBO-Morph~\cite{Ferrara17-FaceDemorphing-TIFS}. More recently, algorithms based on Generative Adversarial Networks (GANs) such as MorGAN~\cite{Damer18-MorGAN-BTAS},  MIPGAN~\cite{Zhang21-MIPGAN-TBIOM} and ReGenMorph \cite{damer2021regenmorph} have also been used to generate face morphing images. These generated face morphing images have demonstrated the high vulnerability of FRS, especially in the passport application scenario, including automatic border control. {{Further, morphing attacks can deceive both human observers (border control officers) and automatic FRS in Automatic Border Control (ABC)}} \cite{Godage22IEEETTS, Venkatesh21-FaceMorphingSurvey-TTS}, \cite{Rancha22_HumanObserver_Arxiv}. Following the initial paper~\cite{Ferrara14-MagicPass-IJCB}, there have been several papers on morphing detection, and the reader is advised to refer to the survey by Venkatesh et al.~\cite{Venkatesh21-FaceMorphingSurvey-TTS} to get a detailed overview on face morphing.

Most face morphing generation works are devised by performing the blending operation on the complete (or total) face images \cite{Venkatesh21-FaceMorphingSurvey-TTS}. However, the success rate of the full-face morphing attack is high when contributory subjects are lookalikes to deceive the super-recognizer and highly trained border guards \cite{Godage22IEEETTS}. Therefore, partial face morphing was introduced in \cite{Qin21-MorphParts-TBIOM} where the blending operation is carried out using Poisson image editing~\cite{Patrick03-Poisson-SIGGRAPH}. The generated composite morphs have shown vulnerabilities of FRS based on deep-learning features such as VGGFace~\cite{Parkhi15-VGG-BMVC},  Arcface~\cite{Deng19-Arcface-CVPR} and commercial-off-the-shelf (COTS) that includes Neurotech~\cite{NeurotechVerilook} and Cognitec~\cite{CognitecFaceVACS}. Further, the human observer analysis is also discussed. However, the work presented in \cite{Qin21-MorphParts-TBIOM} has several limitations, including (1) it is based on landmarks, and this would lead to pixel-based artifacts due to alignment issues, and correction of these would require manual intervention ~\cite{Venkatesh21-FaceMorphingSurvey-TTS} (2) Only a few arbitrary regions are used to generate the composite images (3) Limited only to the base regions like nose, mouth, eye and forehead. (4) Limited only to the single facial attribute composite generation (5) failure to achieve a high vulnerability of FRS.

\begin{figure*}[htb]
\begin{center}
\includegraphics[width=1.0\linewidth]{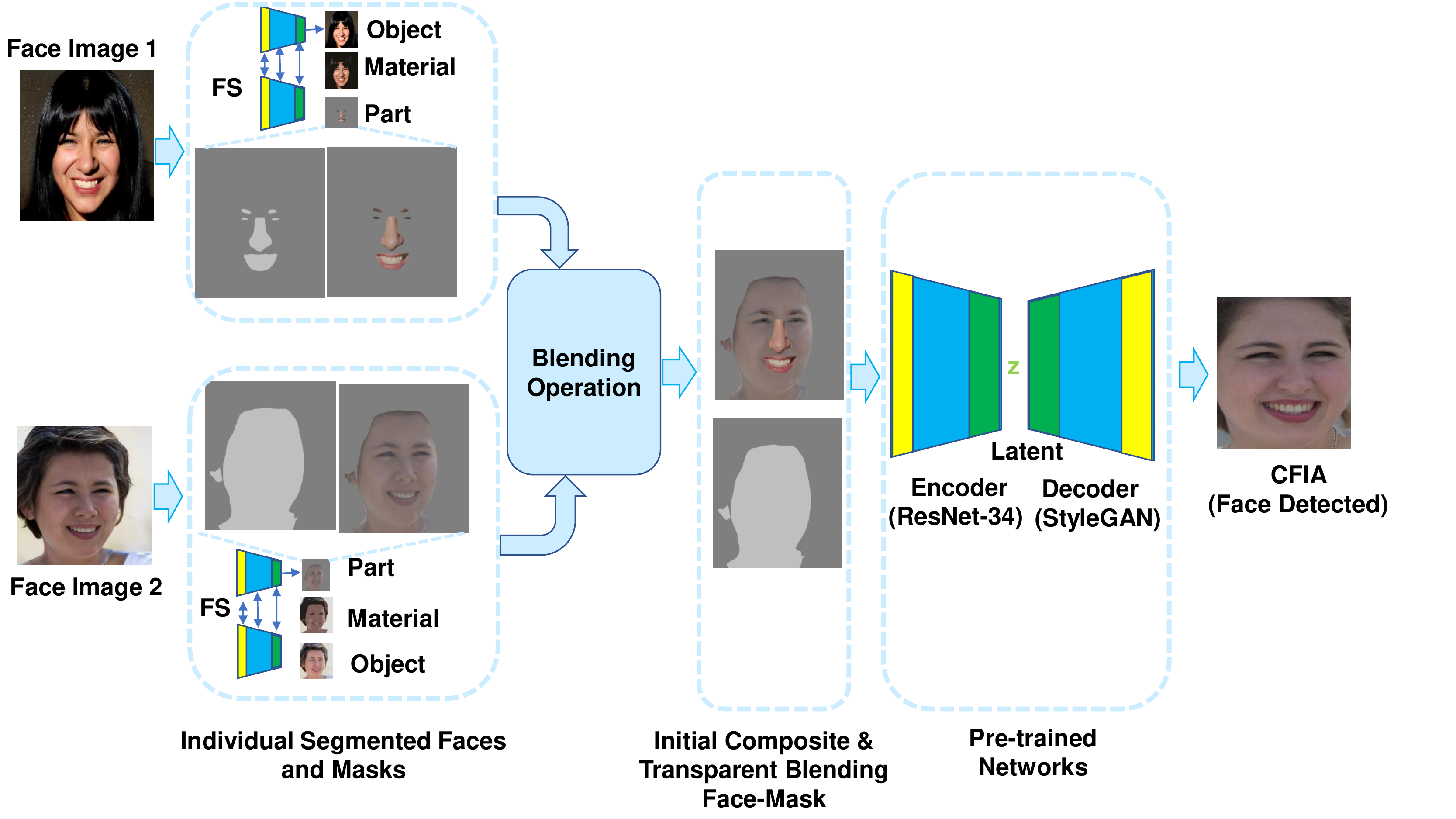}
\end{center}
\caption{Block diagram of the proposed approach where { FS is based on UPerNet Face Segmenter from Zhou et al.~\cite{Zhou17_ADE20KSceneParsing_CVPR}}, the Encoder is based on Resnet-34~\cite{Chai21LatentCompositeICLR}, and Decoder is based on StyleGAN~\cite{Karras19_StyleGAN-CVPR} and the encoder-decoder synthesizes the final composite.
}
\label{fig:BlockDiagram:ProposedMethod}
\end{figure*}
\begin{figure*}[h!]
\begin{center}
\includegraphics[width=0.85\linewidth]{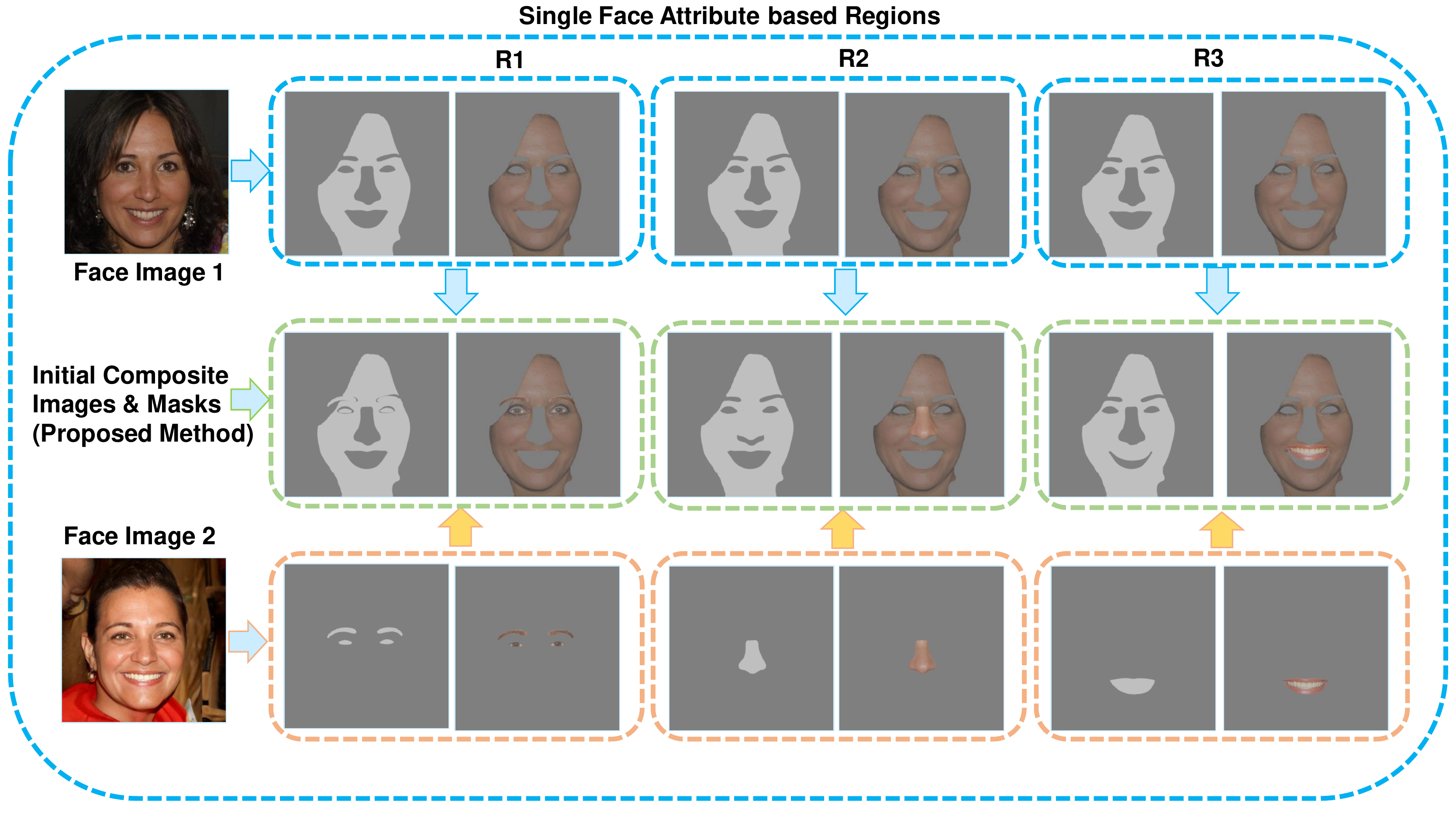}
\includegraphics[width=0.85\linewidth]{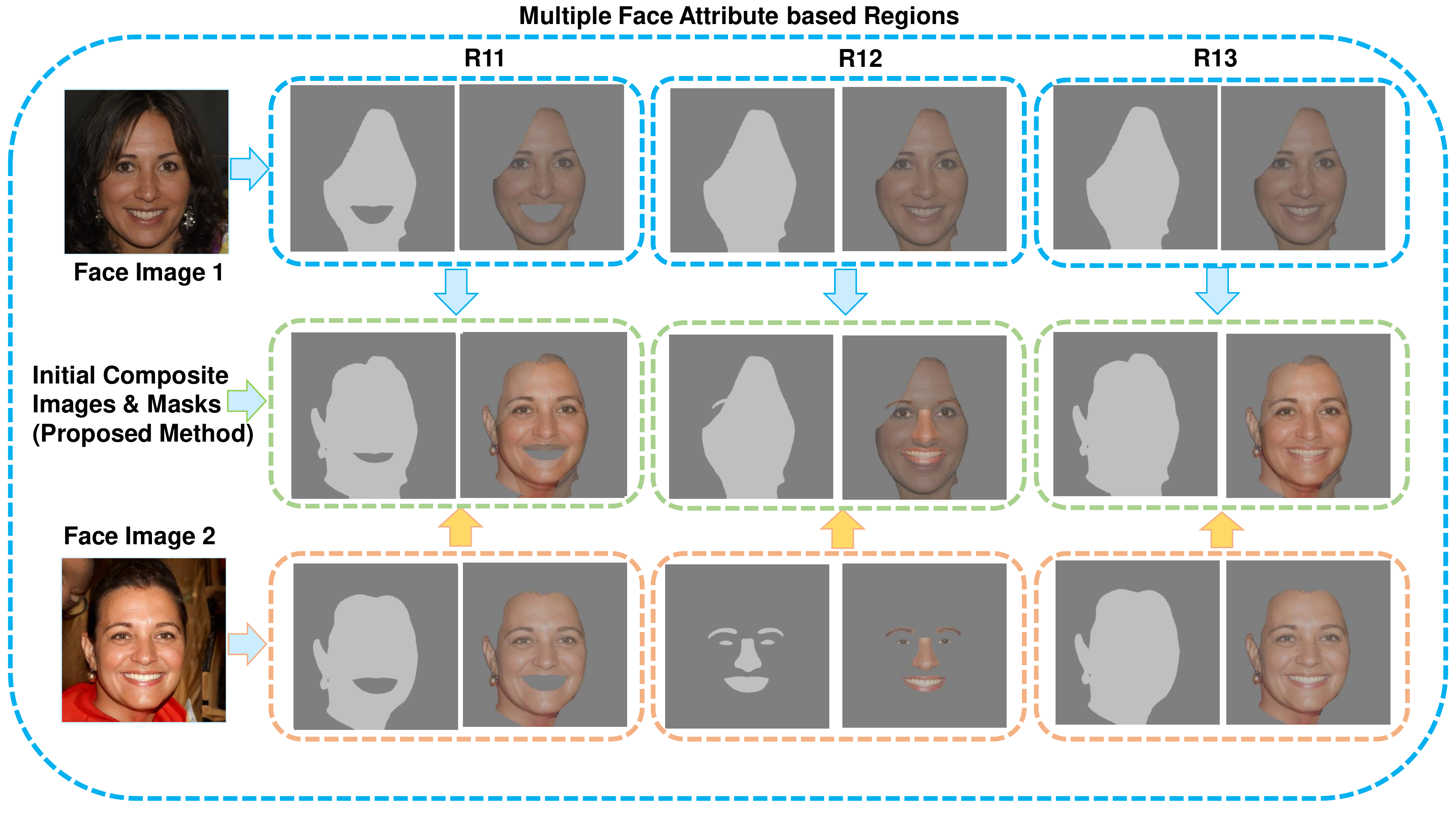}
\end{center}
   \caption{Illustration showing the comparison between the single face attribute based regions and multiple face attribute based regions for the generation of the initial composite using the proposed method.}
\label{fig:SingleRegionandMultipleRegion}
\end{figure*}

Thus, motivated by the limitations of the existing method \cite{Qin21-MorphParts-TBIOM}, we aim to generate the composite images in a  fully automatic fashion using GANs. Even though the GANs are extensively used for full face morphing attack generation \cite{Zhang21-MIPGAN-TBIOM, Venkatesh20-CANGAN-IWBF, Damer21-RegenMorph-AVCSpringer}, the composite (or facial attribute) based attack generation is presented for the first time in this work. 
{{The recent work by  Chai et al.~\cite{Chai21LatentCompositeICLR} presented a  highly realistic facial image synthesis with missing regions using GAN-inversion}}. In this work, we modified the approach from Chai et al.~\cite{Chai21LatentCompositeICLR} to generate the CFIA samples {with the primary motivation to demonstrate the vulnerabilities of FRS to CFIA. Further, we exhaustively varied the regions based on facial attributes to evaluate their attack potential.}  The proposed method for CFIA generation is designed to consider the optimal pairing of the input images used during the compositing process to ensure high-quality CFIA generation. 
The CFIA samples are generated based on multiple facial attributes. Both single and multiple facial attributes are blended using the transparent (or real) value that can further improve the attack potential and challenge the detection of CFIA samples. Use of  facial attributes or partial morphing will not alter the entire face and thus results in less distortion  because the proposed CFIA approach will only choose the facial attributes from the contributory subjects and then synthesize the rest of the facial image using GAN. Hence, the generated  CFIA images  are challenging to be detected by expert border guards. 
The key contributions of our proposed method are as follows:
\begin{itemize}
\item {{We propose a novel framework for Composite Face Image Attack (CFIA) generation using regression and GAN-based image synthesis}}. {{The primary motivation of the proposed CFIA approach is to generate high-quality facial attack images using facial attributes with high attack potential. Further, it should be challenging to detect by both human and automatic morph detection techniques. }} Therefore, we generate CFIA based on single and multiple face attributes for given contributory data subjects. Further, we propose a transparent blending to improve the attack potential of the generated CFIA. 
{{ Thus, we introduce $526$ different types of CFIA based on various combinations of facial attributes from contributory data subjects.}} 
\item We present a new CFIA dataset generated using $1000$ unique data subjects (synthetic identities). The dataset consists of $526000$ CFIA samples and $2000$ bona fide samples. 
\item {{We present extensive vulnerability analysis on the newly generated CFIA dataset using deep learning-based FRS.}}. We also introduce an  vulnerability metric called Generalized Morphing Attack Potential (G-MAP) to benchmark the attack potential effectively by considering real-life scenarios.
\item We present the perceptual image quality analysis of the CFIA dataset using the Peak-Signal-to Noise Ratio (PSNR) and Structural Similarity Index Measure (SSIM) to benchmark the quality of the generated CFIA samples on a sub-set of the CFIA dataset with 14 unique combinations selected from the 526 combinations.
\item We present the human observer study on the newly generated CFIA dataset (subset of 14 combinations) with 43 observers with and without face image manipulation detection background. 
\item We present extensive experiments benchmarking the performance  to automatically detect the CFIA (subset of 14 combinations) using three different existing single image based face MAD techniques. 
\item  The CFIA dataset, together with the source code of the proposed method, will be made publicly available to enable the reproducibility of the results presented in this paper \url{https://github.com/jagmohaniiit/LatentCompositionCode}.   
\end{itemize}

In the rest of the paper we introduce the proposed method in Section~\ref{sec:Proposed}, discussion on database generation methodology in presented in Section~\ref{sec:DataGen},   vulnerability analysis and G-MAP is discussed in  Section~\ref{sec:VulFRS}, Section~\ref{sec:pecpQuality} discuss the quantitative results of the perceptual quality evaluation, human observer study is discussed in Section~\ref{sec:Human},  and  discussion on CFIA  detection (CAD) is presented  in the Section~\ref{sec:CAD}. Lastly, the  Section~\ref{sec:Conc} draws the conclusions and future-work.


\section{Proposed CFIA Generation Technique}\label{sec:Proposed}

 Figure~\ref{fig:BlockDiagram:ProposedMethod} shows the block diagram of the proposed CFIA method. The proposed CFIA method aims to automatically select single and multiple facial attribute regions from the given face images and blend them to generate a composite face image. The proposed CFIA method consists of three main functional blocks (1) generation of individual segmented faces and masks from given face images, (2) computation of the initial composite image and transparent blending face mask and (3)  final CFIA generation based on pre-trained GANs.
\subsection{Individual Segmented Faces and Masks}\label{sec:IFCS}
The proposed CFIA composite image generation is based on the different facial parts from the two contributory data subjects (e.g., skin from the first data subject and eyes from the second data subject). Therefore, we  employed a high precision and accurate method to segment different facial parts. In this work, we choose the unified parsing network (UPerNet)~\cite{Xiao18-UnifiedParsing-ECCV} for automatic facial region segmentation, which is denoted as $\mathbb{FS}$. UPerNet~\cite{Xiao18-UnifiedParsing-ECCV} is based on multi-task learning and semantic segmentation to achieve high-quality results on facial segmentation and classification tasks. Thus, given the face image, UPerNet~\cite{Xiao18-UnifiedParsing-ECCV} provides six facial regions (or attribute) masks, including Skin (S), Eye (E), Nose (N), Mouth (M), Hair (H), and Background (B). 

In this work, we have considered only two contributory face images based on real-time use-case applicability (for e.g. attacks on eMRTD or ID cards) \cite{Qin21-MorphParts-TBIOM,Venkatesh21-FaceMorphingSurvey-TTS}. We denote the first contributory face image by $F_1$ and the corresponding part-based segmented masks obtained using UPerNet~\cite{Xiao18-UnifiedParsing-ECCV} be $SM1_{i}$, where $i = \{1,2, \ldots, 6$\} and its corresponding segmented image be $IS1_{i}$. Similarly,  the second  contributory face image be  $F_2$ and the corresponding part-based segmented masks  be $SM2_{j}$, where $j = \{1,2, \ldots, 6$\}  and its corresponding segmented image be $IS2_{j}$. 
The face region segmentation process to obtain individual segments can be expressed as follows: 
\begin{equation}\label{eqn1}
\begin{split}
\{SM1_{i},IS1_{i}\} = \mathbb{FS}(F_1), \forall i = \{1,2,\ldots, 6 \} \\
\{SM2_{j},IS2_{j}\} = \mathbb{FS}(F_2), 
\forall j = \{1,2,\ldots, 6 \} 
\end{split}
\end{equation}
\begin{table}[h!]

    \scriptsize
    \centering
      \resizebox{.7\linewidth}{!}
      {
      \scriptsize
      \begin{tabular}{|c|c|c|} 
    \hline
   {\bf{ CFIA Region Index}} &  {\bf{ Output Segments}} & {\bf{Possible Pairs (Unique)}}  \\ \hline
     \multicolumn{3}{|c|}{\bf{One Combinations}} \\ \hline
    1 & 2 &   ${{5}\choose{1}}\times{{5}\choose{1}}=25 (13)$ \\ [-0.5em]
    & & \\ \hline
    \multicolumn{3}{|c|}{\bf{Two Combinations}} \\ \hline
  2& 3  &  ${{5}\choose{2}}\times{{5}\choose{1}}=50 (26)$ \\ [-0.5em]
  & & \\ \hline
  3& 4  &  ${{5}\choose{2}}\times{{5}\choose{2}}=100 (100) $ \\ [-0.5em]
  & & \\ \hline
  \multicolumn{3}{|c|}{\bf{Three Combinations}} \\ \hline
  4& 4 &  ${{5}\choose{3}}\times{{5}\choose{1}}=50 (50)$ \\ [-0.5em]
   & & \\ \hline
 5& 5   &  ${{5}\choose{3}}\times{{5}\choose{2}}=100 (78) $ \\ [-0.5em]
  & & \\ \hline
 6& 6   &  ${{5}\choose{3}}\times{{5}\choose{3}}=100 (86) $ \\ [-0.5em]
  & & \\ \hline
  \multicolumn{3}{|c|}{\bf{Four Combinations}} \\ \hline
  7&5  &  ${{5}\choose{4}}\times{{5}\choose{1}}=25 (25) $ \\ [-0.5em]
  & & \\ \hline
  8&6  &  ${{5}\choose{4}}\times{{5}\choose{2}}=50 (50)$ \\ [-0.5em]
   & & \\ \hline
   9&7  &  ${{5}\choose{4}}\times{{5}\choose{3}}=50 (47) $ \\ [-0.5em]
   & & \\ \hline
    10&8  &  ${{5}\choose{4}}\times{{5}\choose{4}}=25 (25) $ \\ [-0.5em]
     & &\\ \hline
     \multicolumn{3}{|c|}{\bf{Five Combinations}} \\ \hline
    11&6  &  ${{5}\choose{5}}\times{{5}\choose{1}}=5 (5)$ \\ [-0.5em]
   &  &\\ \hline
    12&7  &  ${{5}\choose{5}}\times{{5}\choose{2}}=10 (10)$ \\ [-0.5em]
    & & \\ \hline
    13&8  &  ${{5}\choose{5}}\times{{5}\choose{3}}=10 (10) $ \\ [-0.5em]
     && \\ \hline
     14&9  &  ${{5}\choose{5}}\times{{5}\choose{4}}=5 (5) $ \\ [-0.5em]
      && \\ \hline
      15&10  &  ${{5}\choose{5}}\times{{5}\choose{5}}=1 (1) $ \\ [-0.5em]
       && \\ \hline
       \multicolumn{3}{|c|}{\bf{Six Combinations}} \\ \hline
       16&12  &  ${{6}\choose{6}}\times{{6}\choose{6}}=1 (1)$ \\ [-0.5em]
        & & \\ \hline
   \multicolumn{3}{|c|}{\bf{Total Output Segments Possible}} \\ \hline
  \multicolumn{3}{|c|}{\bf{607}}  \\ \hline
   \multicolumn{3}{|c|}{\bf{Total Unique Segments Possible}} \\ \hline
  \multicolumn{3}{|c|}{\bf{526}}  \\ \hline
  \end{tabular}
    }
    \caption{Table showing the generation process of 526 unique CFIA combinations which are listed in detail Table~\ref{table:regionList}.}.
   \label{table:overallSegmentsExhaustive}
    \end{table}

{{Based on these six part-based segmentation masks (or region or facial attributes), we generate an exhaustive list of combinations from $SM1_{i}$ and $SM2_{j}$ that resulted in $526$ unique CFIA samples as listed in Table~\ref{table:regionList}. It needs to be pointed out that the selected areas are exhaustive as listed in Table \ref{table:overallSegmentsExhaustive}.  Table \ref{table:overallSegmentsExhaustive} mentions the CFIA Region Index, through which we give a numerical index to the output segments so that overall, it increases with the number of combinations. 
E.g., if we consider two combinations case, we  select two regions from $SM1$ and choose a maximum of two regions out of six (in a step-wise manner) from $SM2$. Therefore, in two combinations case (see Table \ref{table:overallSegmentsExhaustive}), we have CFIA region index 2, in which, we select 2 regions from $SM1$ and one region from $SM2$. Similarly, for CFIA region index 3 we select 2 regions from $SM1$ and 2 regions from $SM2$. We repeat this process for the various combinations of regions (or facial attributes), such that CFIA region index 2 results in 50 combinations corresponding to 3 output segments.  Similarly, CFIA region index 3 
results in 100 combinations corresponding to 4 output segments. These steps are repeated for different CFIA region indexes from 1 to 16, resulting in a total of 607 combinations.  
However, out of 607 combinations some of the combinations are redundant. For example, selecting face attributes from  $SM1$ and $SM2$ such as SE-NM (SkinEyes-NoseMouth) can occur in two ways, firstly Skin, Eyes from $SM1$ and Nose, Mouth from $SM2$ and secondly SEN-M (SkinEyesNose-Mouth), Skin, Eyes and Nose from $SM1$ and Mouth from $SM2$ resulting in a redundant combination.  Therefore, we removed all such redundant combinations and considered unique combinations. Hence, we generate  526 unique CFIA samples corresponding to two unique facial identities.} }

 \begin{table*}[h!]
    \centering
      \resizebox{0.65\linewidth}{!}
      {
      \begin{tabular}{|c|c|c|c|c|c|c|} 

\hline
\multicolumn{7}{|c|}{{\bf{{Region List}}}} \\ \hline

{\bf{{\textcolor{blue}{S1-S2}}}} &  {\bf{{\textcolor{blue}{S1-S2}}}} &  {\bf{{\textcolor{blue}{S1-S2}}}} & {\bf{{\textcolor{blue}{S1-S2}}}} &  {\bf{{\textcolor{blue}{S1-S2}}}} & {\bf{{\textcolor{blue}{S1-S2}}}} & {\bf{{\textcolor{blue}{S1-S2}}}}  \\ \hline
\hline
E-H  & H-E  & H-H  & H-M  & H-N  & H-S  & M-H \\ \hline
N-H  & S-H  & S-E & S-N & S-M & S-S & EM-H  \\ \hline EN-H  & HE-E  & HE-H  & HE-M & HE-N  & HE-S & HM-E\\ \hline    HM-H  & HM-M  & HM-N  & HM-S & HN-E  & HN-H  & HN-M \\ \hline 
 HN-N  & HN-S  & HS-E  & HS-H & HS-M  & HS-N  & HS-S\\ \hline 
   NM-H  & SE-H  & SM-H  & SN-H & EM-EM  & EM-EN  & EM-HE \\  \hline 
   EM-HM  & EM-HN  & EM-HS  & EM-NM & EM-SE  & EM-SM  & EM-SN \\ \hline 
  EN-EM  & EN-EN  & EN-HE  & EN-HM  & EN-HN  & EN-HS  & EN-NM \\ \hline 
   EN-SE  & EN-SM  & EN-SN  & HE-EM & HE-EN  & HE-HE  & HE-HM\\ \hline 
 HE-HN  & HE-HS  & HE-NM  & HE-SE & HE-SM  & HE-SN  & HM-EM \\ \hline 
HM-EN  & HM-HE  & HM-HM  & HM-HN & HM-HS  & HM-NM  & HM-SE \\ \hline 
  HM-SM  & HM-SN  & HN-EM  & HN-EN & HN-HE  & HN-HM  & HN-HN \\ \hline 
  HN-HS  & HN-NM  & HN-SE  & HN-SM & HN-SN  & HS-EM  & HS-EN \\ \hline 
  HS-HE  & HS-HM  & HS-HN  & HS-HS & HS-NM  & HS-SE  & HS-SM \\ \hline 
  HS-SN  & NM-EM  & NM-EN  & NM-HE & NM-HM  & NM-HN  & NM-HS \\ \hline 
  NM-NM  & NMS-E  & NMS-M  & NMS-N & SEE-M  & SEE-N  & SEH-E \\ \hline 
  SEH-M  & SEH-N  & SEH-S  & SEN-M & SES-E  & SES-M  & SES-N\\ \hline 
   SME-M  & SME-N  & SMH-E  & SMH-M & SMH-N  & SMH-S  & SMN-M\\ \hline 
  SMS-E  & SMS-M  & SMS-N  & SNE-M & SNE-N  & SNH-E  & SNH-M \\ \hline 
  SNH-N  & SNH-S  & SNN-M  & SNS-E & SNS-M  & SNS-N  & ENM-E \\ \hline 
  ENM-H  & ENM-M  & ENM-N  & ENM-S & HEM-E  & HEM-H  & HEM-M\\ \hline 
  HEM-N  & HEM-S  & HEN-E  & HEN-H & HEN-N  & HEN-S  & HNME \\ \hline 
  HNM-H  & HNM-M  & HNM-N  & HNM-S & HSE-E  & HSE-H  & HSE-S \\ \hline 
  HSM-E  & HSM-H  & HSM-M  & HSM-N & HSM-S  & HSN-E  & HSN-H \\ \hline 
HSN-N  & HSN-S  & SEM-E  & SEM-H & SEM-M  & SEM-N  & SEM-S \\ \hline 
SEN-E  & SEN-H  & SEN-N  & SEN-S & SNM-E  & SNM-H  & SNM-M \\ \hline 
 SNM-N  & SNM-S  & ENM-HE  & ENM-HM & SEN-EM & SEN-EN & ENM-HN  \\ \hline 
  ENM-HS  & HEM-EM  & HEM-EN  & HEM-HE & HEM-HM  & HEM-HN & HEM-HS  \\ \hline HEM-NM  & HEM-SE  & HEM-SM  & HEM-SN  & HEN-EM  & HEN-EN & HEN-HE  \\ \hline 
HEN-HM  &  HEN-HN  & HEN-HS  & HEN-NM  & HEN-SE  & HEN-SM & HEN-SN   \\ \hline 
HNM-EM  & HNM-EN  & HNM-HE  & HNM-HM  & HNM-HN  & HNM-HS & HNM-NM  \\ \hline 
 HNM-SE  & HNM-SM  & HNM-SN  & HSE-EM  & HSE-EN  & HSE-HE & HSE-HM  \\ \hline 
 HSE-HN  & HSE-HS  & HSE-NM  & HSE-SE  & HSE-SM  & HSE-SN & HSM-EM   \\ \hline 
 HSM-EN  & HSM-HE  & HSM-HM  & HSM-HN  & HSM-HS  & HSM-NM & HSM-SE  \\ \hline 
 HSM-SM  & HSM-SN  & HSN-EM  & HSN-EN  & HSN-HE  & HSN-HM & HSN-HN  \\ \hline 
 HSN-HS  & HSN-NM  & HSN-SE  & HSN-SM  & HSN-SN  & SEM-HE & SEM-HM  \\ \hline 
 SEM-HN  & SEM-HS  & SEN-HE  & SEN-HM  & SEN-HN  & SEN-HS & SNM-HE  \\ \hline 
 SNM-HM  & SNM-HN  & SNM-HS  & ENM-HEM  & ENM-HEN  & ENM-HNM & ENM-HSE \\ \hline 
 ENM-HSM  & ENM-HSN  & HEM-ENM  & HEM-HEM  & HEM-HEN  & HEM-HNM & HEM-HSE  \\ \hline 
 HEM-HSM  & HEM-HSN  & HEM-SEM  & HEM-SEN  & HEM-SNM  & HEN-ENM & HEN-HEM   \\ \hline 
 HEN-HEN  & HEN-HNM  & HEN-HSE  & HEN-HSM  & HEN-HSN  & HEN-SEM & HEN-SEN   \\ \hline 
 HEN-SNM  & HNM-ENM  & HNM-HEM  & HNM-HEN  & HNM-HNM  & HNM-HSE & HNM-HSM   \\ \hline 
 HNM-HSN  & HNM-SEM  & HNM-SEN  & HNM-SNM  & HSE-ENM  & HSE-HEM & HSE-HEN   \\ \hline 
 HSE-HNM  & HSE-HSE  & HSE-HSM  & HSE-HSN  & HSE-SEM  & HSE-SEN & HSE-SNM   \\ \hline 
 HSM-ENM  & HSM-HEM  & HSM-HEN  & HSM-HNM  & HSM-HSE  & HSM-HSM & HSM-HSN  \\ \hline 
 HSM-SEM  & HSM-SEN  & HSM-SNM  & HSN-ENM  & HSN-HEM  & HSN-HEN & HSN-HNM  \\ \hline 
 HSN-HSE  & HSN-HSM  & HSN-HSN  & HSN-SEM  & HSN-SEN  & HSN-SNM & SEM-HEM  \\ \hline 
 SEM-HEN  & SEM-HNM  & SEM-HSE  & SEM-HSM  & SEM-HSN  & SEN-HEM & SEN-HEN  \\ \hline 
 SEN-HNM  & SEN-HSE  & SEN-HSM  & SEN-HSN  & SNMHEM  & SNM-HEN & SNM-HNM  \\ \hline 
 SNM-HSE  & SNM-HSM  & SNM-HSN  & SEN-SEM & SEN-SEN & HENM-E  & HENM-H   \\ \hline 
 HENM-M & HENM-N  & HENM-S  & HSEM-E  & HSEM-H  & HSEMM  & HSEM-N   \\ \hline 
 HSEM-S & HSEN-E & HSEN-H  & HSEN-N  & HSEN-S  & HSNME  & HSNM-H   \\ \hline 
 HSNM-M & HSNM-N  & HSNM-S  & SENM-E  & SENM-H  & SENM-M  & SENM-N    \\ \hline 
 SENM-S & HENM-EM & HENM-EN  & HENM-HE  & HENM-HM  & HENM-HN  & HENM-HS     \\ \hline 
 HENM-NM & HENM-SE & HENM-SM  & HENM-SN  & HSEM-EM  & HSEM-EN  & HSEM-HE    \\ \hline 
HSEM-HM & HSEM-HN & HSEM-HS  & HSEM-NM  & HSEM-SE  & HSEM-SM  & HSEM-SN    \\ \hline 
 HSEN-EM & HSEN-EN & HSEN-HE  & HSEN-HM  & HSEN-HN  & HSEN-HS  & HSEN-NM    \\ \hline 
HSEN-SE & HSEN-SM & HSEN-SN  & HSNM-EM  & HSNM-EN  & HSNM-HE  & HSNM-HM    \\ \hline 
HSNM-HN & HSNM-HS & HSNM-NM  & HSNM-SE  & HSNM-SM  & HSNM-SN  & SENM-EM    \\ \hline 
SENM-EN & SENM-HE & SENM-HM  & SENM-HN  & SENM-HS  & SENM-NM  & SENM-SE    \\ \hline 
 SENM-SM & SENM-SN &  SENM-ENM  & HENM-ENM  & HENM-HEM  & HENM-HEN  & HENM-HNM   \\ \hline 
HENM-HSE  & HENM-HSM & HENM-HSN  & HENM-SEM  & HENM-SEN  & HENM-SNM  & HSEM-ENM   \\ \hline 
HSEM-HEM  & HSEM-HEN & HSEMH-NM  & HSEMH-SE  & HSEMH-SM  & HSEM-HSN  & HSEM-SEM   \\ \hline 
HSEM-SEN  & HSEM-SNM & HSEN-ENM  & HSEN-HEM  & HSEN-HEN  & HSEN-HNM  & HSEN-HSE   \\ \hline 
 HSEN-HSM  & HSEN-HSN & HSEN-SEM  & HSEN-SEN  & HSEN-SNM  & HSNM-ENM  & HSNM-HEM   \\ \hline 
HSNM-HEN  & HSNM-HNM & HSNM-HSE  & HSNM-HSM  & HSNM-HSN  & HSNM-SEM  & HSNM-SEN   \\ \hline 
 HSNM-SNM  & SENM-HEM & SENM-HEN  & SENM-HNM  & SENM-HSE  & SENM-HSM  & SENM-HSN   \\ \hline 
HENMH-ENM  & HENMH-SEM & SENM-SENM & HENM-HSEN  & HENM-HSNM  & HENM-SENM  & HSEM-HENM     \\ \hline 
HSEM-HSEM  & HSEM-HSEN & HSEM-HSNM & HSEM-SENM  & HSEN-HENM  & HSEN-HSEM  & HSENH-SEN    \\ \hline 
HSENH-SNM  & HSEN-SENM & HSNM-HENM  & HSNM-HSEM  & HSNM-HSEN  & HSNM-HSNM  & HSNM-SENM     \\ \hline 
SENM-HENM  & SENM-HSEM & SENM-HSEN & SENM-HSNM  & HSENM-E  & HSENM-H  & HSENM-M    \\ \hline 
HSENM-N  & HSENM-S & HSENME-M & HSENME-N  & HSENMH-E  & HSENMH-M  & HSENMH-N     \\ \hline 
HSENMH-S  & HSENMN-M & HSENMS-E & HSENMS-M  & HSENMS-N  & HSENMEN-M  & HSENMH-EM     \\ \hline 
HSENMH-EN  & HSENMH-NM & HSENMH-SE & HSENMH-SM  & HSENMH-SN  & HSENMS-EM  & HSENMS-EN     \\ \hline 
HSENMS-NM  & HSENMH-ENM & HSENMH-SEM & HSENMH-SEN & HSENMH-SNM & HSENM-SENM & HSENM-HSENM   \\ \hline
 HBSENM-HBSENM &&&&&& \\ \hline
\end{tabular}
    }\caption{Exhaustive List of Regions used for Composition where the compositions S1 are used for Subject 1 and S2 are used for Subject 2 where the facial attributes are B=Background, S=Skin, E=Eye, N=Nose and M=Mouth. The compositions listed in left to right order are in increasing order of Composition Region Index (for Composition Region Index, please refer Table~\ref{table:overallSegmentsExhaustive})}.
   \label{table:regionList}
    \end{table*}

\subsection{\bf{Initial Composite Image and Transparent Blending Face-Mask}}
In the next step, we generate the initial composite image and transparent blending of face segments by applying the blending operation on the individual segmented faces ($IS1_{i}$ \& $IS2_{j}$) and their corresponding masks ($SM1_{i}$ \& $SM2_{j}$) from contributory data subjects ($F_1$ \& $F_2$). The blending operation is carried out independently for the mask and the individual segmented faces. The blended mask  $m_c$ is generated by a simple union operation that can represent the combined facial region from $SM1_{i}$ and $SM2_{j}$ as described in Equation~\ref{eqn2}. The generation of the initial composite image ($IC$) is done in three consecutive steps shown in Equation~\ref{eqn3}, where first $IC$ is initialized $0$, then in the next step, $IC$ is updated using the compositing equation with the segmented region ($IS1_{i}$) from the data subject $F_1$ as input. Finally, $IC$ is updated using the compositing equation with the segmented region ($IS2_{j}$) from data subject $F_2$, and its segmentation masks $SM2_{j}$  as an input. These steps are mathematically presented in Equation~\ref{eqn3}.  

\begin{equation}\label{eqn2}
m_c  = SM1_{i} \bigcup SM2_{j}   
\end{equation}

\begin{equation}\label{eqn3}
\begin{split}
IC &= 0\\
IC &= IS1_{i}\\
IC &= IS2_{j}+(1-SM2_{j}){\times}IC
\end{split}
\end{equation}

Figure~\ref{fig:SingleRegionandMultipleRegion} shows the qualitative results of the initial composite image and the corresponding mask for both single-face attribute-based composite regions \& multiple-face attribute-based composite regions.

\subsection{Final CFIA  samples Generation}
\begin{figure*}[htb]
\begin{center}
\includegraphics[width=0.85\linewidth]{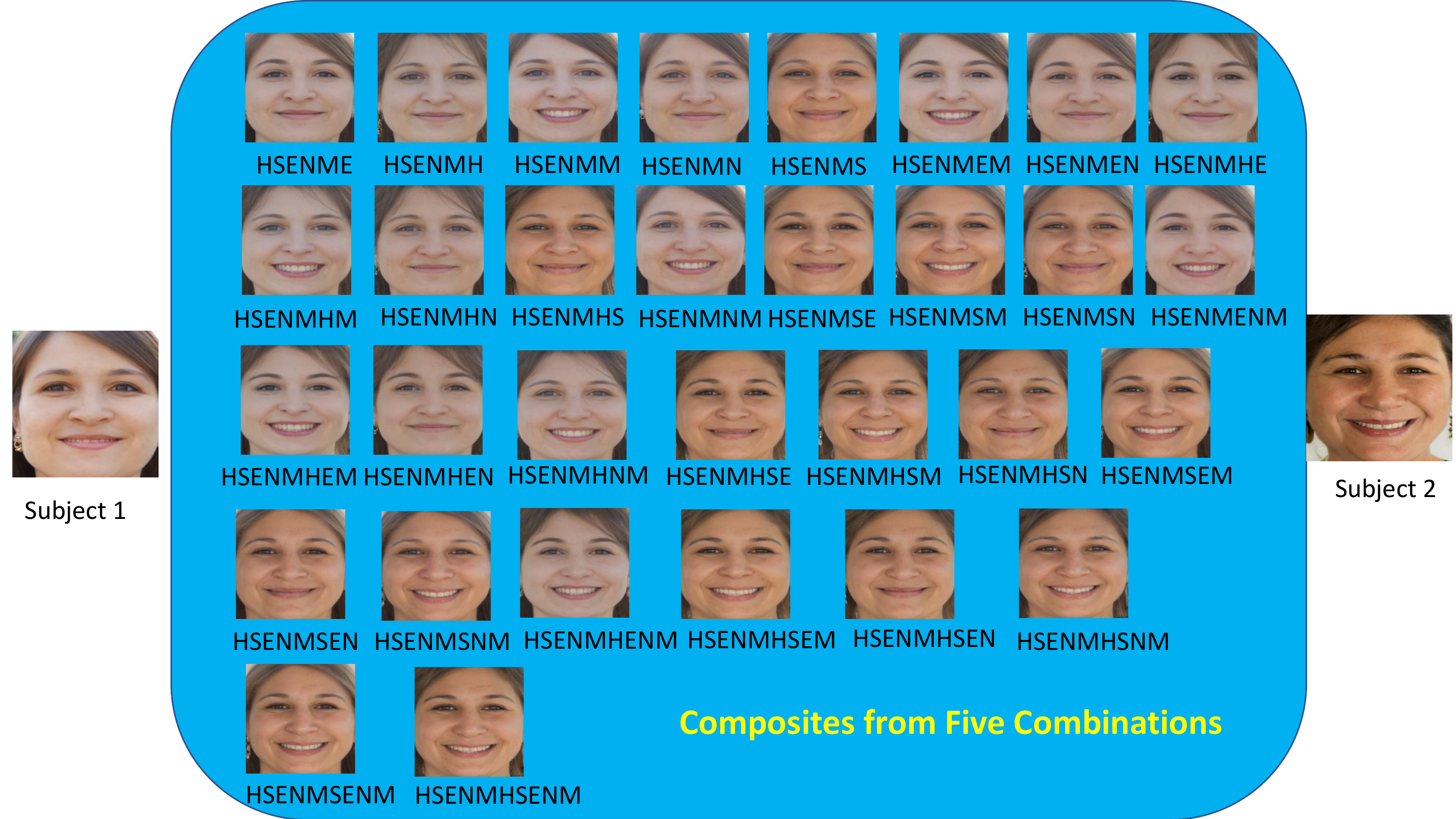}
\end{center}
   \caption{Illustration showing five combinations based composites.}
\label{fig:fiveCombinations}
\end{figure*}

Once the initial composite image and the transparent blending face mask are generated, we generate the final CFIA samples  using the image inpainting based on pre-trained regressor  and GAN~\cite{Chai21LatentCompositeICLR}. The input composite image and its mask are passed through a pre-trained encoder ($\mathbb{E}$)  and then to the decoder ($\mathbb{G}$) to generate the final composite image ($FCI$). The process of generating the CFIA sample is as indicated in Equation~\ref{eqn6}.
\begin{equation}\label{eqn6}
CFIA = \mathbb{D}(\mathbb{E}(IC,m_c))
\end{equation}
 The encoder network ($\mathbb{E}$) selected in our work is pre-trained Resnet-34~\cite{Chai21LatentCompositeICLR}, and the decoder network ($\mathbb{G}$) is a pre-trained StyleGAN-I decoder which was trained on FFHQ dataset~\cite{Karras19_StyleGAN-CVPR}. The primary motivation for the choice of the encoder and decoder networks was that image to latent conversion is posed as a regression problem~\cite{Chai21LatentCompositeICLR}. Further, it is found that Resnet-34 is suitable for regressing the latent from a face image with missing information and renders the high-quality face image. Lastly, we use the decoder ($\mathbb{D}$) based on StyleGAN-I as it provides a linear latent subspace~\cite{Shen20_InterfaceGAN_TPAMI}. Hence, reconstruction from the generated latent is of good perceptual quality even with missing information in the input image. Figure~\ref{fig:fiveCombinations} shows example  results corresponding to five combinations  generated using the proposed method. For the simplicity, we have included the illustration for five combination and full 526 CFIA samples are included in the supplement material. 

\section{CFIA Database Generation}\label{sec:DataGen}

\begin{figure*}[htb]
\begin{center}
\includegraphics[width=0.85\linewidth]{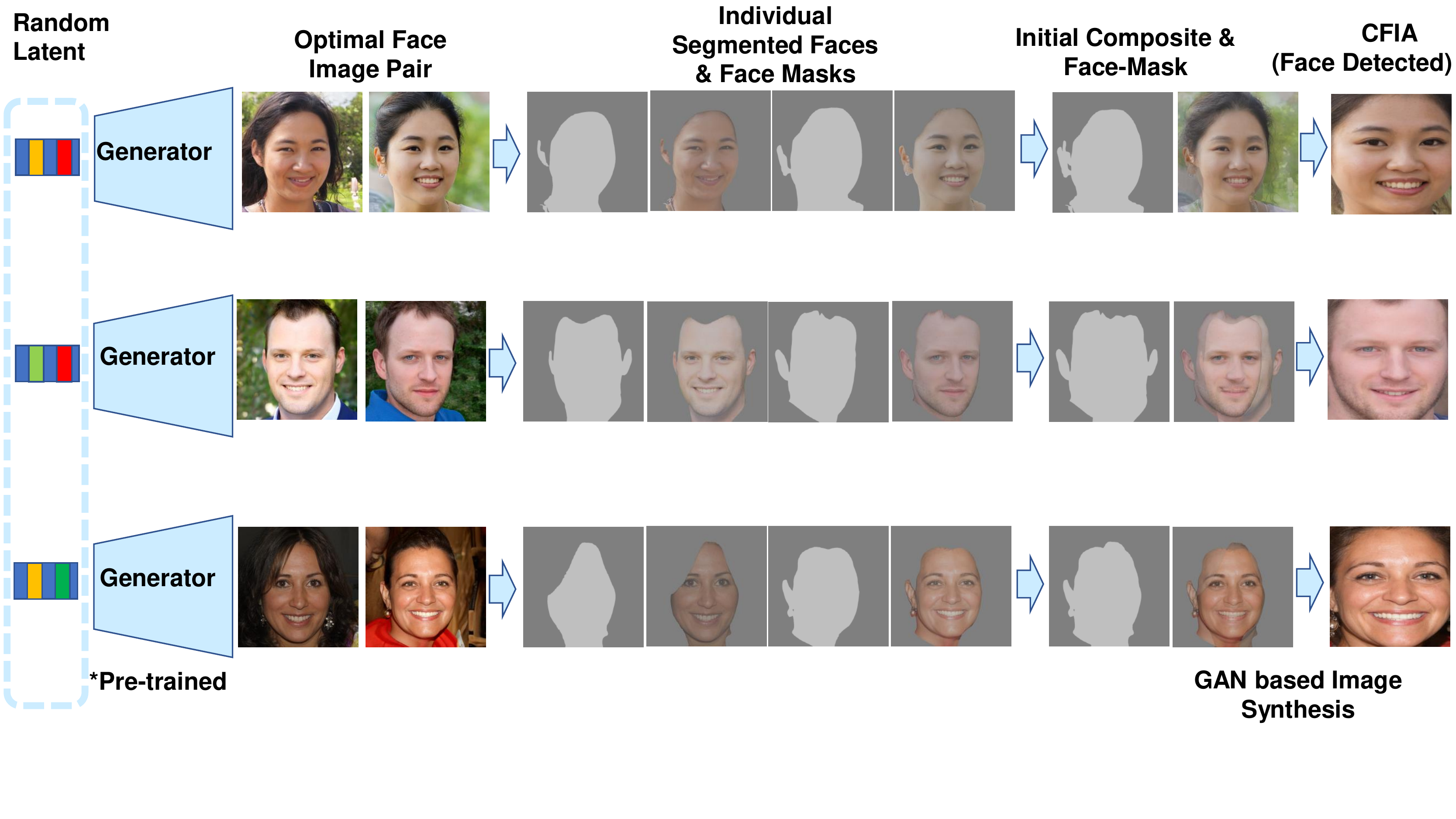}
\end{center}
   \caption{Illustration showing the CFIA  dataset generation process}
\label{fig:blockDiagram1}
\end{figure*}

This section presents the dataset generation process used to evaluate the proposed composite image generation. Owing to the ethical and legal challenges with face biometric datasets that will eventually limit the distribution, in this work, we generate the synthetic face images corresponding to the unique identities using StyleGAN inversion~\cite{Chai21LatentCompositeICLR}. Earlier works~\cite{Sarkar20-VulnerabilityMAD-Arxiv,Zhang21-MIPGAN-TBIOM,Damer22-PrivacySynthetic-Arxiv} indicated that  generating the synthetic face images have demonstrated both realness in terms of quality, uniqueness and verification accuracy. Further, synthetic face images will overcome the need for privacy and legal limitations to make the database public, which is vital for reproducible research. Figure \ref{fig:blockDiagram1} illustrates the CFIA dataset generation process.

\begin{figure}[htp]
\begin{center}
\includegraphics[width=0.85\linewidth]{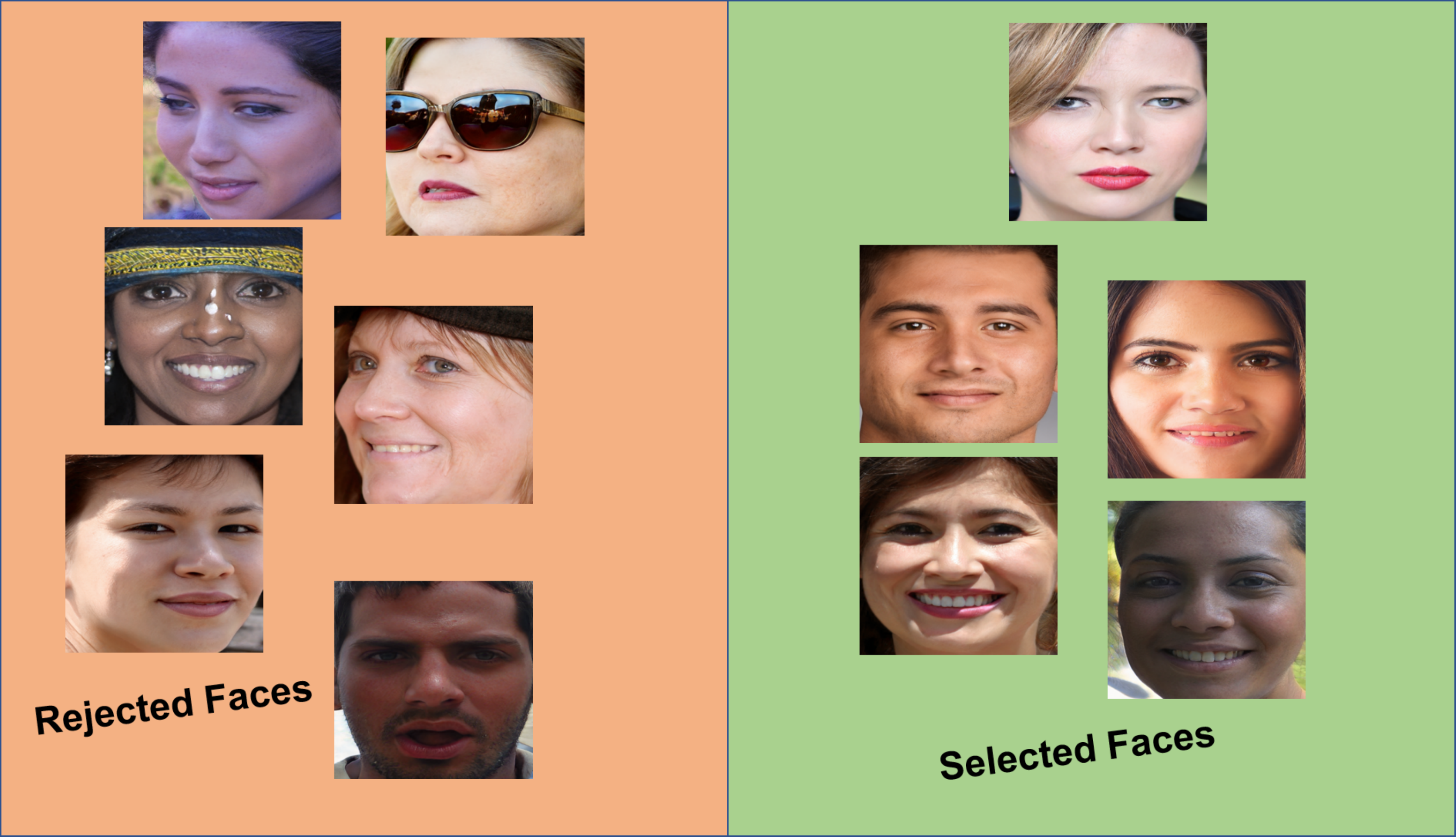}
\end{center}
   \caption{Illustration showing faces selected and rejected by our proposed frontal-pose detection Algorithm~\ref{Alg:NFComputeAlgo}. Note face images with glasses and GAN-based artifacts are rejected manually.}
\label{fig:FrontalFaceSelection}
\end{figure}

\subsection{Synthetic Face Image Generation}
Given a random latent vector, we use the approach from Chai et al.~\cite{Chai21LatentCompositeICLR} to generate a synthetic face corresponding to unique data subjects using StyleGAN inversion. We further perturb the random latent by an $\epsilon$ amount to generate the mated face image corresponding to the given identity. The choice of $\epsilon$ is made empirically, which is small enough not to alter the identity of the generated face. However, the generation of synthetic face images with corresponding mated face images with unique identities will result in non-ICAO compliant photos with glasses, non-frontal pose, and a non-neutral face expression, as shown in Figure~\ref{fig:FrontalFaceSelection}. { Therefore, it is necessary to detect the ICAO-compliant faces for which we select faces with frontal pose automatically and remove photos with glasses and non-neutral face expressions manually.}

{
 \begin{figure*}[htp]
\begin{center}
\includegraphics[width=0.9\linewidth]{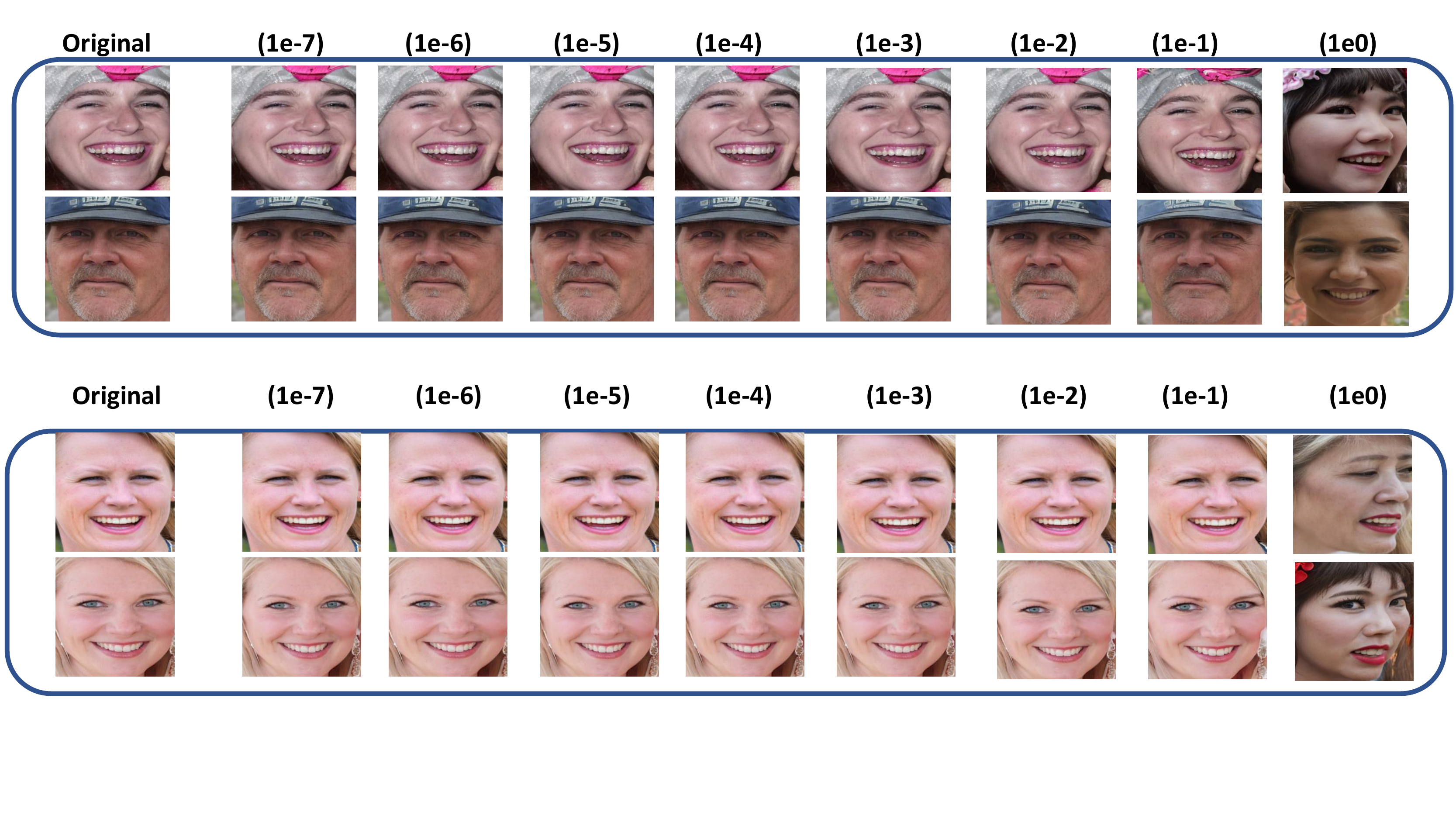}
\end{center}
   \caption{Illustration of the effect of perturbation based on epsilon ($\epsilon$) for synthetic face generation, note artifacts start appearing when ($\epsilon=0.1$) and results in change in identity when ($\epsilon=1$)}
\label{fig:latentComparison}
\end{figure*}
\subsection{Hyperparameters Selection}
\begin{table}[h!]
    \centering
      \resizebox{0.98\linewidth}{!}{
      \begin{tabular}{|c|c|c|} 
    \hline
    {\bf{Hyper-parameters}} & {\bf{{SOTA~\cite{Chai21LatentCompositeICLR}}}} & {\bf{Proposed Method}} \\ \hline
    {\bf{Frontal Pose Selection }} & No                                     & {\bf{Yes}}  \\ \hline
    {\bf{Optimal Pairing }} & No                                     & {\bf{Yes}}  \\ \hline
    {\bf{Epsilon($\epsilon$)}} & No & $10^{-7}$ \\ \hline
    {\bf{Alpha($\alpha$)}} & 1 & 0.5 \\ \hline
    \end{tabular}
    }\caption{ Different Hyper-parameters used for the proposed method. Note the proposed method modifies a large number of hyper-parameters compared with SOTA~\cite{Chai21LatentCompositeICLR}.}
    \label{table:hyperparametersModified}
    \end{table}
This section discusses the choices of the parameters associated with SOTA and the proposed method as tabulated in Table \ref{table:hyperparametersModified}. In total, we have four different hyperparameters that are discussed as follows:  
\begin{itemize}
    \item {\bf{Epsilon($\epsilon$)}}: The value of $\epsilon$ is empirically chosen as $10^{-7}$. Since values higher than $10^{-1}$ lead to artifacts and a sample of different identities as shown in Figure~\ref{fig:latentComparison}. Thus, we choose an $\epsilon$ conservatively.
    \item {\bf{Alpha}}: We choose $\alpha$=0.5 as it is known to create the highest vulnerability towards FRS for Face Morphing Image Attack (FMIA)~\cite{Venkatesh21-FaceMorphingSurvey-TTS}.
    of segments possible is shown in Table~\ref{table:regionList}
  \end{itemize}}  
   

\subsection{Frontal Face Pose Selection}
We have developed the algorithm to automatically select the ICAO compliant face images corresponding to each unique identity as indicated in the Algorithm~\ref{Alg:NFComputeAlgo}.
The primary motivation behind this algorithm is that the face in a frontal pose would have similar angles between Left-Eye, Nose, and Mouth (Left Part) and Right-Eye, Nose, and Mouth (Right Part). A slight change in the face pose from a frontal face to a profile face would result in a skew, which would cause these two angles to be different. The qualitative results of the proposed frontal face selection algorithm are as shown in Figure~\ref{fig:FrontalFaceSelection}. Since we are currently not interested in the computation of exact face pose, the heuristic works sufficiently well for our dataset, which does not consist of extreme face poses.
\begin{algorithm}[h!]
\caption{Non-Frontal Pose Identification}\label{Alg:NFComputeAlgo} 
\KwInput{Face Image with 5 Landmarks (Left-Eye ($LE$), Right-Eye ($RE$), Nose ($N$), Left-Mouth ($LM$), and Right-Mouth ($RM$)}
\KwOutput{True if Face Image is Frontal}
     \begin{algorithmic}[1]
        \STATE Compute the angle between the vectors of Left-Eye, Nose, and Left-Mouth, Nose \\$\theta_1$ $\gets$ $\arccos((\overrightarrow{LEN}\cdot\overrightarrow{LMN}))$.
         \STATE Compute the angle between the vectors of Nose, Right-Eye, and Nose, Right-Mouth \\$\theta_2$ $\gets$ $\arccos((\overrightarrow{NRE}\cdot\overrightarrow{NRM}))$.
         \STATE Compute absolute difference between the angles, as \emph{angleDiff} $\gets$ $\lvert{\theta_1-\theta_2}\rvert$
          \IF {\emph{angleDiff} $\leq$ $\tau$}
            \STATE Face is Frontal \RETURN{} True
        \ENDIF
       \RETURN{}  False
      \end{algorithmic}
\end{algorithm}

\subsection{Optimal Face Pair Generation for composite image generation}\label{OptimalSec}
\begin{algorithm}[h!]
\caption{Optimal Pair Finding Algorithm}\label{Alg:PairFindingAlgo} 
\KwInput{Random Image Pairs ($I_1^1,I_2^1$),$\cdots$,($I_1^N,I_2^N$)}
\KwOutput{Optimal Image Pairs ($O_1^1$,$O_2^1$),$\cdots$,($O_1^N$,$O_2^N$)}
     \begin{algorithmic}[1]
         \STATE Compute Arcface features on the input face images.
          \STATE{Optimal-Pair $\gets$ $\left[ \right]$}
         \FOR{$i \gets 1$ to $N$}
        \STATE Compute Index of nearest arcface feature $j$  to $i$
        \IF {$\left(j,i\right)$ $\notin$  Optimal-Pair}
            \STATE Append $\left(i,j\right)$ to Optimal-Pair
        \ELSE
        \STATE Compute Index of second-nearest arcface feature $k$  to $i$
        \STATE Append $\left(i,k\right)$ to Optimal-Pair
        \ENDIF
         \ENDFOR
         \RETURN{} Optimal-Pair
      \end{algorithmic}
\end{algorithm}
It is essential to select the look-alike data subjects to achieve the optimal attack potential with the proposed composite face image generation. We choose the optimal pairs to generate the composite face images to this extent. Given $n$ synthetic samples, the total number of pairs possible is $((n){\times}(n-1))/2$, and thus finding optimal pairs using this approach is quadratic ($O(n^2)$) as we have to compute the pair-wise distance for all pairs. The quadratic time for pair-finding is within the computing limits as our dataset now consists of 1000 unique data subjects. We have put an additional constraint in the pair-finding algorithm not to return swapped pairs, i.e., if ($i$,$j$) is the list, then ($j$,$i$) is not added to the optimal pair list. The approach for optimal pair finding is summarized in an algorithmic format in Algorithm~\ref{Alg:PairFindingAlgo} and a few optimal pairs are shown in Figure~\ref{fig:blockDiagram1}. The distance metric used in our approach is cosine-distance from Arcface~\cite{Deng19-Arcface-CVPR} features.

Thus, the CFIA dataset has 1000 unique identities with 2000 bona fide samples and 526000 CFIA samples. The whole dataset will made  publicly available for research purposes along with code at the following link \url{https://github.com/jagmohaniiit/LatentCompositionCode}.

\section{Vulnerability Analysis}
\label{sec:VulFRS}

\begin{table*}[h!]
    \centering
      \resizebox{0.8\linewidth}{!}{
      \scriptsize
      \begin{tabular}{|c|c|c|c|c|} 
    \hline
    {\bf{Utility Features}} & {\bf{MMPMR \cite{Ulrich17_MMPMR_Biosig}}} & {\bf{FMMPMR \cite{Sushma20_FMMPMR_IJCB}}} & {\bf{MAP \cite{Ferrara22_MAP_IWBF}}} & {\bf{G-MAP}}
    \\ \hline
    {Multiple Attempts for individual morphing Image} &\cmark  & \cmark &  \xmark & \cmark \\ \hline
    {Pairwise comparison of probe samples} &\xmark  & \cmark & \cmark  & \cmark \\ \hline
    {Multiple FRS} &\xmark  & \xmark & \cmark  & \cmark \\ \hline
    {Multiple Morphing Types} &\xmark  & \xmark & \xmark & \cmark \\ \hline
    {Accountability for FTAR}&\xmark  & \xmark & \xmark & \cmark \\ \hline
    {Vulnerability as a single number} & \cmark & \cmark & \xmark & \cmark \\ \hline
    \end{tabular}
    }\caption{Utility Features of existing and proposed vulnerability metrics}
    \label{table:mapTable}
    \end{table*}

This section presents the vulnerability analysis of the proposed CFIA samples on the automatic FRS. We have benchmarked four different FRS based on deep learning. The deep learning FRS employed in this work are Arcface~\cite{ArcfaceGithub} (Model R100 V1), VGGFace~\cite{VGGFaceGithub} (Version 2), Facenet~\cite{FacenetGithub} and Magface~\cite{MagfaceGithub}. 
{{
The proposed CFIA samples are generated based on the face images corresponding to two contributory subjects. Therefore, we benchmark the attack potential of CFIA by comparing the FRS scores computed from both contributory subjects against the pre-set threshold of FAR = 0.1\%. The comparison scores from FRS are computed by enrolling the attack samples to FRS and then probing the face images from the contributory subjects. \\
In the literature, the vulnerability of FRS can be calculated using three different types of metrics namely: Mated Morphed Presentation Match Rate (MMPMR) \cite{Ulrich17_MMPMR_Biosig}, Fully Mated Morphed Presentation Match Rate (FMMPMR) \cite{Sushma20_FMMPMR_IJCB} and Morphing Attack Potential (MAP) \cite{Ferrara22_MAP_IWBF}. The MMPMR metric is based on the independent attempts, while FMMPMR employs pair-wise probe attempts of the contributory subjects. The MAP metric improves existing metrics by accommodating multiple FRS together with pair-wise probe attempts. However, the MAP metric will represent the vulnerability results in the matrix form as  attempts versus multiple FRS. Hence, MAP does not quantify the vulnerability as a single  number. Further, the constant number of attempts will also limit the evaluation as it enforces all enrolled attack samples to have the same number of attempts which is not true in a real-life scenario. Additionally, while computing the vulnerability, the existing metrics do not consider accommodating Failure-to-Acquire Rate (FTAR) and multiple morphing generation techniques. Even though the enroled face image (attack/CFIA/morphing or bona fide) is captured in the constrained conditions, the probe images are not essentially captured in the constrained conditions due to the nature of ID verification scenarios (for example, in border control gates, smartphone authentication, etc.). Further, the availability of different types of morphing (or attack) generation techniques (full face/partial face/facial attribute) allows an attacker to generate various attack samples. Hence, the vulnerability computation needs to accommodate different types of morphing generation.  These factors motivated us to enhance the existing vulnerability metrics (MAP) to include more utility features such as (a) Dynamic attempts per morph image, (b) Accountability for FTAR, (3) Accountability for multiple morphing techniques, and (4) Single numeric value indicating the vulnerability. The enhanced vulnerability metric is termed as Generalised Morphing Attack Potential (G-MAP). Table \ref{table:mapTable}  presents utility features of the proposed G-MAP compared to existing metrics such as MMPMR \cite{Ulrich17_MMPMR_Biosig}, FMMPMR \cite{Sushma20_FMMPMR_IJCB} and MAP\cite{Ferrara22_MAP_IWBF}.} }

\subsection{\bf{Mathematical Formulation of G-MAP}}
{{
Let $\mathbb{P}$ denote the set of paired probe images (which can also be denoted as number of attempts), $\mathbb{F}$ denote the set of FRS, $\mathbb{D}$ denote the set of Morphing Attack Generation Type, $\mathbb{M}_d$ denote the face morphing image set corresponding to Morphing Attack Generation Type $d$, $\tau_l$ indicate the similarity score threshold for FRS ($l$), and $||$ represents the count of elements in a set during metric evaluation. The G-MAP metric is presented as below: }

\begin{equation}\label{eqn:G-MAP}
\begin{aligned}
&{\textrm{G-MAP}} ={\frac{1}{|\mathbb{D}|}}{\sum_{d}^{|\mathbb{D}|}}{\frac{1}{|\mathbb{P}|}}{\frac{1}{|\mathbb{M}_d|}} \min_{{l}}\\
& \sum_{i,j}^{|\mathbb{P}|,|\mathbb{M}_d|}\bigg\{\left[ (S1_i^j > \tau_l) \wedge \cdots ( Sk_i^j > \tau_l) \right]\\ &{\times}  \left[ (1-FTAR(i,l)) \right] \bigg\}\\
\end{aligned}
\end{equation}
where,  $FTAR(i,l)$ is the failure to acquire probe image in attempt $i$ using  FRS ($l$). The algorithm for G-MAP is presented in \ref{Alg:G-MAP} and the code is made available in the link~\cite{GMAPGithub}. 

\begin{algorithm}[h!]
\caption{Generalized Morph Attack Potential (G-MAP)}\label{Alg:G-MAP} 
\KwInput{Set of Probe Images $\mathbb{P}$, Set of FRS $\mathbb{F}$, Set of Morphing Attack Generation Type $\mathbb{D}$, Set of Morphing Attack Images in $d^{\textrm{th}}$ attack $\mathbb{M}_d$, $\tau_{l}$ indicate the similarity score threshold for FRS.}
\KwOutput{G-MAP}
     \begin{algorithmic}[1]
      \STATE Compute G-MAP Metric as follows.
          \FOR{$j \gets 1$ to $|\mathbb{M}_d|$}
         \FOR{$d \gets 1$ to $|\mathbb{D}|$}
         \FOR{$l \gets 1$ to $|\mathbb{F}|$}
         \FOR{$i \gets 1$ to $|\mathbb{P}|$}
         \STATE Compute QF(i,l)=(1-FTAR(i,l))
        \STATE Compute $\textrm{G-MAP}$(d)=${\frac{1}{|\mathbb{P}|}}{\frac{1}{|\mathbb{M}_d|}}\min_{l}\sum_{i,j}^{|\mathbb{P}|,|\mathbb{M}_d|} (S1_i^j > \tau_l) \wedge \cdots $\\$( Sk_i^j > \tau_l){\times}QF(i,l)$
        \ENDFOR{}
        \ENDFOR{}
         \ENDFOR{}
           \ENDFOR{}
           \STATE Compute G-MAP= ${\frac{1}{|\mathbb{D}|}}  
           {G-MAP(d)}$
        \RETURN{G-MAP}
      \end{algorithmic}
\end{algorithm}

\subsection{\bf{Computing G-MAP}}
Given the fact that G-MAP can be computed with different parameters, which include multiple probe attempts, multiple FRS and the morph attack generation types. {\bf{G-MAP with multiple probe attempts}} is calculated from Equation \ref{eqn:G-MAP} by setting D = 1 and F = 1 where the similarity scores ($S1_i^j$)  should be greater than threshold ($\tau_l$) and FTAR(i,l) is calculated for each probe attempt and FRS. Thus, making {\bf{G-MAP with multiple probe attempts}} identical to FMMPMR when FTAR=0. Further, {\bf{G-MAP with Multiple FRS and multiple probe attempts}} is computed by taking minimum across FRS and using D=1. Finally, the full {\bf{G-MAP metric}} would provide a single value indicating the vulnerability which is by taking the average as shown in Equation  \ref{eqn:G-MAP}. 

\subsection{Quantitative evaluation of vulnerability}
\begin{figure*}
     \centering
     \begin{subfigure}[b]{0.40\textwidth}
         \centering
         \includegraphics[width=\textwidth]{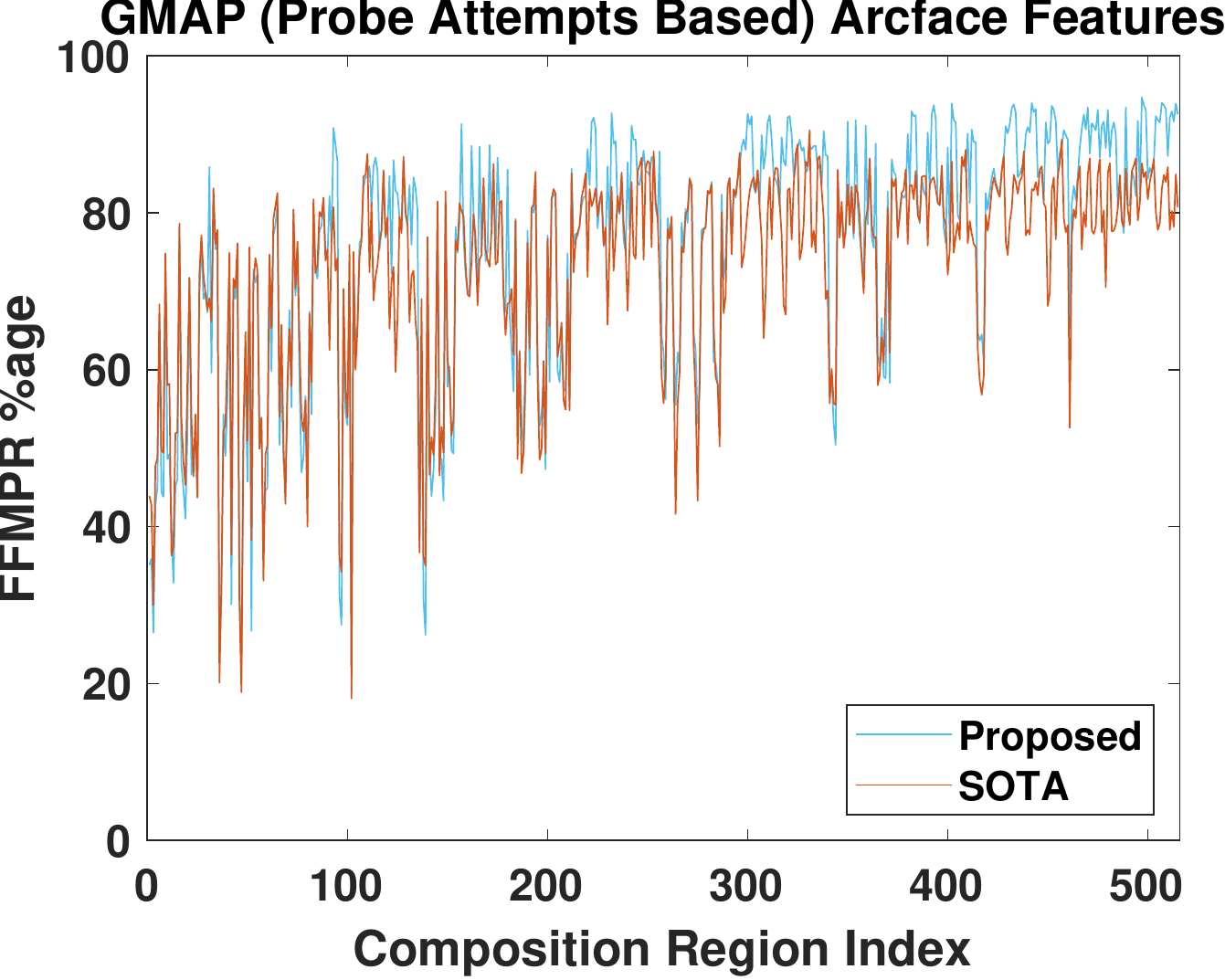}
         \caption{}
         \label{fig:y equals x}
     \end{subfigure}
     \begin{subfigure}[b]{0.40\textwidth}
         \centering
         \includegraphics[width=\textwidth]{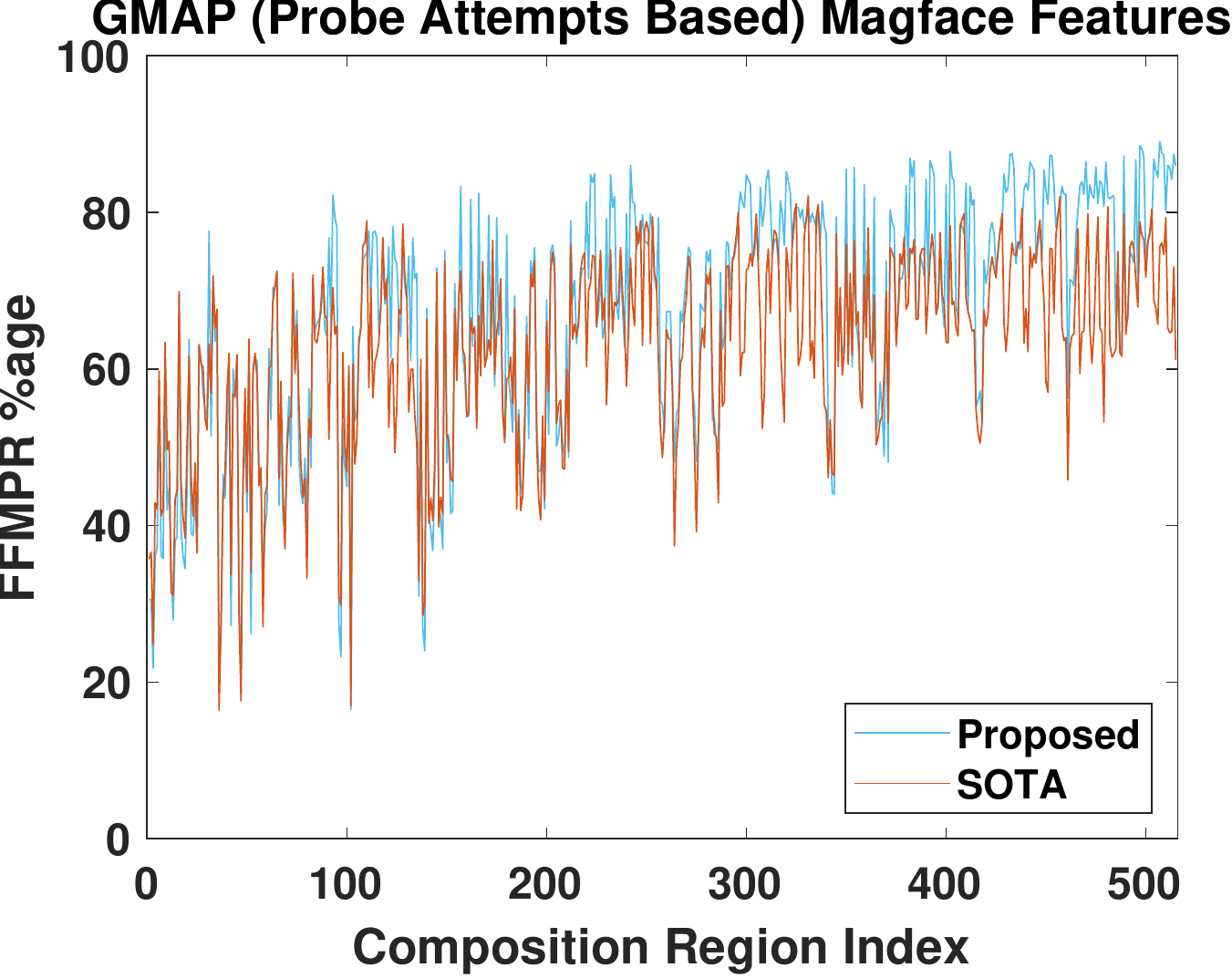}
         \caption{}
         \label{fig:three sin x}
     \end{subfigure}
     \\
     \begin{subfigure}[b]{0.40\textwidth}
         \centering
         \includegraphics[width=\textwidth]{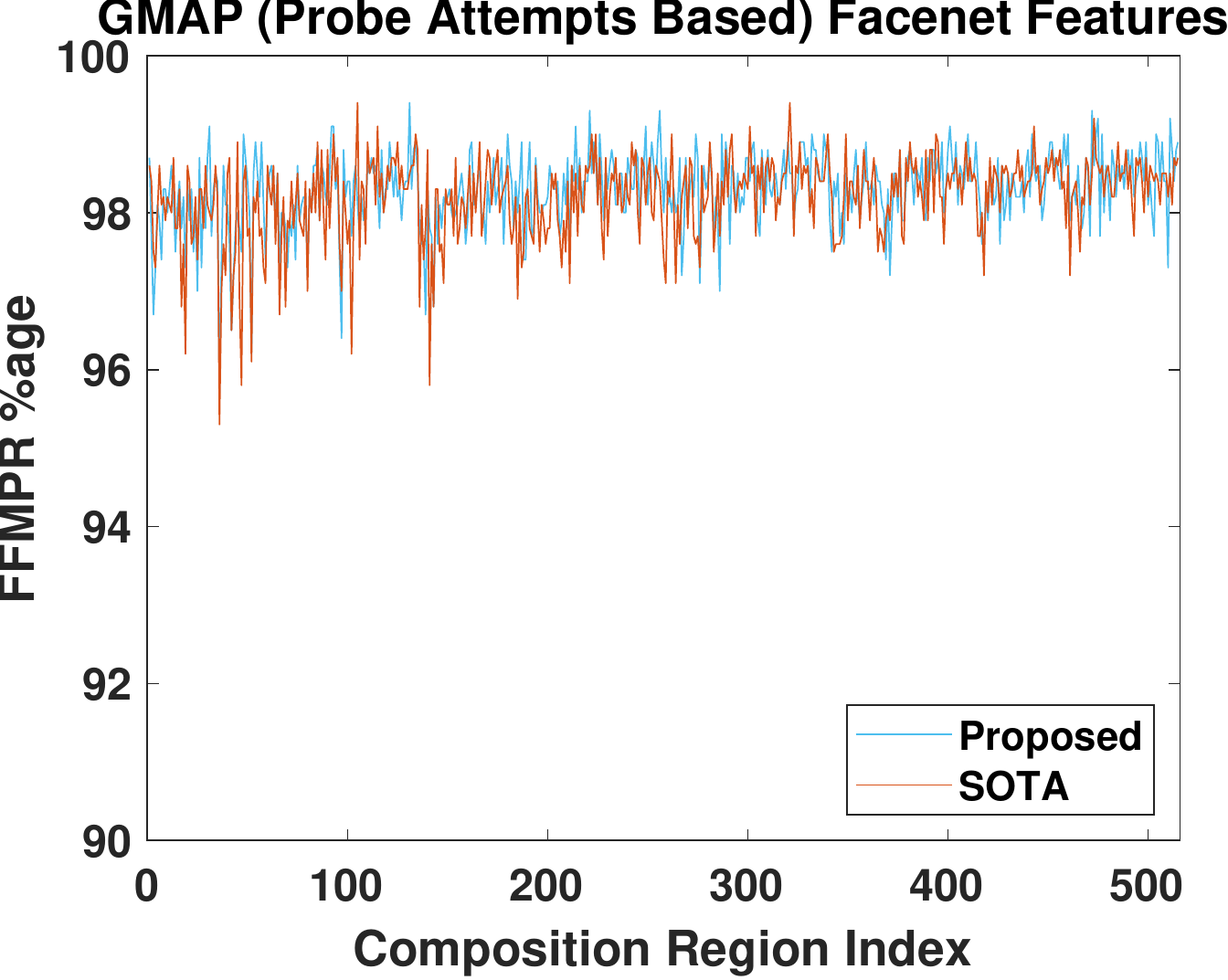}
         \caption{}
         \label{fig:five over x}
     \end{subfigure}
     \begin{subfigure}[b]{0.40\textwidth}
         \centering
         \includegraphics[width=\textwidth]{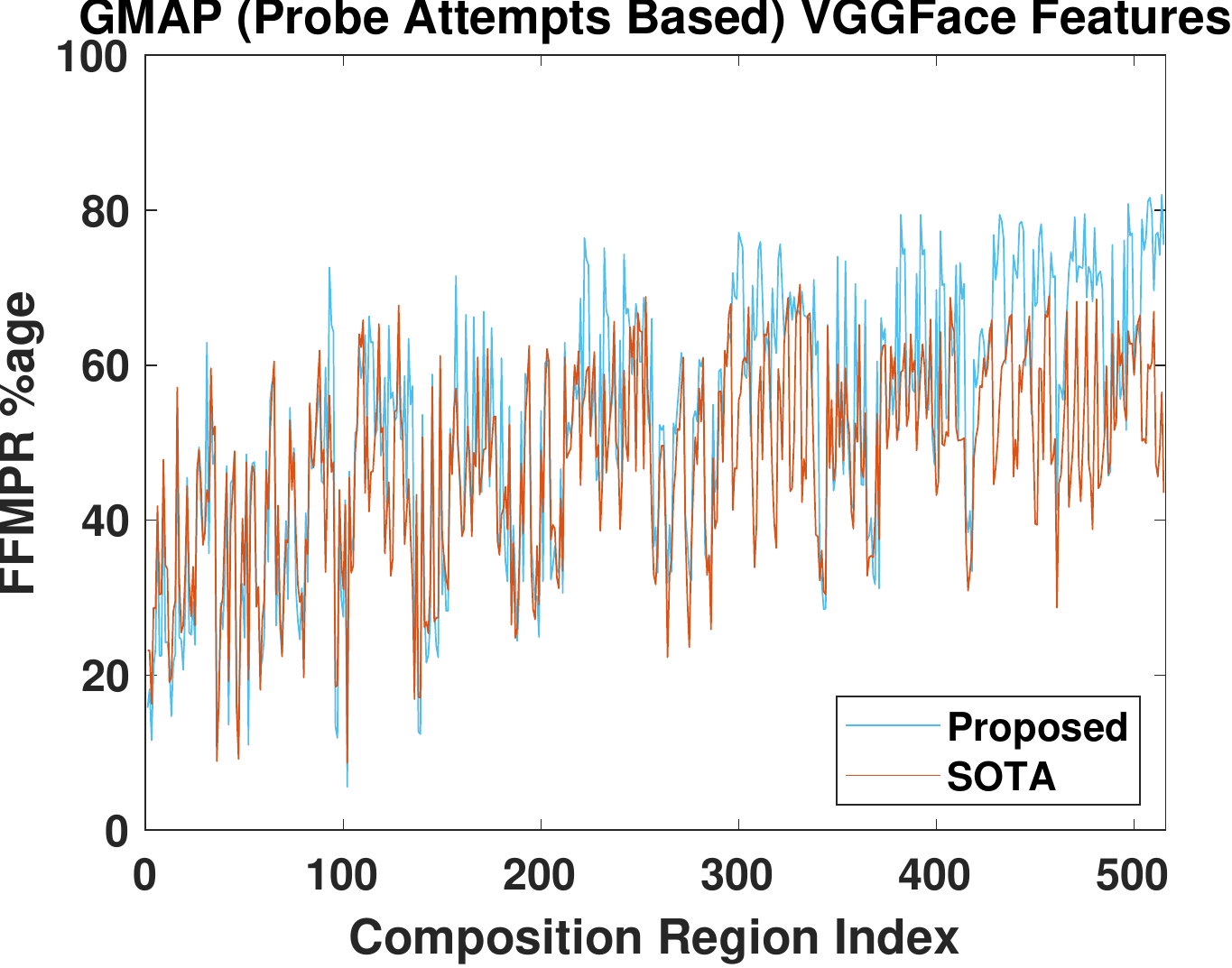}
         \caption{}
         \label{fig:five over x}
     \end{subfigure}
        \caption{Vulnerability Plots G-MAP (Probe Attempts). X-axis indicates the number of unique CFIA generated where the index 0 corresponds to E-H, the index 1 corresponds to H-E, and the following indices in the left to right order corresponding to Table~\ref {table:overallSegmentsExhaustive}. Thus, finally, index 525 to HBSENM-HBSENM.}
        \label{fig:G-MAPPlots1}
\end{figure*}

\begin{figure}[htp!]
\centering
\includegraphics[width=0.9999\linewidth]
{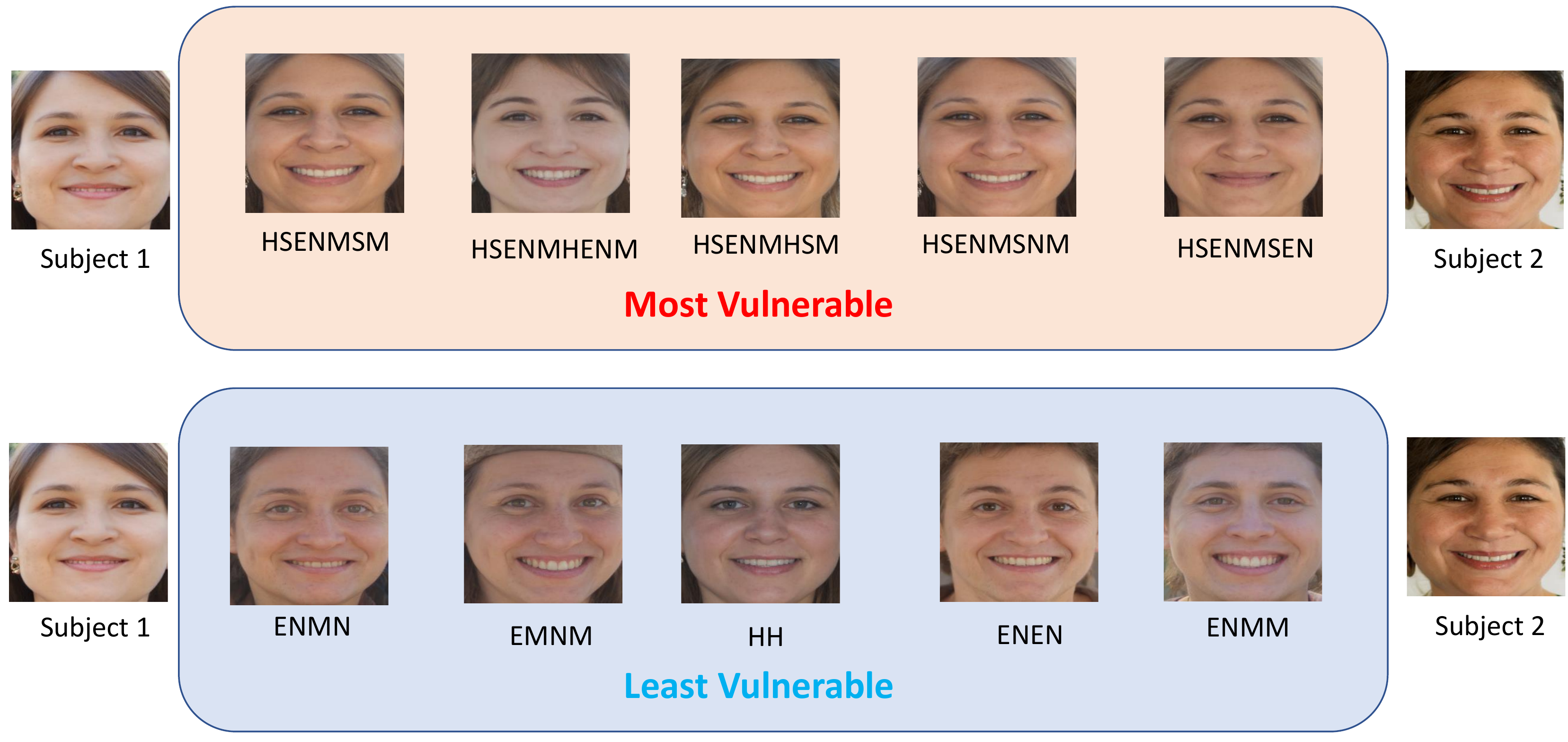}
\caption{Most and Least Vulnerable CFIA Samples from the dataset.}
\label{fig:mostLeastVulnerable}
\end{figure}

\begin{figure}[htp!]
\centering
\includegraphics[width=0.95\linewidth]
{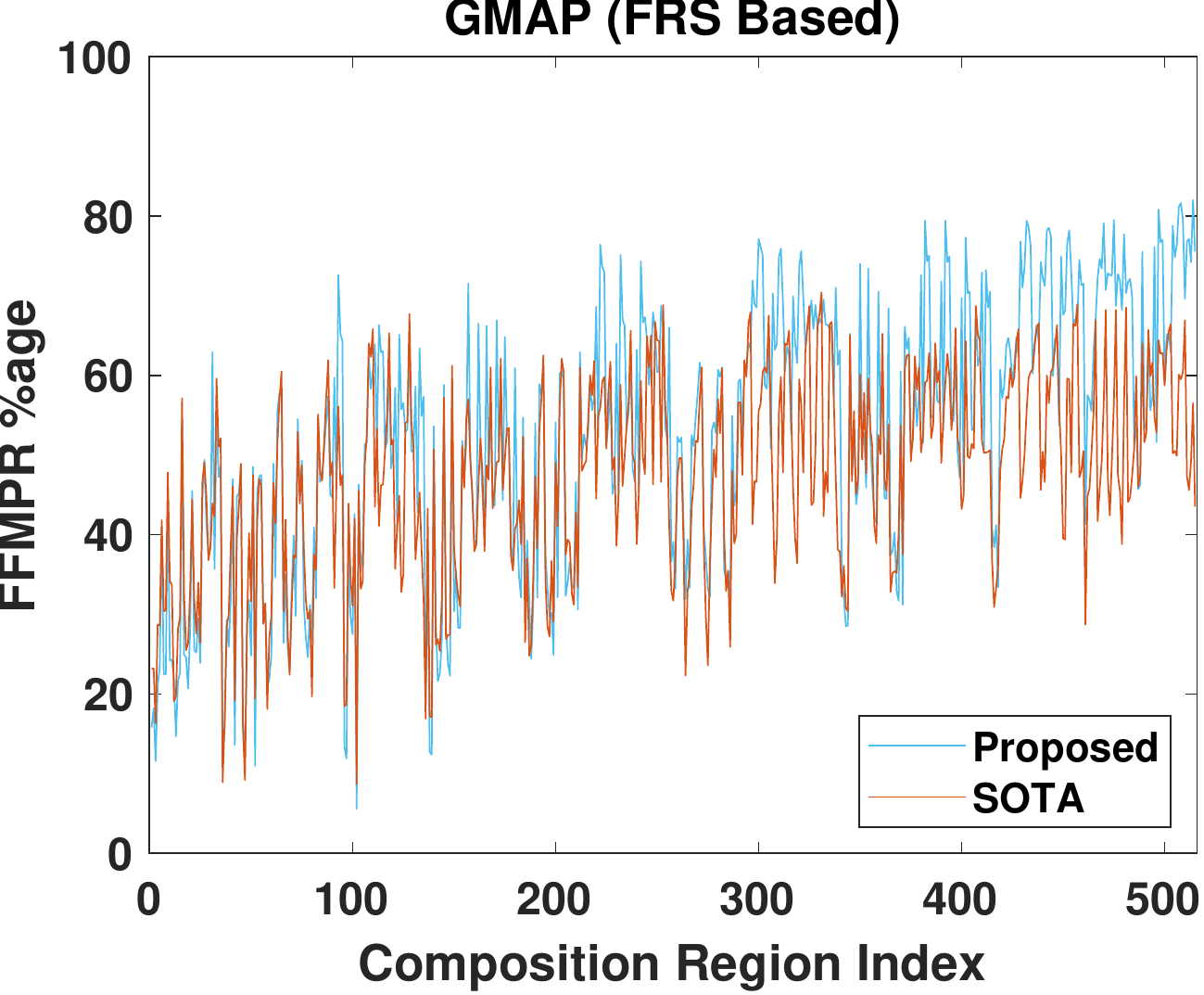}
\caption{Vulnerability Plots G-MAP (Multiple  FRS and multiple probe attempts-based). X-axis indicates the number of unique CFIA generated where the index 0 corresponds to E-H, the index 1 corresponds to H-E, and the following indices in the left to right order corresponding to Table~\ref {table:overallSegmentsExhaustive}. Thus, finally, index 525 to HBSENM-HBSENM.}
\label{fig:G-MAPPlots}
\end{figure}
{{ In this section, we present the qualitative and quantitative evaluation of the vulnerability corresponding to FRS for all 526 CFIA samples generated using a different combination of facial attributes. Since G-MAP is a function of attempts, FRS, and morphing types, this will allow one to analyse the quantitative results corresponding to (a) probe attempts independently to FRS and attack image generation type (b) Multiple FRS with multiple attempts independent of attack image generation type (c) Final G-MAP value as a function of attempts, multiple FRS and different types of attack image generation together with FTAR.\\
In this work, we first present the vulnerability of the full CFIA dataset using four different FRS such as Arcface~\cite{ArcfaceGithub} (Model R100 V1), VGGFace~\cite{VGGFaceGithub} (Version 2), Facenet~\cite{FacenetGithub} and Magface~\cite{MagfaceGithub}. The vulnerability reported in this work is computed by setting the threshold of FRS at FAR = 0.1\%. Figure \ref{fig:G-MAPPlots1} shows the plot of G-MAP values that are computed for multiple probe attempts independent of FRS and CFIA generation type. The composite region index started from the output segment with two regions (left extreme of x-axis in Figure \ref{fig:G-MAPPlots1}) and continued till six combinations (right extreme of x-axis in Figure \ref{fig:G-MAPPlots1}). Table \ref{table:vulnerabilityMAP1} shows the quantitative values of G-MAP with probe attempts corresponding to four different FRS. For simplicity, we have only indicated the quantitative results to 14 combinations sampled from 526 regions. \textit{It needs to be pointed out that these 14 regions are indicative of least, moderate and most vulnerable regions from 526 unique CFIA combinations.}   
}}

{{Based on the obtained results following are the main observations: 
\begin{itemize}
    \item  The number of composite regions used to generate the CFIA samples plays a vital role in the vulnerability of FRS. Using a smaller number of regions (for example, 2, 3 and 4) to generate the CFIA will result in a lower vulnerability of FRS. This it can be attributed to the fact that in these regions, the blending for the generation of composite happens in a small region and the remainder of the face is generated by GAN-based image inpainting. For example, if we consider the two regions (or facial attribute) CFIA generation, then one region is taken from the contributory subject 1 and another region is taken from the contributory subject 2, from these selected regions, the whole face is generated using the GAN. This process results in the loss of identity information in the generated CFIA due to the availability of a few regions. Figure \ref{fig:mostLeastVulnerable} illustrates the example of low vulnerable CFIA samples generated using two and three region combinations. The lower vulnerability is noted with both SOTA and the proposed CFIA generation. 
\item  The CFIA samples generated using 4, 5 and 6 regions have indicated higher vulnerability of FRS. This can be attributed to the fact that the larger the number of facial attributes used from both the contributory data subjects, the higher the vulnerability of the FRS. This trend is noticed equally with both SOTA and the proposed CFIA generation. Figure \ref{fig:mostLeastVulnerable} shows the  CFIA samples for the top 5 highest vulnerable combinations indicating the rich identity features corresponding to both contributory subjects. 
\item   Among the four different FRS employed in this work, the Facenet \cite{FacenetGithub} indicates the higher vulnerability across different region combinations. The lowest vulnerability is noted with the VGG FRS \cite{VGGFaceGithub}. 
\item  The proposed CFIA generation technique indicates the higher vulnerability of FRS when compared with the SOTA \cite{Chai21LatentCompositeICLR}. The higher vulnerability of FRS to the proposed technique is noted with the CFIA samples that are generated using five and six-region combinations.
\item Additional experiments on Commercial-Off-The-Shelf (COTS) to indicate the importance of FTAR is included in the Appendix A. 
\end{itemize}
}}

\begin{table*}[htp]
\centering
 \resizebox{1.0\linewidth}{!}{
\begin{tabular}{|c|c|c|c|c|c|c|c|c|c|c|c|c|c|c|c|} 
\hline
\multicolumn{16}{|c|}{{\bf{ G-MAP \% (Multiple probe attempts)}}} \\ \hline
{{\bf{FRS}}} & {{\bf{Method}}} & {\bf{R1}} & {\bf{R2}} & {\bf{R3}} & {\bf{R4}} & {\bf{R5}} & {\bf{R6}} & {\bf{R7}} & {\bf{R8}} & {\bf{R9}} & {\bf{R10}} & {\bf{R11}} & {\bf{R12}} & {\bf{R13}} & {\bf{R14}} \\ \hline
\multirow{2}{*}{{\bf{Arcface (FAR=0.1\%)}}} & {\bf{SOTA~\cite{Chai21LatentCompositeICLR}}} &70.5 & 58.7 & 60.6 & 52.7 & 70.8 & 69.2 & 72.9 & 71.6 & 69.1 & 69.1 & 67.6 & 74.1 & 69.6 & 72.3 \\ \cline{2-16}  
 & {\bf{Proposed}} &67.3 & 58.1 & 60.2 & 72.4 & 70.4 & 68.4 & 72.5 & 71.9 & 71.4 & 84.2 & 82.8 & 76.4 & 86.8 & {\bf{89.9}} \\ \hline 
\multirow{2}{*}{{\bf{MagFace (FAR=0.1\%)}}} & {\bf{SOTA~\cite{Chai21LatentCompositeICLR}}} & 57.6 & 45.0 & 48.7 & 42.7 & 58.0 & 57.3 & 61.0 & 60.7 & 61.1 & 59.0 & 52.6 & 65.1 & 57.7 & 54.0 \\ \cline{2-16}  
 & {\bf{Proposed}} & 67.3 & 58.2 & 60.1 & 72.4 & 70.4 & 68.6 & 72.5 & 72.0 & 71.4 & 84.2 & 82.8 & 76.4 & 86.7 & {\bf{89.8}}\\ \hline 
\multirow{2}{*}{{\bf{VGGFace (FAR=0.1\%)}}}  & {\bf{SOTA~\cite{Chai21LatentCompositeICLR}}} &65.2 & 64.2 & 62.7 & 63.6 & 64.9 & 63.9 & 66.1 & 65.7 & 67.0 & 67.2 & 65.4 & 65.9 & 68.1 & 67.7 \\ \cline{2-16} 
 & {\bf{Proposed}} &65.4 & 64.4 & 63.0 & 65.9 & 66.4 & 68.1 & 69.1 & 66.0 & 68.5 & 70.5 & 70.5 & 68.6 & 71.0 & {\bf{71.9}} \\ \hline 
\multirow{2}{*}{{\bf{Facenet (FAR=0.1\%)}}} & {\bf{SOTA~\cite{Chai21LatentCompositeICLR}}} &95.8 & 96.9 & 95.4 & 93.8 & 95.3 & 96.5 & 95.9 & 95.8 & 96.1 & 95.6 & 94.5 & 96.4 & 94.7 & 95.2 \\ \cline{2-16} 
& {\bf{Proposed}} &96.1 & 97.7 & 96.3 & 97.4 & 95.5 & 97.3 & 95.4 & 96.4 & 97.4 & 97.2 & 97.3 & 96.6 & 96.6 & {\bf{97.0}} \\ \hline 
\end{tabular}}
    \caption{ Vulnerability analysis using the G-MAP metric (probe attempts-based) for the proposed method and the SOTA~\cite{Chai21LatentCompositeICLR}, where the description of regions is provided in Table~\ref{table:overallSegmentsExhaustive}. Where R1 is (S-E), R2 is (S-N), R3 is (S-M), R4 is (S-S), R5 is (SEN-M), R6 is (SEM-N), R7 is (SNM-E), R8 is (SEN-EM), R9 is (SEN-EN), R10 is (SEN-SEM), R11 is (SEN-SEN), R12 is (SENM-ENM), R13 is (SENM-SENM), and R14 is (HBSENM-HBSENM)).}
    \label{table:vulnerabilityMAP1}
\end{table*}

\begin{table*}[htp]
\centering
 \resizebox{1.0\linewidth}{!}{
\begin{tabular}{|c|c|c|c|c|c|c|c|c|c|c|c|c|c|c|} 
\hline
\multicolumn{15}{|c|}{{\bf{ G-MAP \% (Multiple  FRS and multiple probe attempts)}}} \\ \hline
{{\bf{Method}}} & {\bf{R1}} & {\bf{R2}} & {\bf{R3}} & {\bf{R4}} & {\bf{R5}} & {\bf{R6}} & {\bf{R7}} & {\bf{R8}} & {\bf{R9}} & {\bf{R10}} & {\bf{R11}} & {\bf{R12}} & {\bf{R13}} & {\bf{R14}} \\ \hline

{\bf{SOTA~\cite{Chai21LatentCompositeICLR}}} & 57.6& 45.0& 48.7& 42.7& 58.0& 57.3& 61.0& 60.7& 61.1& 59.0& 52.6& 65.1& 57.7& 54.0 \\ \hline
{\bf{Proposed}} & 65.4& 58.1& 60.1& 65.9& 66.4& 68.1& 69.1& 66.0& 68.5& 70.5& 70.5& 68.6& 71.0& 71.9\\ \hline

\end{tabular}}
    \caption{ Vulnerability analysis using the G-MAP metric (Multiple  FRS and multiple probe attempts-based) for the proposed method and the SOTA~\cite{Chai21LatentCompositeICLR}.}
    \label{table:vulnerabilityMAP2}
\end{table*}

{{ Figure \ref{fig:G-MAPPlots} shows the vulnerability of FRS with G-MAP computed across multiple FRS and multiple attempts for both SOTA and proposed CFIA with 526 combinations. Given CFIA sample is said to be vulnerable if the multiple probe attempts must successfully deceive the multiple FRS. Thus, the G-MAP will provide a single value indicating the vulnerability by taking the average probe attempts while accounting for FTAR. Table \ref{table:vulnerabilityMAP2} indicates the G-MAP (multiple FRS and multiple probes) for 14 different regions (that are the same as Table \ref{table:vulnerabilityMAP1}) for simplicity. Based on the obtained results\\ following are the main observations: 
\begin{itemize}
    \item The CFIA samples generated with five and six regions combinations indicate higher vulnerability of multiple FRS. This is noted with both SOTA and the proposed CFIA technique. 
\item The proposed CFIA samples indicate the higher vulnerability of FRS compared to SOTA. 
\item Figure \ref{fig:G-MAPCombinations} shows the box plots of proposed method and SOTA  computed across CFIA region index as mentioned in Table \ref{table:overallSegmentsExhaustive} indicates the mean and variance computed by taking the average of G-MAP values computed over all region combinations within the CFIA region index. As noticed from Figure \ref{fig:G-MAPCombinations}  and Table \ref{table:G-MAPBoxPlot}, the combinations with less number of regions do not significantly increase the vulnerability. The combination of five regions with CFIA region index of 13, 14 and  15  indicates the higher vulnerability of FRS with the proposed CFIA technique. 
\end{itemize}
}}

\begin{table}[htp]
    \centering
      \resizebox{0.8\linewidth}{!}{
      \scriptsize
    \begin{tabular}{|c|c|} 
    \hline
\multicolumn{2}{|c|}{{\bf{ G-MAP \% }}} \\ \hline
 {\bf{SOTA Method~\cite{Chai21LatentCompositeICLR}}} & {\bf{Proposed Method}} \\ \hline
 46.9\% & 52.4\% \\ \hline
\end{tabular}}
    \caption{G-MAP for SOTA Method and the Proposed Method computed using 526 CFIA compositions.}
    \label{table:vulnerabilityMAP3}
\end{table}

\begin{figure*}[th!]
\centering
\begin{subfigure}[b]{0.45\textwidth}
         \centering
         \includegraphics[width=\textwidth]{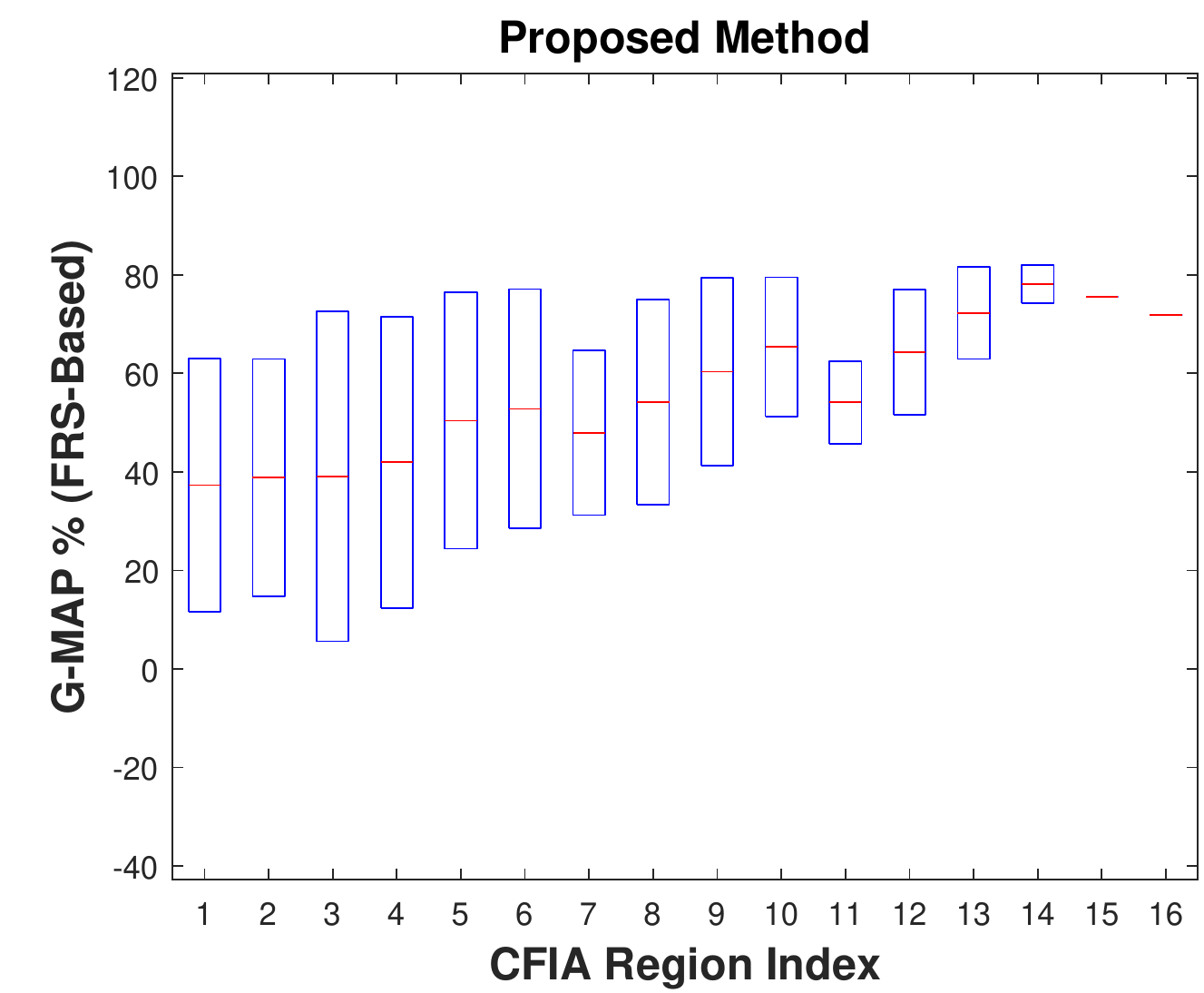}
         \caption{}
         \label{fig:y equals x}
     \end{subfigure}
      \hfill
     \begin{subfigure}[b]{0.45\textwidth}
         \centering
         \includegraphics[width=\textwidth]{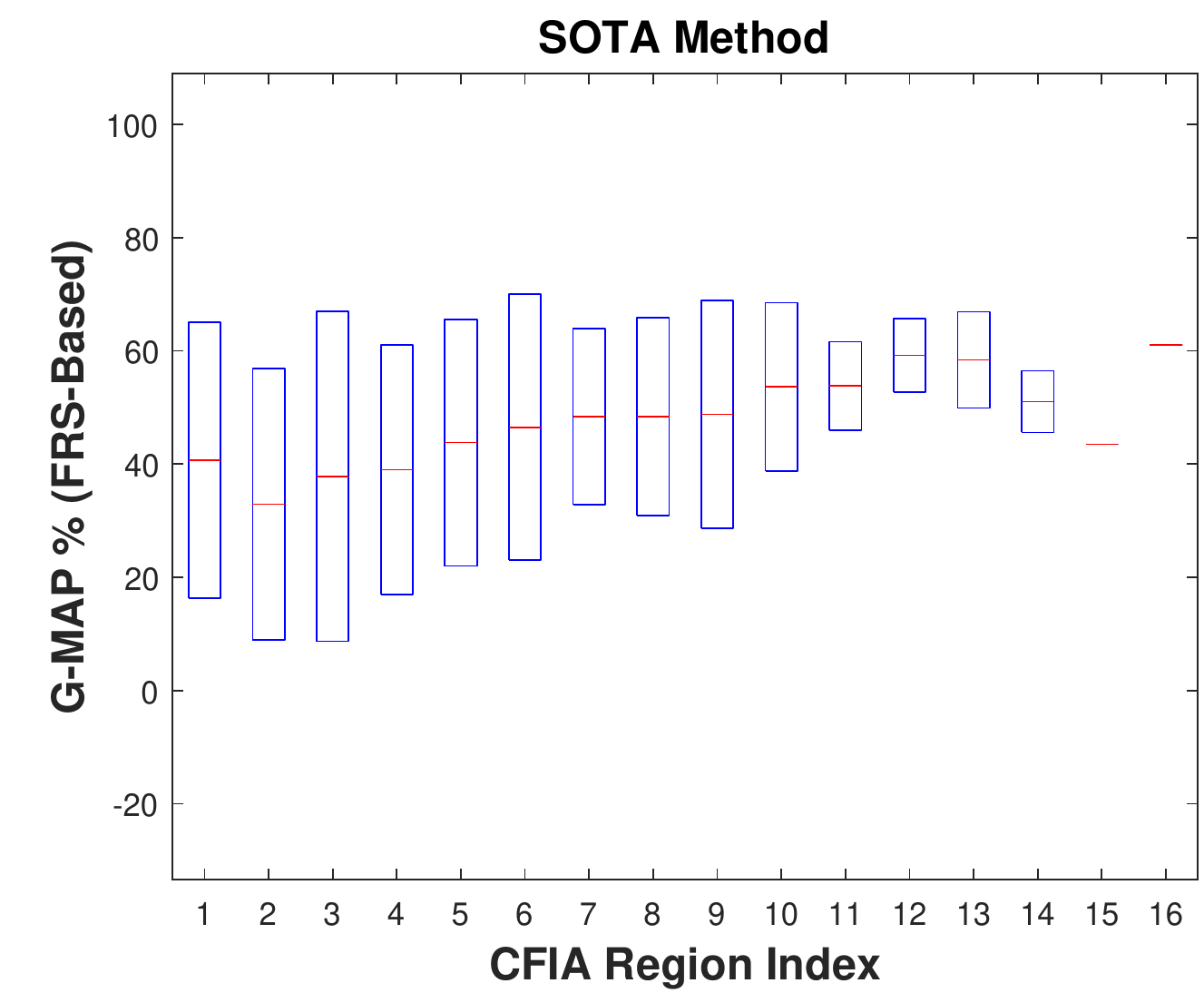}
         \caption{}
         \label{fig:three sin x}
     \end{subfigure}
\caption{G-MAP Combinations (FRS-Based) where the number denotes the CFIA Region Index of (a) Proposed and (b) SOTA Method ~\cite{Chai21LatentCompositeICLR}  (Table~\ref{table:overallSegmentsExhaustive})}
\label{fig:G-MAPCombinations}
\end{figure*}

\begin{table}[htp]
\centering
 \resizebox{1.0\linewidth}{!}{
\begin{tabular}{|c|c|c|}
\hline
{\bf{CFIA Region Index}} &  {{\bf{Proposed Method}}} & {{\bf{SOTA Method~\cite{Chai21LatentCompositeICLR}}}} \\ \hline
1 & 37.3$\pm$36.3 & 40.7$\pm$34.5 \\ \hline
2 & 38.8$\pm$34.0 & 32.9$\pm$33.9 \\ \hline
3 & 39.1$\pm$47.3 & 37.8$\pm$41.2  \\ \hline
4 & 41.9$\pm$41.7 & 39$\pm$31.1  \\ \hline
5 & 50.4$\pm$36.7 & 43.8$\pm$30.8 \\ \hline
6 & 52.8$\pm$34.3 & 46.5$\pm$33.2 \\ \hline
7 & 47.9$\pm$23.6 & 48.4$\pm$22.0 \\ \hline
8 & 54.2$\pm$29.4 & 48.4$\pm$24.7 \\ \hline
9 & 60.3$\pm$26.9 & 48.8$\pm$28.4 \\ \hline
10 & 65.3$\pm$20.0 & 53.6$\pm$21.0 \\ \hline
11 & 54.1$\pm$11.8 & 53.8$\pm$11.0 \\ \hline
12 & 64.3$\pm$17.9 & 59.2$\pm$9.1 \\ \hline
13 & 72.2$\pm$13.2 & 58.4$\pm$12.0 \\ \hline
14 & 78.1$\pm$5.5  & 51.0$\pm$7.7 \\ \hline
15 & 75.5$\pm$0 & 43.5$\pm$0 \\ \hline
16 & 71.8$\pm$0 &  61.1$\pm$0 \\ \hline
\end{tabular}}
    \caption{Table showing mean and standard deviation for each CFIA region index based on SOTA~\cite{Chai21LatentCompositeICLR} and the Proposed Method.
    (for CFIA region index please refer Table~\ref{table:overallSegmentsExhaustive})}
    \label{table:G-MAPBoxPlot}
\end{table}

{{
Table \ref{table:vulnerabilityMAP3} indicates the vulnerability computed with full capacity of G-MAP in which multiple attempts, multiple FRS, multiple attack types and FTAR. The G-MAP values indicated in the \ref{table:vulnerabilityMAP3} quantify the vulnerability of the proposed and SOTA for the complete CFIA dataset with 526 attack types and four different FRS. The obtained results indicate that the proposed method gives higher bounds of vulnerability for all 526 attack types. 
}}
\section{Perceptual quality evaluation of the composite images} 
\label{sec:pecpQuality}
\begin{table}[h!]
\centering
\resizebox{0.8\linewidth}{!}{
\begin{tabular}{|c|c|c|c|c|c|c|}
\hline
{\bf{Region}} &  \multicolumn{2}{|c|}{\bf{PSNR}}  & \multicolumn{2}{|c|}{\bf{SSIM}} \\ \hline 
{\bf{R1}} & {\bf{SOTA~\cite{Chai21LatentCompositeICLR}}} & {\bf{Proposed}} & {\bf{SOTA~\cite{Chai21LatentCompositeICLR}}} & {\bf{Proposed}}  \\ \hline 
{\bf{R1}}  & 15.4$\pm$10.2 & 15.6$\pm$7.0 & 0.68$\pm$0.01 & 0.71$\pm$0.00  \\ \hline 
{\bf{R2}}  & 15.5$\pm$9.5 & 15.7$\pm$6.6 & 0.68$\pm$0.01 & 0.71$\pm$0.00  \\ \hline 
{\bf{R3}}  & 15.5$\pm$10.2 & 15.6$\pm$7.0 & 0.68$\pm$0.01 & 0.71$\pm$0.00  \\ \hline 
{\bf{R4}}  & 15.6$\pm$8.6 & 15.9$\pm$4.6 & 0.69$\pm$0.01 & 0.71$\pm$0.00  \\ \hline 
{\bf{R5}}  & 15.4$\pm$10.6 & 15.6$\pm$7.4 & 0.68$\pm$0.01 & 0.71$\pm$0.00  \\ \hline 
{\bf{R6}}  & 15.5$\pm$10.0 & 15.7$\pm$6.9 & 0.68$\pm$0.01 & 0.71$\pm$0.00  \\ \hline 
{\bf{R7}}  & 15.5$\pm$10.6 & 15.7$\pm$7.4 & 0.68$\pm$0.01 & 0.71$\pm$0.00  \\ \hline 
{\bf{R8}}  & 15.4$\pm$9.6 & 15.6$\pm$6.8 & 0.68$\pm$0.01 & 0.71$\pm$0.00  \\ \hline 
{\bf{R9}}  & 15.4$\pm$8.6 & 15.7$\pm$6.7 & 0.68$\pm$0.01 & 0.73$\pm$0.00  \\ \hline 
{\bf{R10}}  & 15.7$\pm$8.7 & 16.0$\pm$4.7 & 0.69$\pm$0.01 & 0.72$\pm$0.00  \\ \hline 
{\bf{R11}}  & 15.6$\pm$9.9 & 16.0$\pm$5.0 & 0.69$\pm$0.01 & 0.72$\pm$0.00  \\ \hline 
{\bf{R12}}  & 15.3$\pm$7.8 & 15.7$\pm$6.4 & 0.67$\pm$0.00 & 0.71$\pm$0.00  \\ \hline 
{\bf{R13}}  & 15.7$\pm$10.3 & 16.0$\pm$5.2 & 0.69$\pm$0.01 & 0.72$\pm$0.00  \\ \hline 
{\bf{R14}}  & 15.8$\pm$14.4 & 16.0$\pm$6.4 & 0.68$\pm$0.01 & 0.71$\pm$0.00  \\ \hline 
\end{tabular}
    }
    \caption{Perceptual Image Quality Metrics PSNR and SSIM comparison for SOTA~\cite{Chai21LatentCompositeICLR} and proposed Method on 14 different regions mentioned in the Table \ref{table:vulnerabilityMAP1}}
    \label{table:psnrSSIM}
\end{table}
\begin{figure*}[htp]
\centering
 \begin{tabular}[b]{c}
       \includegraphics[width=0.22\linewidth]{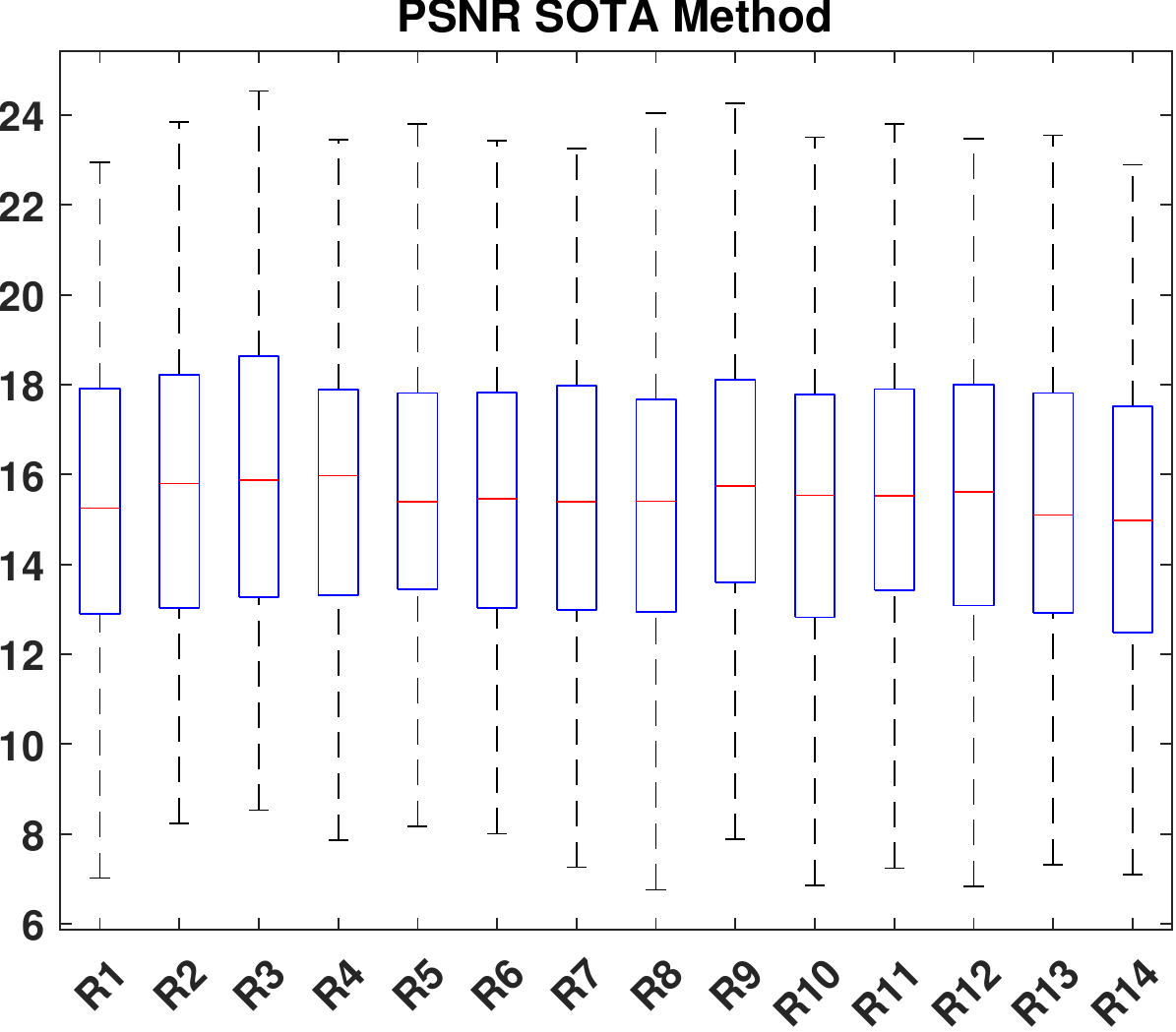}\\ 
    \small {(a) PSNR SOTA}
  \end{tabular}
  \begin{tabular}[b]{c}
       \includegraphics[width=0.22\linewidth]{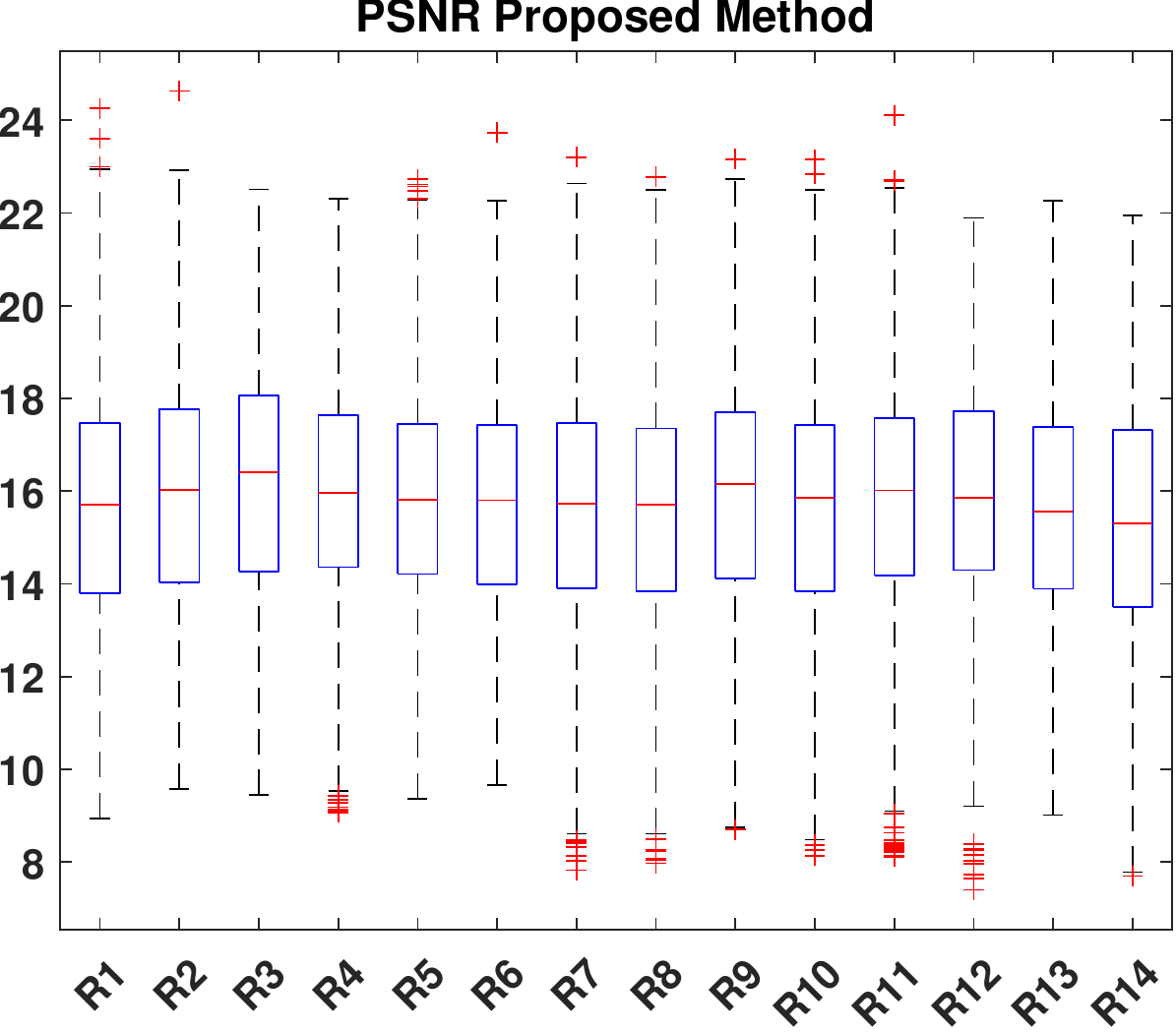}\\ 
    \small {(b) PSNR Proposed}
  \end{tabular}
\begin{tabular}[b]{c}
       \includegraphics[width=0.22\linewidth]{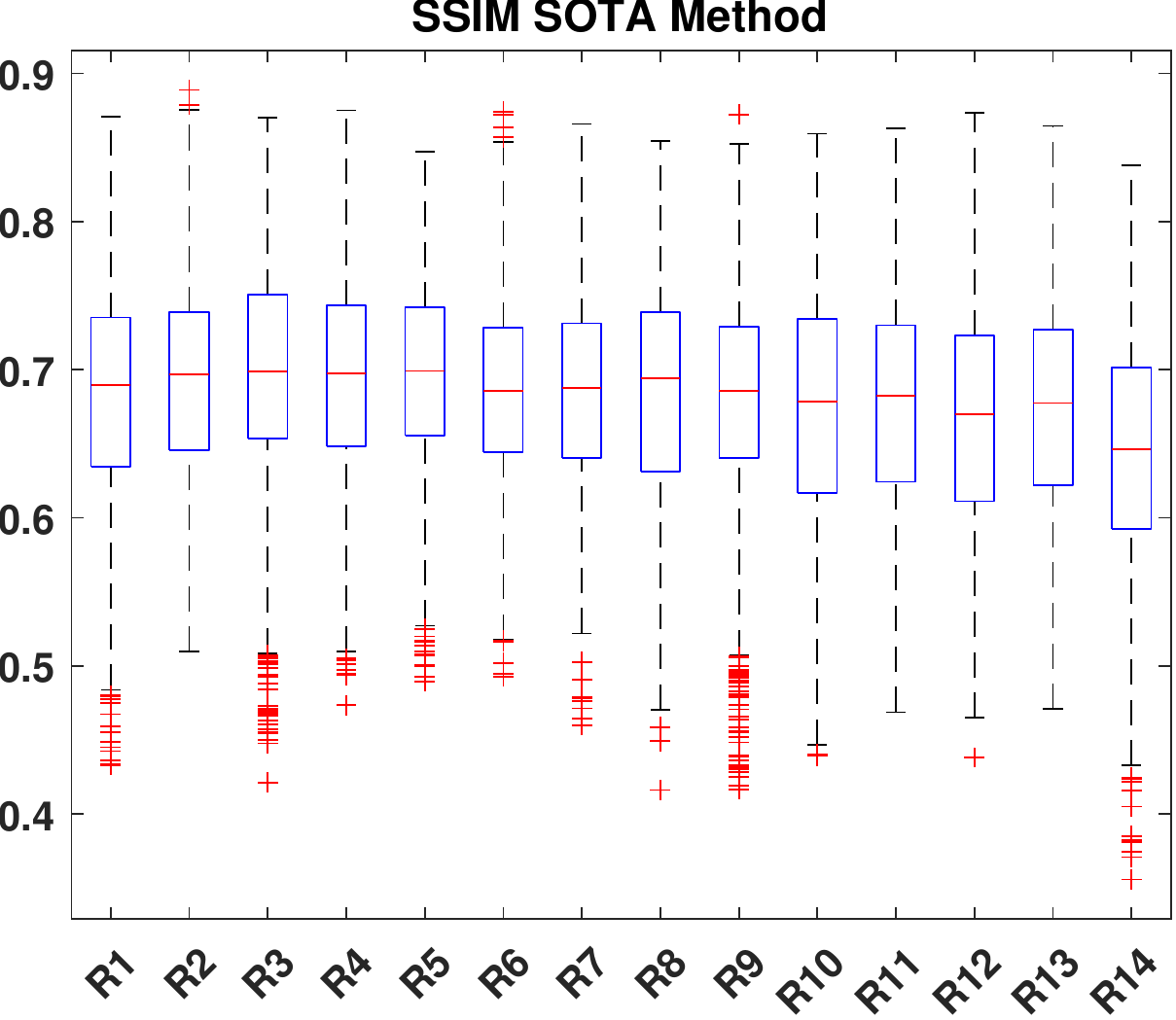}\\ 
    \small {(c) SSIM SOTA}
  \end{tabular}
\begin{tabular}[b]{c}
       \includegraphics[width=0.22\linewidth]{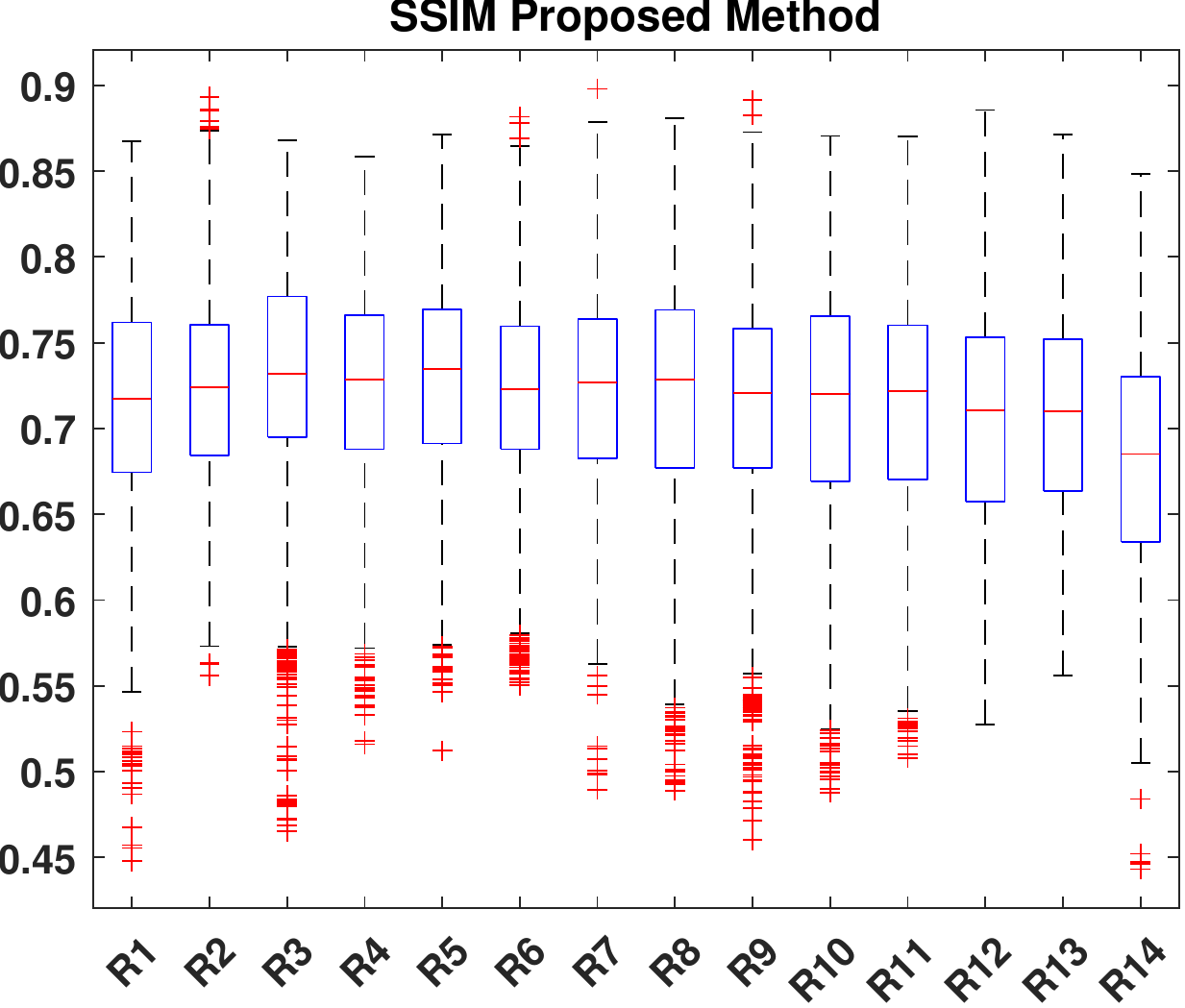}\\
    \small {(d) SSIM Proposed}
  \end{tabular}
 \caption{Box plots showing PSNR of SOTA~\cite{Chai21LatentCompositeICLR} and the proposed Method for 14 regions. These  14 regions  are same as indicated in Table~\ref{table:vulnerabilityMAP1}}
     \label{fig:PSNRPlots}
\end{figure*}
This section presents the quantitative analysis of the proposed CFIA samples using two perceptual image quality metrics, namely, PSNR (Peak Signal-to-Noise Ratio) and SSIM (Structural Similarity Index). {{ We present the results pertaining to 14 regions out of 526 unique regions for the simplicity and these regions are same as mention in Section \ref{sec:VulFRS} and in Table \ref{table:vulnerabilityMAP1}. It is worth noting that, these  14 regions will represent the lower, moderate and high vulnerability of FRS.  
}}
Both PSNR and SSIM are reference image-based quality metrics and thus require a pair of images for evaluation (face image from the contributory data subject and the generated face composite image). Table \ref{table:psnrSSIM} indicates the quantitative analysis of the perceptual quality analysis on both SOTA~\cite{Chai21LatentCompositeICLR} and the proposed CFIA method. Figure \ref {fig:PSNRPlots} illustrates the box plots corresponding to both SSIM and PSNR computed on all $14$ regions. Following are the main observations from the obtained results: 

\begin{itemize}
    \item The PSNR metric has a higher mean-value and less variance for the proposed CFIA method compared with SOTA~\cite{Chai21LatentCompositeICLR} indicating lesser noise in the face composites generated using the proposed CFIA method. This is expected as transparent blending would produce a lower contrast image, as the choice of blending-factor ($\alpha=0.5$) would generate a pixel value lower than those from contributory data subjects as the blending equation is applied twice refer Equation~\ref{eqn3}. Thus, the proposed CFIA method generates a more consistent image quality irrespective of the region compared with SOTA~\cite{Chai21LatentCompositeICLR}.
    \item The SSIM metric produces a more stable value for both the proposed CFIA method and SOTA~\cite{Chai21LatentCompositeICLR}. The proposed CFIA method gives a higher value for SSIM than the SOTA~\cite{Chai21LatentCompositeICLR}. Since SSIM is a metric more tuned to the Human Visual System (HVS), ~\cite{Hore10-PSNRSSIM-ICPR} as it measures luminance distortion, contrast distortion, and loss of correlation. Thus, our proposed CFIA method generates higher-quality composites for HVS.
\end{itemize}

\section{Human Observer Study}
\label{sec:Human}
\begin{figure*}
\begin{center}
\includegraphics[height=0.30\linewidth,trim=4 4 4 4,clip]{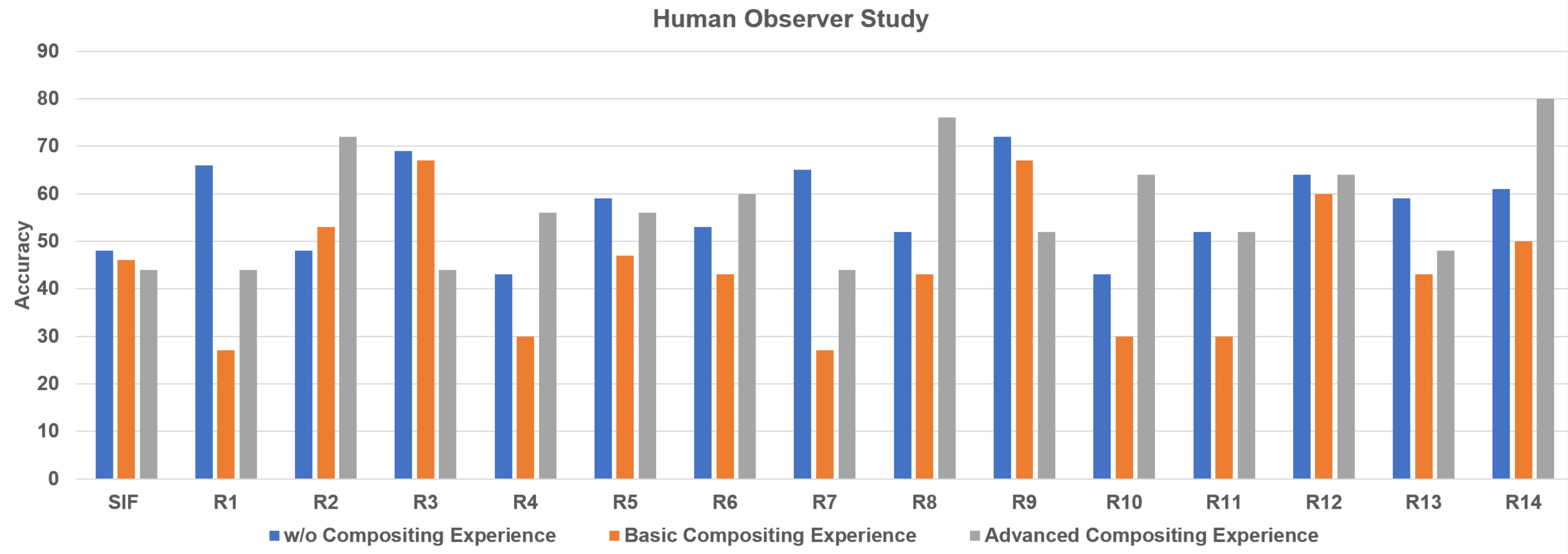}
\end{center}
   \caption{Illustration showing average accuracy quantitatively for the human observer study where bona fide or Synthetic Face Image (without any modification) is denoted as SIF.}
\label{fig:compositeSurvey}
\end{figure*}
\begin{figure}[h!]
\centering
\begin{center}
\includegraphics[width=0.9\linewidth]{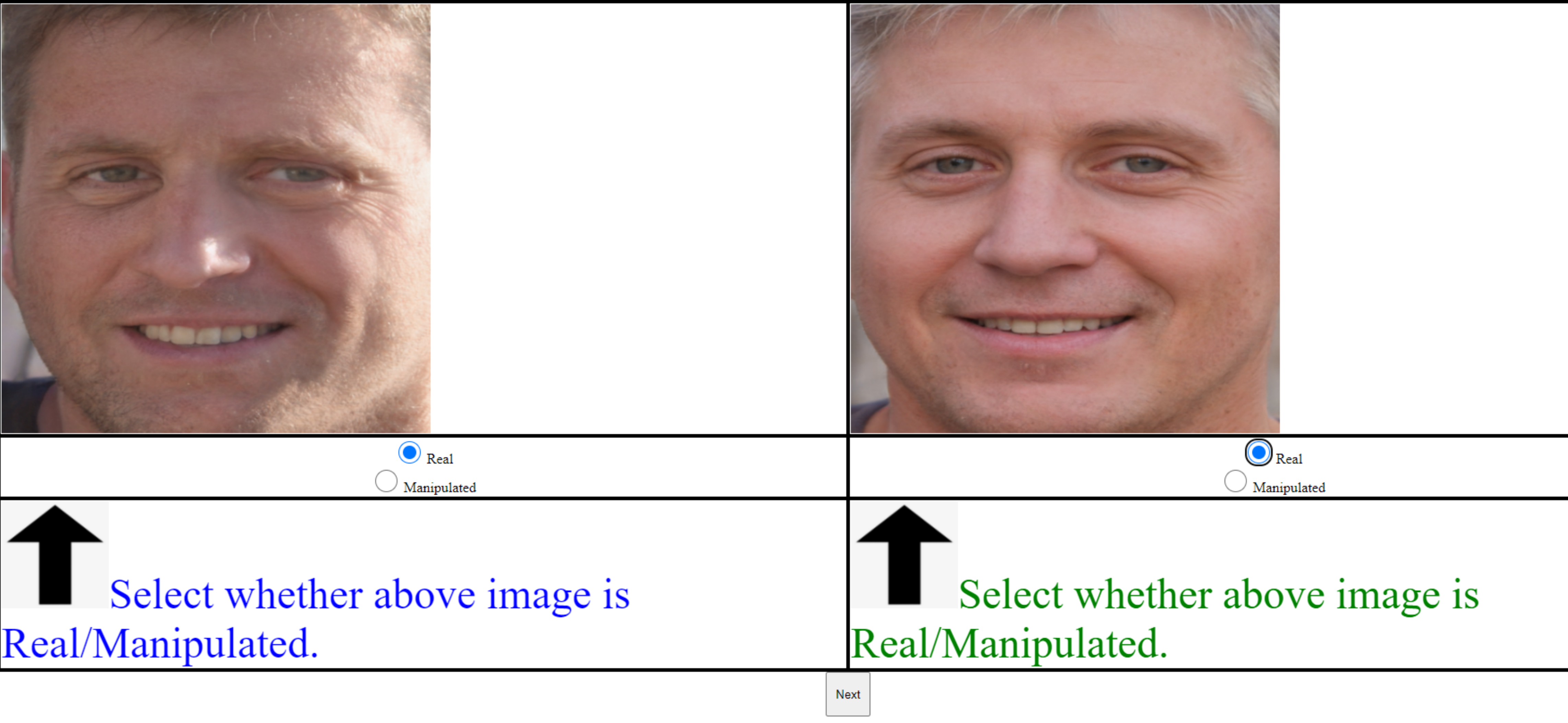}\\ 
\centering
\end{center}
\caption{Screenshot from the GUI (Full Page) of human observer web page.}
\label{fig:Screenshot}
\end{figure}

We perform a Human Observer Study (HOS) of the generated composites to evaluate the detection performance by human experts. {{We present the results pertaining to 14 regions out of 526 unique regions for the simplicity and these regions are same as mention in Section \ref{sec:VulFRS} and in Table \ref{table:vulnerabilityMAP1}. It is worth noting that, these  14 regions will represent the lower, moderate and high vulnerability of FRS.  
}}
The HOS is conducted using a web-based application  \footnote{\url{https://folk.ntnu.no/jagms/indexCompositeUpdated.html}} where a dedicated web page is set up with the use of PHP and HTML-CSS. In this study, GDPR norms are respected, and we only store the individual's email, gender, experience with the composite problem, and age group. We have made sure that the user remains anonymous during the study. 
Figure \ref{fig:Screenshot} shows the screenshot of the GUI of our website where the HOS is carried out. In this study, a human observer is shown a webpage with two images at a time where the observer has to decide independently on whether each of them is real/composite (or manipulated). The current study shows $43$ image pairs, and it takes around 20 minutes to complete the study. The study includes synthetic face images and $14$ different types of composites as mentioned in Table~\ref{table:vulnerabilityMAP1}. Further, the human observer is explained in detail the step-wise instructions to perform the study. This enables people without awareness of the image manipulation problem and those with basic and advanced awareness of the composition problem to participate in the study. In the current evaluation, $51$ human observers have participated and completed the study, including 40 participants without awareness, 6 with basic awareness, and 5 with an advanced awareness of the composition problem. 

The quantitative results of HOS are as shown in Figure~\ref{fig:compositeSurvey} and the following are the important observations: 
\begin{itemize}
\item The average detection accuracy is similar for human observers without awareness of the composition problem and those with basic awareness. This can be attributed to the innate human ability to detect composites. However, the average detection accuracy for human observers with advanced awareness of the composition problem is much higher than both without awareness and basic awareness.
\item The average accuracy is not very high for faces based on the composition, which utilizes a single facial attribute except for {\bf{R2}} with advanced awareness. This can be attributed to the fact that a large part of the facial region needs to be inpainted in the case of single facial attribute composition.
\item The average detection accuracy is  high for the regions {\bf{R8}}, {\bf{R10}}, {\bf{R12}}, and {\bf{R14}}. {\bf{R8}} has moderate parts of faces being used for compositing from the two contributory data subjects. The reason for high detection accuracy can be attributed to the fact that {\bf{R8}} has only eyes from both the contributory data subjects but his nose and mouth from different contributory data subjects. The same reasoning with more significant facial parts used for compositing can be extended to {\bf{R10}} where the nose and mouth are from different contributory data subjects but have skin and eyes from both contributory data subjects. Now for the compositing region {\bf{R12}}, the skin region is only from one contributory data subject. Thus, in all three cases, the asymmetry in the regions from the contributory data subjects aids the human observer in performing the detection at high accuracy. 
\item However, the average performance of the human observers for detecting a normal face image (or non-composite) is 46\%. Further, it is also interesting to observe that degraded performance is noted in the advanced experience group. Thus, our analysis indicates that human observers are limited in detecting the normal face images compared to the composite face images.  
\item Now, for the compositing region {\bf{R14}} all facial parts from the contributory data subjects are being used. Thus, the composited image can be distinguished from a synthetic face image using global image-based cues. 
\item In summary, we could say that either asymmetric regions or global level cues can help the human observer perform detection at high accuracy rates. However, our analysis indicates that it is very challenging for humans to detect composite attacks.    
\end{itemize}

\begin{table*}[h!]
\centering
\scriptsize{
\resizebox{0.8\linewidth}{!}{
\begin{tabular}{|c|c|c|c|c|c|c|}
\hline
{\bf{Detection Method (Region)}} &  \multicolumn{2}{|c|}{\bf{D-EER (\%)}}  & \multicolumn{4}{|c|}{\bf{BPCER @ APCER =}}   \\ \cline{4-7} 
& \multicolumn{2}{|c|}{} &  \multicolumn{2}{|c|}{\bf{5\%}} & \multicolumn{2}{|c|}{\bf{10\%}} \\ \hline
{\bf{R1}} & {\bf{SOTA~\cite{Chai21LatentCompositeICLR}}} & {\bf{Proposed}} & {\bf{SOTA~\cite{Chai21LatentCompositeICLR}}} & {\bf{Proposed}} & {\bf{SOTA~\cite{Chai21LatentCompositeICLR}}} & {\bf{Proposed}} \\ \hline DetAlgo1~\cite{Venkatesh19-MorphDeepCN-IPTA} &50.0 & 42.9 & 96.0 & 92.1 & 92.5 & 86.3 \\ \hline 
 DetAlgo2~\cite{Ramachandra19-MADScaleSpace-ISBA} &50.0 & 50.0 & 95.9 & 94.3 & 92.4 & 89.2 \\ \hline 
 DetAlgo3~\cite{Venkatesh20-MorphCAN-WACV} &38.2 & 28.7 & 85.4 & 74.0 & 78.5 & 57.2 \\ \hline 
{\bf{R2}} & {\bf{SOTA~\cite{Chai21LatentCompositeICLR}}} & {\bf{Proposed}} & {\bf{SOTA~\cite{Chai21LatentCompositeICLR}}} & {\bf{Proposed}} & {\bf{SOTA~\cite{Chai21LatentCompositeICLR}}} & {\bf{Proposed}} \\ \hline DetAlgo1~\cite{Venkatesh19-MorphDeepCN-IPTA} &50.0 & 44.6 & 96.0 & 93.0 & 92.3 & 86.0 \\ \hline 
 DetAlgo2~\cite{Ramachandra19-MADScaleSpace-ISBA} &50.0 & 50.0 & 96.2 & 94.4 & 92.3 & 91.2 \\ \hline 
 DetAlgo3~\cite{Venkatesh20-MorphCAN-WACV} &39.5 & 29.3 & 87.1 & 78.5 & 78.9 & 64.8 \\ \hline 
{\bf{R3}} & {\bf{SOTA~\cite{Chai21LatentCompositeICLR}}} & {\bf{Proposed}} & {\bf{SOTA~\cite{Chai21LatentCompositeICLR}}} & {\bf{Proposed}} & {\bf{SOTA~\cite{Chai21LatentCompositeICLR}}} & {\bf{Proposed}} \\ \hline DetAlgo1~\cite{Venkatesh19-MorphDeepCN-IPTA} &50.0 & 47.0 & 95.5 & 93.9 & 92.1 & 88.7 \\ \hline 
 DetAlgo2~\cite{Ramachandra19-MADScaleSpace-ISBA} &50.0 & 50.0 & 96.3 & 94.7 & 92.8 & 91.5 \\ \hline 
 DetAlgo3~\cite{Venkatesh20-MorphCAN-WACV} &40.6 & 32.2 & 88.3 & 79.1 & 80.3 & 65.8 \\ \hline 
{\bf{R4}} & {\bf{SOTA~\cite{Chai21LatentCompositeICLR}}} & {\bf{Proposed}} & {\bf{SOTA~\cite{Chai21LatentCompositeICLR}}} & {\bf{Proposed}} & {\bf{SOTA~\cite{Chai21LatentCompositeICLR}}} & {\bf{Proposed}} \\ \hline DetAlgo1~\cite{Venkatesh19-MorphDeepCN-IPTA} &49.0 & 39.6 & 94.3 & 90.5 & 89.0 & 81.7 \\ \hline 
 DetAlgo2~\cite{Ramachandra19-MADScaleSpace-ISBA} &50.0 & 50.0 & 96.1 & 92.6 & 92.8 & 88.9 \\ \hline 
 DetAlgo3~\cite{Venkatesh20-MorphCAN-WACV} &42.8 & 32.4 & 89.8 & 77.7 & 82.2 & 64.6 \\ \hline 
{\bf{R5}} & {\bf{SOTA~\cite{Chai21LatentCompositeICLR}}} & {\bf{Proposed}} & {\bf{SOTA~\cite{Chai21LatentCompositeICLR}}} & {\bf{Proposed}} & {\bf{SOTA~\cite{Chai21LatentCompositeICLR}}} & {\bf{Proposed}} \\ \hline DetAlgo1~\cite{Venkatesh19-MorphDeepCN-IPTA} &50.0 & 45.0 & 95.0 & 93.3 & 91.0 & 86.6 \\ \hline 
 DetAlgo2~\cite{Ramachandra19-MADScaleSpace-ISBA} &50.0 & 50.0 & 96.6 & 93.9 & 92.7 & 90.8 \\ \hline 
 DetAlgo3~\cite{Venkatesh20-MorphCAN-WACV} &42.0 & 31.6 & 87.9 & 76.5 & 80.7 & 64.8 \\ \hline 
{\bf{R6}} & {\bf{SOTA~\cite{Chai21LatentCompositeICLR}}} & {\bf{Proposed}} & {\bf{SOTA~\cite{Chai21LatentCompositeICLR}}} & {\bf{Proposed}} & {\bf{SOTA~\cite{Chai21LatentCompositeICLR}}} & {\bf{Proposed}} \\ \hline DetAlgo1~\cite{Venkatesh19-MorphDeepCN-IPTA} &50.0 & 44.1 & 95.5 & 92.5 & 91.8 & 84.6 \\ \hline 
 DetAlgo2~\cite{Ramachandra19-MADScaleSpace-ISBA} &50.0 & 50.0 & 95.3 & 92.7 & 91.0 & 88.9 \\ \hline 
 DetAlgo3~\cite{Venkatesh20-MorphCAN-WACV} &38.0 & 29.3 & 85.3 & 73.2 & 78.5 & 60.2 \\ \hline 
{\bf{R7}} & {\bf{SOTA~\cite{Chai21LatentCompositeICLR}}} & {\bf{Proposed}} & {\bf{SOTA~\cite{Chai21LatentCompositeICLR}}} & {\bf{Proposed}} & {\bf{SOTA~\cite{Chai21LatentCompositeICLR}}} & {\bf{Proposed}} \\ \hline DetAlgo1~\cite{Venkatesh19-MorphDeepCN-IPTA} &50.0 & 43.5 & 95.0 & 92.5 & 90.8 & 85.8 \\ \hline 
 DetAlgo2~\cite{Ramachandra19-MADScaleSpace-ISBA} &50.0 & 49.7 & 95.8 & 92.8 & 92.1 & 87.5 \\ \hline 
 DetAlgo3~\cite{Venkatesh20-MorphCAN-WACV} &39.5 & 29.8 & 87.9 & 73.7 & 78.9 & 60.0 \\ \hline 
{\bf{R8}} & {\bf{SOTA~\cite{Chai21LatentCompositeICLR}}} & {\bf{Proposed}} & {\bf{SOTA~\cite{Chai21LatentCompositeICLR}}} & {\bf{Proposed}} & {\bf{SOTA~\cite{Chai21LatentCompositeICLR}}} & {\bf{Proposed}} \\ \hline DetAlgo1~\cite{Venkatesh19-MorphDeepCN-IPTA} &50.0 & 44.6 & 96.2 & 94.0 & 92.2 & 85.9 \\ \hline 
 DetAlgo2~\cite{Ramachandra19-MADScaleSpace-ISBA} &50.0 & 49.8 & 96.0 & 92.5 & 92.0 & 87.4 \\ \hline 
 DetAlgo3~\cite{Venkatesh20-MorphCAN-WACV} &41.7 & 30.6 & 87.4 & 75.4 & 81.3 & 61.5 \\ \hline 
{\bf{R9}} & {\bf{SOTA~\cite{Chai21LatentCompositeICLR}}} & {\bf{Proposed}} & {\bf{SOTA~\cite{Chai21LatentCompositeICLR}}} & {\bf{Proposed}} & {\bf{SOTA~\cite{Chai21LatentCompositeICLR}}} & {\bf{Proposed}} \\ \hline DetAlgo1~\cite{Venkatesh19-MorphDeepCN-IPTA} &50.0 & 43.4 & 95.7 & 91.8 & 90.6 & 84.5 \\ \hline 
 DetAlgo2~\cite{Ramachandra19-MADScaleSpace-ISBA} &50.0 & 50.0 & 95.8 & 91.6 & 91.5 & 86.3 \\ \hline 
 DetAlgo3~\cite{Venkatesh20-MorphCAN-WACV} &40.5 & 28.4 & 87.5 & 78.5 & 81.7 & 63.2 \\ \hline 
{\bf{R10}} & {\bf{SOTA~\cite{Chai21LatentCompositeICLR}}} & {\bf{Proposed}} & {\bf{SOTA~\cite{Chai21LatentCompositeICLR}}} & {\bf{Proposed}} & {\bf{SOTA~\cite{Chai21LatentCompositeICLR}}} & {\bf{Proposed}} \\ \hline DetAlgo1~\cite{Venkatesh19-MorphDeepCN-IPTA} &48.2 & 38.5 & 94.2 & 86.6 & 87.9 & 77.9 \\ \hline 
 DetAlgo2~\cite{Ramachandra19-MADScaleSpace-ISBA} &50.0 & 48.0 & 94.7 & 91.8 & 90.7 & 87.4 \\ \hline 
 DetAlgo3~\cite{Venkatesh20-MorphCAN-WACV} &41.6 & 30.2 & 87.0 & 76.2 & 80.9 & 61.9 \\ \hline 
{\bf{R11}} & {\bf{SOTA~\cite{Chai21LatentCompositeICLR}}} & {\bf{Proposed}} & {\bf{SOTA~\cite{Chai21LatentCompositeICLR}}} & {\bf{Proposed}} & {\bf{SOTA~\cite{Chai21LatentCompositeICLR}}} & {\bf{Proposed}} \\ \hline DetAlgo1~\cite{Venkatesh19-MorphDeepCN-IPTA} &49.0 & 37.3 & 95.2 & 88.0 & 89.6 & 76.3 \\ \hline 
 DetAlgo2~\cite{Ramachandra19-MADScaleSpace-ISBA} &50.0 & 50.0 & 93.7 & 92.7 & 92.2 & 88.6 \\ \hline 
 DetAlgo3~\cite{Venkatesh20-MorphCAN-WACV} &41.9 & 31.7 & 87.1 & 80.0 & 80.7 & 67.8 \\ \hline 
{\bf{R12}} & {\bf{SOTA~\cite{Chai21LatentCompositeICLR}}} & {\bf{Proposed}} & {\bf{SOTA~\cite{Chai21LatentCompositeICLR}}} & {\bf{Proposed}} & {\bf{SOTA~\cite{Chai21LatentCompositeICLR}}} & {\bf{Proposed}} \\ \hline DetAlgo1~\cite{Venkatesh19-MorphDeepCN-IPTA} &50.0 & 43.7 & 95.4 & 90.6 & 91.4 & 83.0 \\ \hline 
 DetAlgo2~\cite{Ramachandra19-MADScaleSpace-ISBA} &50.0 & 48.8 & 94.8 & 91.4 & 91.4 & 86.0 \\ \hline 
 DetAlgo3~\cite{Venkatesh20-MorphCAN-WACV} &41.8 & 30.8 & 87.6 & 76.4 & 80.5 & 64.1 \\ \hline 
{\bf{R13}} & {\bf{SOTA~\cite{Chai21LatentCompositeICLR}}} & {\bf{Proposed}} & {\bf{SOTA~\cite{Chai21LatentCompositeICLR}}} & {\bf{Proposed}} & {\bf{SOTA~\cite{Chai21LatentCompositeICLR}}} & {\bf{Proposed}} \\ \hline DetAlgo1~\cite{Venkatesh19-MorphDeepCN-IPTA} &48.7 & 37.4 & 94.2 & 86.8 & 89.0 & 76.6 \\ \hline 
 DetAlgo2~\cite{Ramachandra19-MADScaleSpace-ISBA} &50.0 & 49.1 & 93.6 & 91.6 & 91.4 & 86.7 \\ \hline 
 DetAlgo3~\cite{Venkatesh20-MorphCAN-WACV} &41.4 & 32.2 & 86.9 & 78.5 & 80.2 & 64.8 \\ \hline 
{\bf{R14}} & {\bf{SOTA~\cite{Chai21LatentCompositeICLR}}} & {\bf{Proposed}} & {\bf{SOTA~\cite{Chai21LatentCompositeICLR}}} & {\bf{Proposed}} & {\bf{SOTA~\cite{Chai21LatentCompositeICLR}}} & {\bf{Proposed}} \\ \hline DetAlgo1~\cite{Venkatesh19-MorphDeepCN-IPTA} &50.0 & 42.6 & 97.6 & 90.2 & 94.8 & 83.7 \\ \hline 
 DetAlgo2~\cite{Ramachandra19-MADScaleSpace-ISBA} &50.0 & 49.5 & 97.2 & 95.7 & 93.4 & 88.0 \\ \hline 
 DetAlgo3~\cite{Venkatesh20-MorphCAN-WACV} &46.4 & 34.0 & 92.2 & 83.8 & 83.2 & 72.1 \\ \hline 
\end{tabular}
}}
\caption{CFIA Attack Detection  using DetAlgo1~\cite{Venkatesh19-MorphDeepCN-IPTA}, DetAlgo2~\cite{Ramachandra19-MADScaleSpace-ISBA}, and  DetAlgo3~\cite{Venkatesh20-MorphCAN-WACV}}\label{table:cadResultsBasedonMAD}
\end{table*}

\section{Composite Face Image  Attack Detection}
\label{sec:CAD}
In this section, we benchmark CFIA detection based on a single image. Since the generation of CFIA is procedurally similar to morphing generation with transparent blending. Therefore, we have employed three different Face Morphing Attack Detection (MAD) techniques to benchmark the CFIA detection. MAD methods are selected by considering their detection performance on various morphing data sources, including NIST FRVT MORPH benchmarking. To this extent, we have chosen three different S-MAD approaches, namely: Color denoising based S-MAD (DetAlgo1)~\cite{Venkatesh19-MorphDeepCN-IPTA}, Hybrid features (DetAlgo2)~\cite{Ramachandra19-MADScaleSpace-ISBA} and Residual noise-based S-MAD Network (DetAlgo3)~\cite{Venkatesh20-MorphCAN-WACV}. {{We also report the performance of CAD algorithms on 14 different regions for the same reasons that were descried in previous section \ref{sec:VulFRS}. }} These algorithms are briefly explained as follows:

\textbf{Color denoising based S-MAD (DetAlgo1)}~\cite{Venkatesh19-MorphDeepCN-IPTA}: DetAlgo1 is based on using the color information by converting the RGB image HSV color space. Then, each color channel is denoised using a Deep Convolutional Neural Network to compute the corresponding residual noise. In the next step, Pyramid LBP (P-LBP) and an SRKDA classifier for final detection.

\textbf{Hybrid features (DetAlgo2)~\cite{Ramachandra19-MADScaleSpace-ISBA}:}  DetAlgo2 is based on two different colors spaces. Given the RGB image, firstly, it is converted to HSV and YCbCr color space. In the next step, micro-texture features are computed using pyramid-LBP and passed through the SRKDA classifier. The final classification is performed using SUM rule fusion to make the final decision on detection.

\textbf{Residual noise-based S-MAD Network (DetAlgo3)~\cite{Venkatesh20-MorphCAN-WACV}:}  DetAlgo3 is based on the computing the residual noise using the Multi-Scale Context Aggregation Network (MS-CAN). The residual noise is further processed through Alexnet to obtain the classified features using the Collaborative Representative Classifier (CRC) to make the final decision to detect the attack.

{{To benchmark CFIA detection performance we resort to the off-the-shelf S-MAD.}} Three different S-MAD methods employed in this work are trained using different morph generation types (landmark-based and deep learning) and three different mediums (Digital, print-scanned, and print-scanned compression) generated using the publicly available FRGC face database. The quantitative results are presented using the ISO/IEC metrics \cite{ISO-IEC-30107-3-PAD-metrics-170227} which are as follows:  1) Attack Presentation Classification Error Rate (APCER ($\%$)) defining the percentage of attack images  (morph images) incorrectly classified as bona fide images  \cite{ISO-IEC-30107-3-PAD-metrics-170227} , 2) Bonafide Presentation Classification Error Rate (BPCER ($\%$)) defining the percentage of  bona fide images incorrectly classified as attack images \cite{ISO-IEC-30107-3-PAD-metrics-170227} and 3) Detection Equal Error Rate (D-EER ($\%$)) \cite{Zhang21-MIPGAN-TBIOM}. The detection performance is benchmarked with both SOTA and proposed CFIA images and quantitative results are presented in Table \ref{table:cadResultsBasedonMAD} and bar chart with D-EER (\%) on all 14 different regions. Based on the obtained results following are the main observations: 
\begin{itemize}
\item The CFIA detection performance is degraded with all three detection algorithms. 
\item Among three different detection algorithms. DetAlgo3 indicates the better detection accuracy attributed to the quantification of residual noise. 
\item Among the 14 different regions, the degraded detection performance is noted with the R14 on all three detection algorithms. 
\end{itemize}
Thus, based on the obtained results, we can conclude that the detection of CFIA attacks is very challenging and this needs more sophisticated detection algorithms to be devised for reliable detection. 

\section{Conclusion}
\label{sec:Conc}
{{
In this work, we presented a new type of digital attack based on the facial attributes and we termed it as Composite Face Image Attack (CFIA). Given the facial images from the two contributory data subjects, the proposed CFIA will first segment the face images into six different attributes independently. Then, these segments are blended using a transparent mask based on both single face-attribute and multiple face attributes. These attributes are processed using the image inpainting based on pre-trained GAN to generate the final CFIA samples. In this work, given the face images from two contributory data subjects, we generate 526 different composite face images based on single and multiple face attributes. We contributed a new dataset with 1000 unique identities that will result in 526000 CFIA samples. Extensive experiments are performed to evaluate the attack potential of the newly generated CFIA using four different FRS. To effectively benchmark the vulnerability of the generated CFIA, we have introduced a generalized vulnerability metric. Further, we benchmark the detection accuracy using both human and automatic detection techniques. Our results demonstrated that the proposed CFIA could indicate the vulnerability of the FRS while it is difficult to detect using both human and automatic detection techniques. In the future work, we would like to extend the present work in several directions: 1) Generation of composites of higher quality, 2) Evaluation of the proposed method on real face images on public datasets, 3) Development of novel detection techniques.}}

\bibliographystyle{ieee}
\bibliography{access}

\newpage
\appendix
\section{Role of FTAR in computing vulnerability}
In this appendix, we present additional results on the vulnerability of COTS to illustrate the importance of FTAR in computing the G-MAP. The use of academic FRS does not include quality estimation to optimize the verification performance; thus, FTAR can be assumed to be zero. However, with COTS FRS (which is more practical), the captured face quality is imposed because of which the FRS seeks good-quality face images to optimize the verification performance. The requirement of good quality will result in the rejection of probe attempts deemed low-quality face capture and, thus, the failure of verification with reasonable attempts. Hence the proposed FTAR will penalise the failure to verify with a reasonable attempt. 

Table \ref{table:vulnerability} and \ref{table:vulnerabilityMAP} indicates the quantitative results of two different Commercial-Off-The-Shelf (COTS) such as Neurotechnology Version 10.0 \cite{NeurotechVerilook} and Cognitec FaceVACS-SDK Version 9.4.2 \cite{CognitecFaceVACS} \footnote{Disclaimer: These results were produced in experiments conducted by us and should; therefore, the outcome does not necessarily constitute the best the algorithm can do.} in which G-MAP is computed with the multiple attempts on 14 different combinations. These 14 regions are the same as those used in the earlier sections of the papers that are representative of low, moderate and high vulnerability combinations. As noticed from the Tables \ref{table:vulnerability} and \ref{table:vulnerabilityMAP} the G-MAP with FTAR will indicate the less vulnerability meaning that, the COTS FRS fail to perform the verification. Therefore accountability to FTAR is important to be consider for vulnerability calculation.

\begin{table}[htp]
    \centering
      \resizebox{1.0\linewidth}{!}{
    \begin{tabular}{|c|c|c|c|c|c|c|c|c|c|c|c|c|c|c|c|} 
    \hline
\multicolumn{16}{|c|}{{\bf{ G-MAP \% (Probe Attempts) with FTAR}}} \\ \hline
{{\bf{FRS}}} & {{\bf{Method}}} & {\bf{R1}} & {\bf{R2}} & {\bf{R3}} & {\bf{R4}} & {\bf{R5}} & {\bf{R6}} & {\bf{R7}} & {\bf{R8}} & {\bf{R9}} & {\bf{R10}} & {\bf{R11}} & {\bf{R12}} & {\bf{R13}} & {\bf{R14}} \\ \hline
\multirow{2}{*}{{\bf{Neurotech (FAR=0.1\%)}}} & {\bf{SOTA~\cite{Chai21LatentCompositeICLR}}} &18.2 & 10.4 & 10.9 & 8.1 & 16.2 & 14.6 & 19.0 & 18.8 & 19.3 & 14.2 & 14.0 & 21.6 & 15.0 & 11.3  \\ \cline{2-16}
& {\bf{Proposed}} &13.8 & 10.2 & 9.7 & 17.6 & 13.5 & 14.4 & 16.0 & 17.1 & 19.3 & 22.2 & 22.9 & 21.0 & 23.7 & {\bf{23.3}}  \\ \hline 
\multirow{2}{*}{{\bf{Cognitec (FAR=0.1\%)}}} & {\bf{SOTA~\cite{Chai21LatentCompositeICLR}}} &31.6 & 22.2 & 23.3 & 19.8 & 27.7 & 28.8 & 33.1 & 34.0 & 37.9 & 30.0 & 24.8 & 41.1 & 25.1 & 21.3  \\  \cline{2-16}
 & {\bf{Proposed}} &30.5 & 22.9 & 22.3 & 43.9 & 28.1 & 26.9 & 31.9 & 34.7 & 35.3 & 54.6 & 55.1 & 43.0 & 57.6 & {\bf{60.7}}  \\ \hline 
 
\end{tabular}}
    \caption{Vulnerability analysis using the proposed GMAP metric (probe attempts-based with FTAR) for the proposed method and the SOTA~\cite{Chai21LatentCompositeICLR}}
    \label{table:vulnerability}
\end{table}
\begin{table}[htp]
\centering
 \resizebox{1.0\linewidth}{!}{
\begin{tabular}{|c|c|c|c|c|c|c|c|c|c|c|c|c|c|c|c|} 
\hline
\multicolumn{16}{|c|}{{\bf{ G-MAP \% (Probe Attempts) without FTAR}}} \\ \hline
{{\bf{FRS}}} & {{\bf{Method}}} & {\bf{R1}} & {\bf{R2}} & {\bf{R3}} & {\bf{R4}} & {\bf{R5}} & {\bf{R6}} & {\bf{R7}} & {\bf{R8}} & {\bf{R9}} & {\bf{R10}} & {\bf{R11}} & {\bf{R12}} & {\bf{R13}} & {\bf{R14}} \\ \hline
\multirow{2}{*}{{\bf{Neurotech (FAR=0.1\%)}}} & {\bf{SOTA~\cite{Chai21LatentCompositeICLR}}} &54.4 & 33.1 & 35.1 & 32.4 & 50.0 & 44.3 & 59.1 & 58.1 & 59.5 & 51.3 & 50.3 & 65.4 & 55.0 & 45.3 \\ \cline{2-16} 
& {\bf{Proposed}} &43.1 & 31.6 & 33.1 & 57.8 & 41.8 & 43.5 & 49.7 & 51.9 & 56.9 & 72.2 & 75.2 & 63.9 & 79.5 & 79.2 \\ \hline 
\multirow{2}{*}{{\bf{Cognitec (FAR=0.1\%)}}} & {\bf{SOTA~\cite{Chai21LatentCompositeICLR}}} &31.9 & 22.5 & 23.5 & 20.0 & 28.0 & 29.2 & 33.5 & 34.4 & 38.3 & 30.3 & 25.2 & 41.6 & 25.5 & 21.6 \\ \cline{2-16} 
& {\bf{Proposed}} &30.8 & 23.2 & 22.6 & 44.4 & 28.4 & 27.2 & 32.2 & 35.1 & 35.7 & 55.2 & 55.8 & 43.4 & 58.3 & 61.4 \\ \hline 
\end{tabular}}
    \caption{ Vulnerability analysis using the G-MAP metric (Probe Attempts- without FTAR) for the proposed method and the SOTA~\cite{Chai21LatentCompositeICLR}}
    \label{table:vulnerabilityMAP}
\end{table}

\begin{IEEEbiography}[{\includegraphics[width=1in,height=1.25in,clip,keepaspectratio]{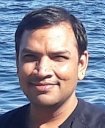}}]{Jag Mohan Singh} (Member, IEEE) received the
B.Tech. (Hons.) and M.S. by research in computer
science degrees from the International Institute
of Information Technology (IIIT), Hyderabad, in
2005 and 2008, respectively. He is currently in the
final year of his Ph.D. with the Norwegian Biometrics
Laboratory (NBL), Norwegian University
of Science and Technology (NTNU), Gjøvik. He
worked with the industrial research and development
departments of Intel, Samsung, Qualcomm,
and Applied Materials, India, from 2010 to 2018. He has published several
papers at international conferences focusing on presentation attack detection,
morphing attack detection and ray-tracing. His current research interests
include generalizing classifiers in the cross-dataset scenario and neural
rendering.
\end{IEEEbiography}

\begin{IEEEbiography}[{\includegraphics[width=1in,height=1.25in,clip,keepaspectratio]{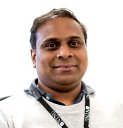}}]{Raghavendra Ramachandra} 
obtained a
Ph.D. in computer science and technology from
the University of Mysore, Mysore India and Institute
Telecom, and Telecom Sudparis, Evry, France
(carried out as collaborative work) in 2010. He
is currently a full professor at the Institute of
Information Security and Communication Technology
(IIK), Norwegian University of Science
and Technology (NTNU), Gjøvik, Norway. He is
also working as R\&D chief at MOBAI AS. He
was a researcher with the Istituto Italiano di Tecnologia, Genoa, Italy,
where he worked with video surveillance and social signal processing.
His main research interests include deep learning, machine learning, data
fusion schemes, and image/video processing, with applications to biometrics,
multi-modal biometric fusion, human behaviour analysis, and crowd
behaviour analysis. He has authored several papers and is a reviewer for
several international conferences and journals. He also holds several patents
in biometric presentation attack detection and morphing attack detection.
He has also been involved in various conference organising and program
committees and has served as an associate editor for various journals. He
has participated (as a PI, co-PI or contributor) in several EU projects, IARPA
USA and other national projects. He is serving as an editor of the ISO/IEC
24722 standards on multi-modal biometrics and an active contributor to the
ISO/IEC SC 37 standards on biometrics. He has received several best paper
awards, and he is also a senior member of IEEE.
\end{IEEEbiography}
\SetKwInput{KwInput}{Input}
\SetKwInput{KwOutput}{Output}

\def\BibTeX{{\rm B\kern-.05em{\sc i\kern-.025em b}\kern-.08em
    T\kern-.1667em\lower.7ex\hbox{E}\kern-.125emX}}
\history{Date of publication xxxx 00, 0000, date of current version xxxx 00, 0000.}
\doi{10.1109/ACCESS.2017.DOI}

\title{Supplementary Material: Deep Composite Face Image Attacks: Generation, Vulnerability and Detection}
\author{Jag Mohan Singh,~\IEEEmembership{Member,~IEEE,}
  and    Raghavendra Ramachandra,~\IEEEmembership{Senior Member,~IEEE}}
\address[]{Norwegian University of Science and Technology (NTNU), Norway\\ (e-mail: {jag.m.singh; raghavendra.ramachandra}@ntnu.no)}
\tfootnote{}

\markboth
{Singh \headeretal: Deep Composite Face Image Attacks: Generation, Vulnerability and Detection}
{Singh \headeretal: Deep Composite Face Image Attacks: Generation, Vulnerability and Detection}

\corresp{Corresponding author: Jag M. Singh (e-mail: jag.m.singh@ntnu.no).}

\titlepgskip=-15pt

\maketitle

\section{Full Composition Results for two contributory data subjects.}
In this section, we present the 526 composition images for bona fide images from Figure~\ref{fig:BonafideSubjects} in Figures~\ref{fig:BonafideSubjects1}-~\ref{fig:BonafideSubjects2},~\ref{fig:BonafideSubjects3},~\ref{fig:BonafideSubjects4},~\ref{fig:BonafideSubjects5},~\ref{fig:BonafideSubjects6} and ~\ref{fig:BonafideSubjects7}. Note the composition figures are in left to right order of the composition regions mentioned in Table 2 from the main manuscript. Further. each figure mentions the combination for  the starting and ending CFIA.
\begin{figure}
     \centering
     \includegraphics[width=0.35\textwidth]{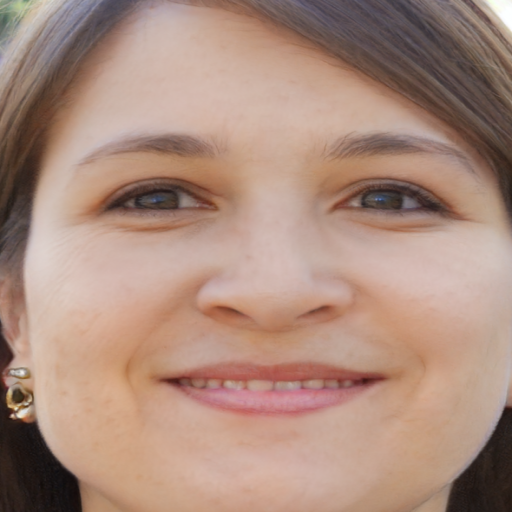}
     \hfill
     \includegraphics[width=0.35\textwidth]{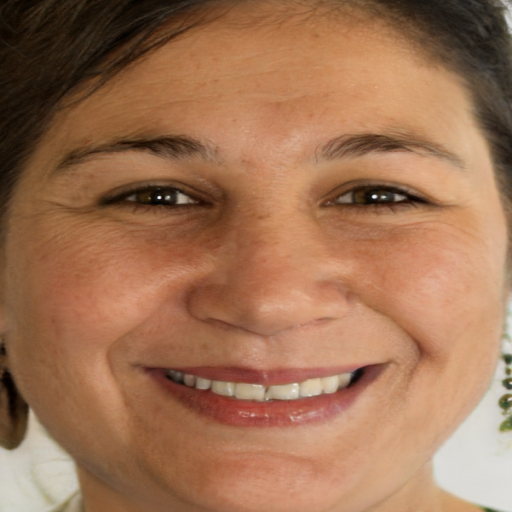}
     \caption{Bona fide subjects used for composition results as shown in Figures~\ref{fig:BonafideSubjects1}-~\ref{fig:BonafideSubjects2},~\ref{fig:BonafideSubjects3},~\ref{fig:BonafideSubjects4},~\ref{fig:BonafideSubjects5},~\ref{fig:BonafideSubjects6} and ~\ref{fig:BonafideSubjects7}}
        \label{fig:BonafideSubjects}
\end{figure}

\begin{figure*}
     \centering
    \includegraphics[width=0.85\textwidth]{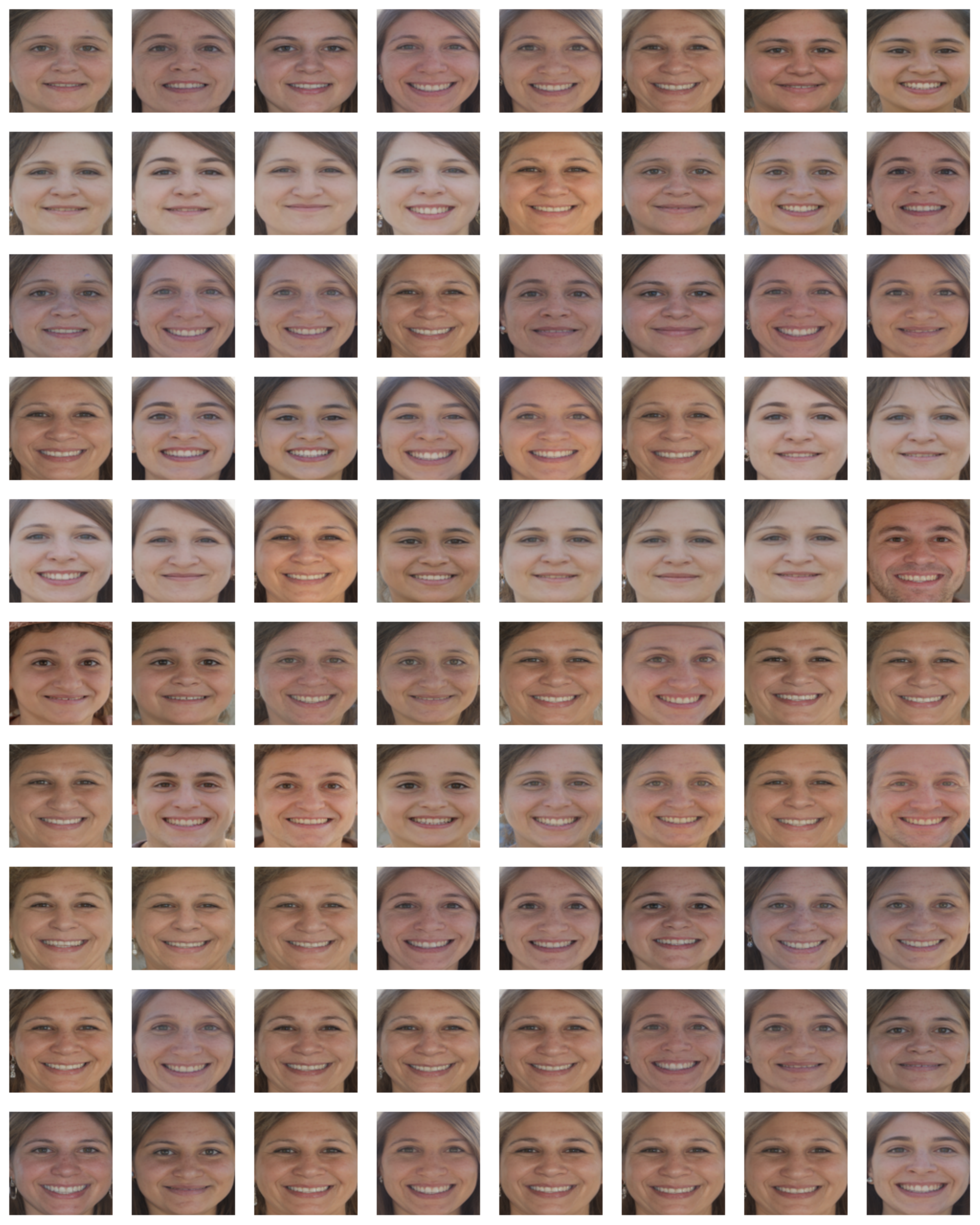}
      \caption{Bona fide subjects used for composition results where starting composition is E-H and ending composition is HN-EM}
      \label{fig:BonafideSubjects1}
\end{figure*}
\begin{figure*}
     \centering
    \includegraphics[width=0.85\textwidth]{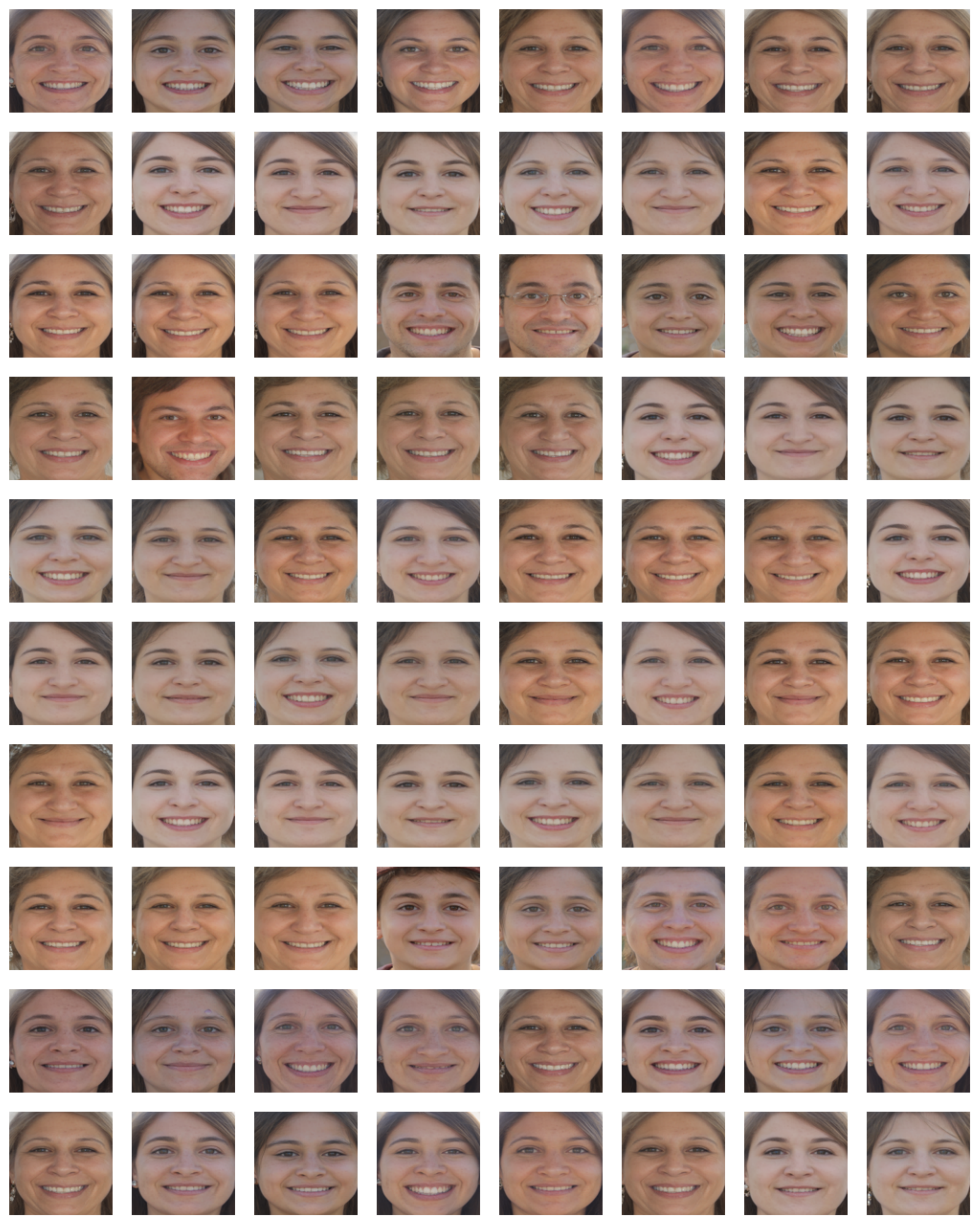}
      \caption{Bona fide subjects used for composition results where starting composition is HN-EN and ending composition is HSE-H}
      \label{fig:BonafideSubjects2}
\end{figure*}
\begin{figure*}
     \centering
    \includegraphics[width=0.85\textwidth]{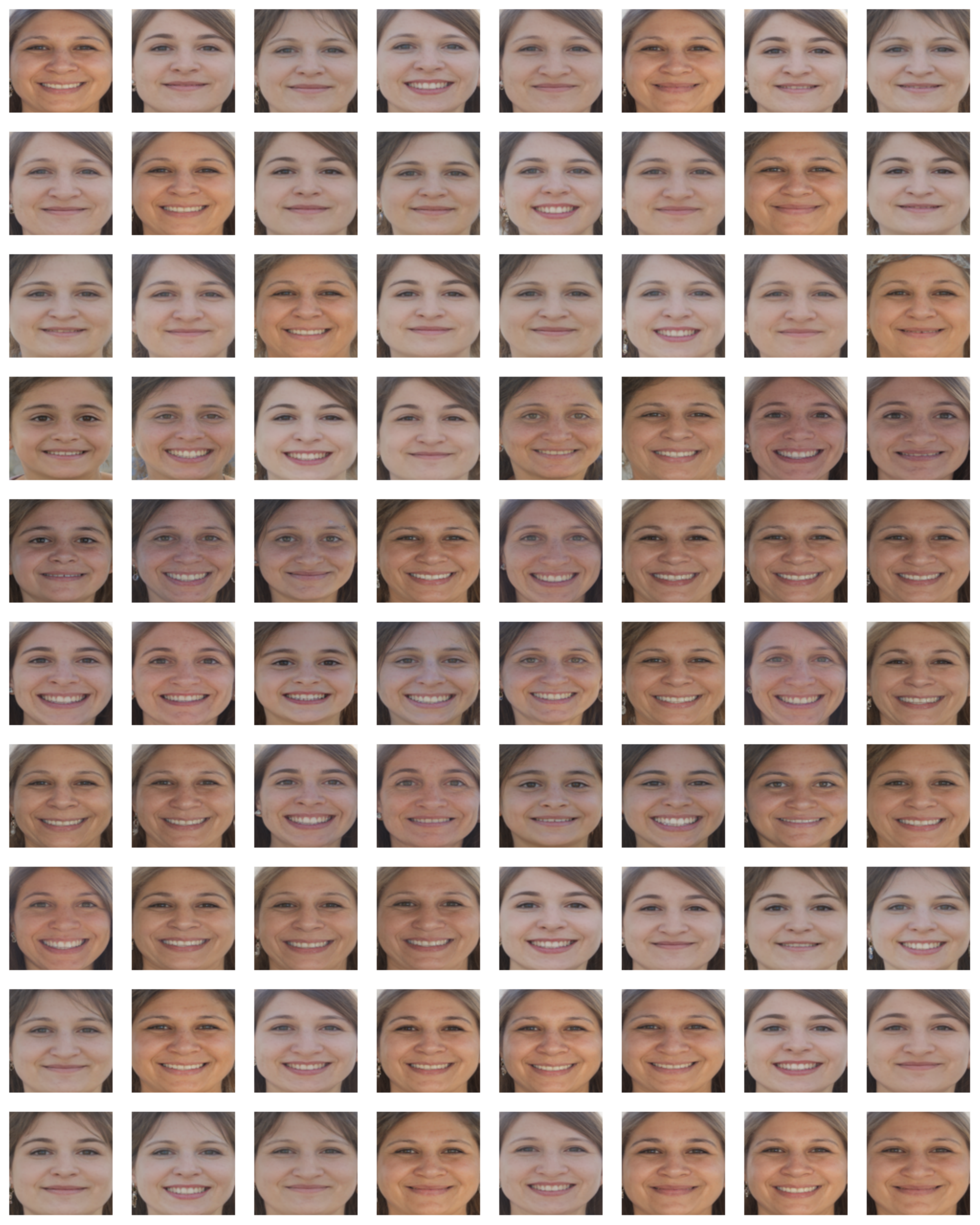}
      \caption{Bona fide subjects used for composition results where starting composition is HSE-S and ending composition is HSM-SN}
      \label{fig:BonafideSubjects3}
\end{figure*}

\begin{figure*}
     \centering
    \includegraphics[width=0.85\textwidth]{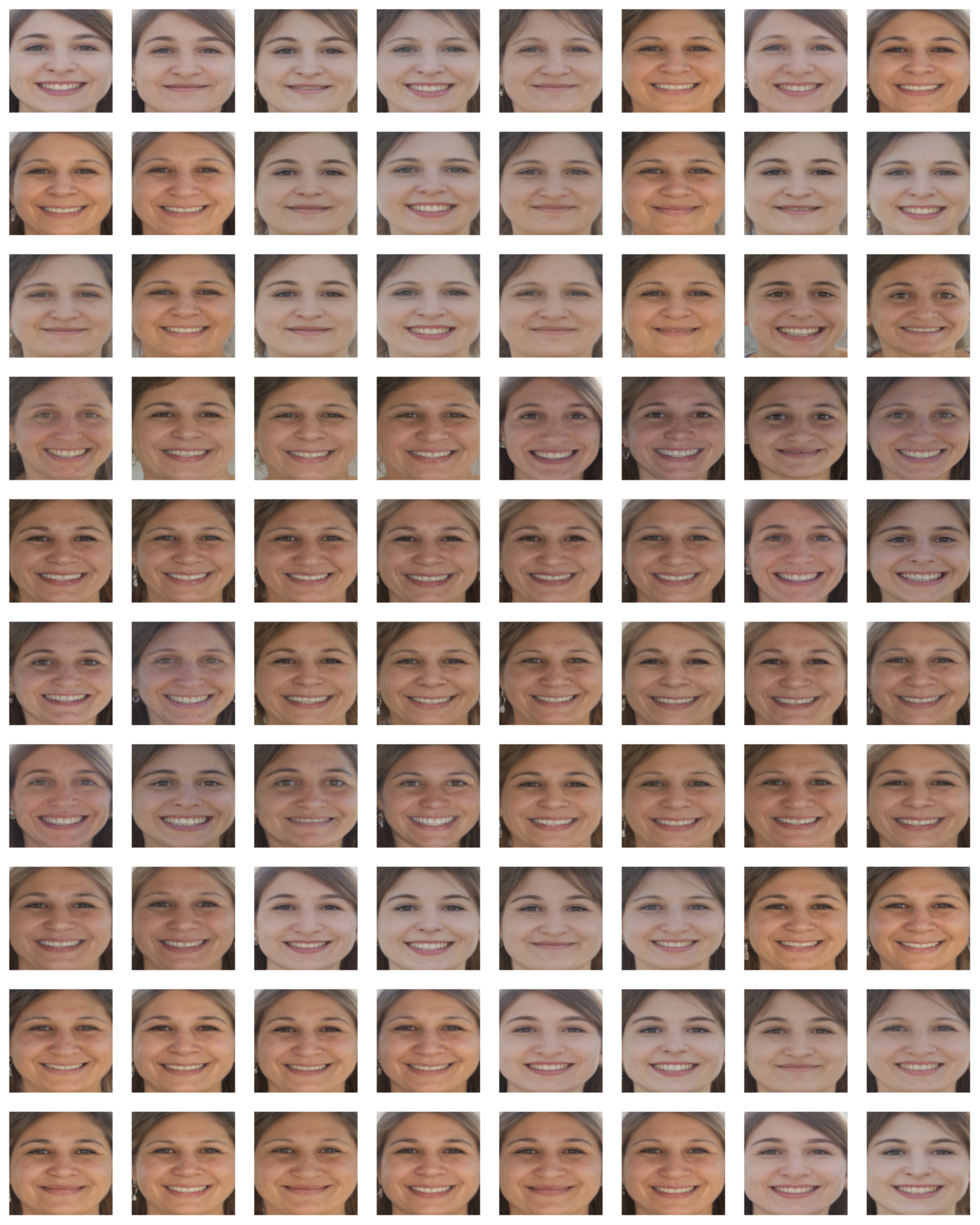}
      \caption{Bona fide subjects used for composition results where starting composition is HSN-EM and ending composition is HSN-HEM}
      \label{fig:BonafideSubjects4}
\end{figure*}

\begin{figure*}
     \centering
    \includegraphics[width=0.85\textwidth]{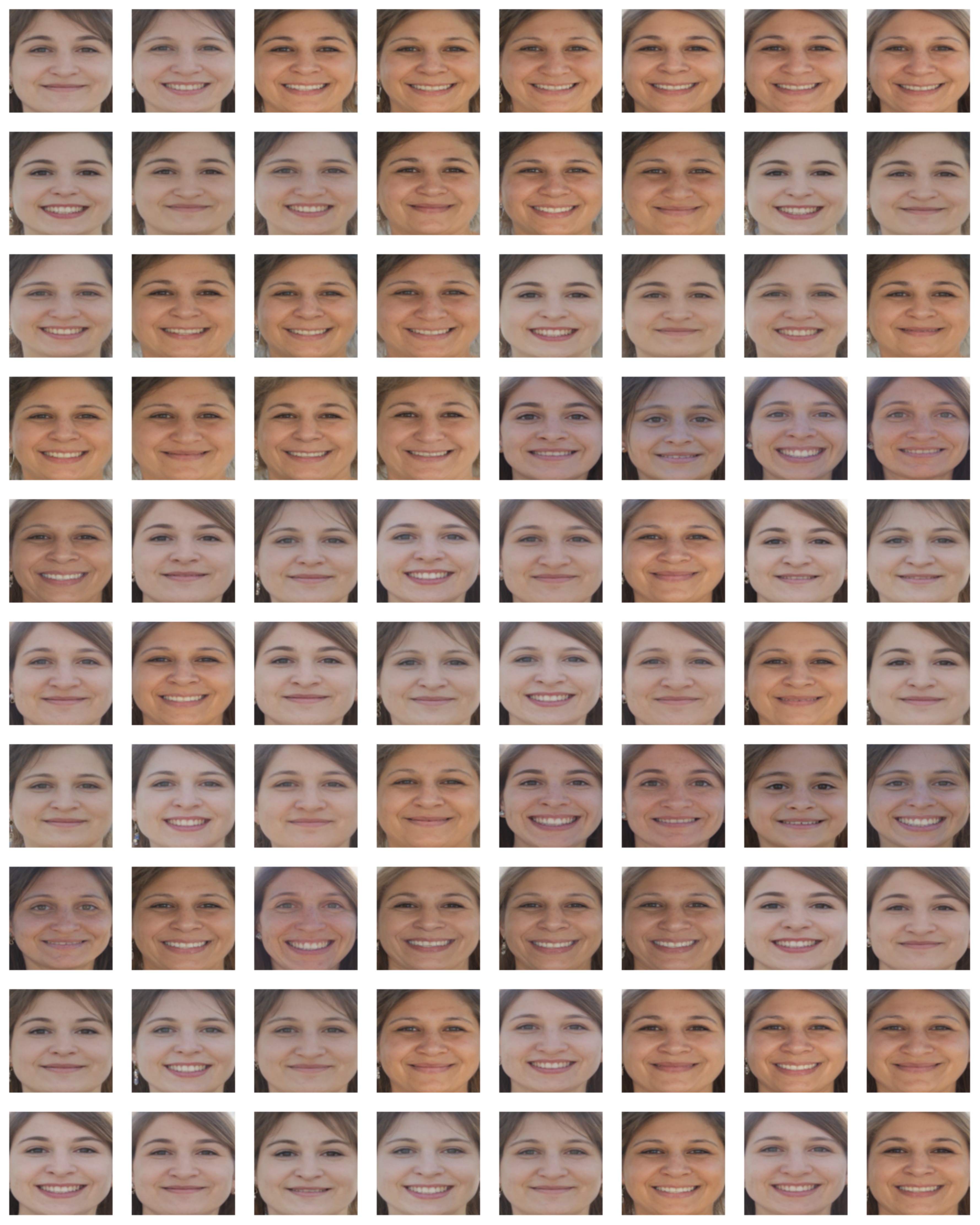}
      \caption{Bona fide subjects used for composition results where starting composition is HSN-HEN and ending composition is HSEN-SE}
      \label{fig:BonafideSubjects5}
\end{figure*}
\begin{figure*}
     \centering
    \includegraphics[width=0.85\textwidth]{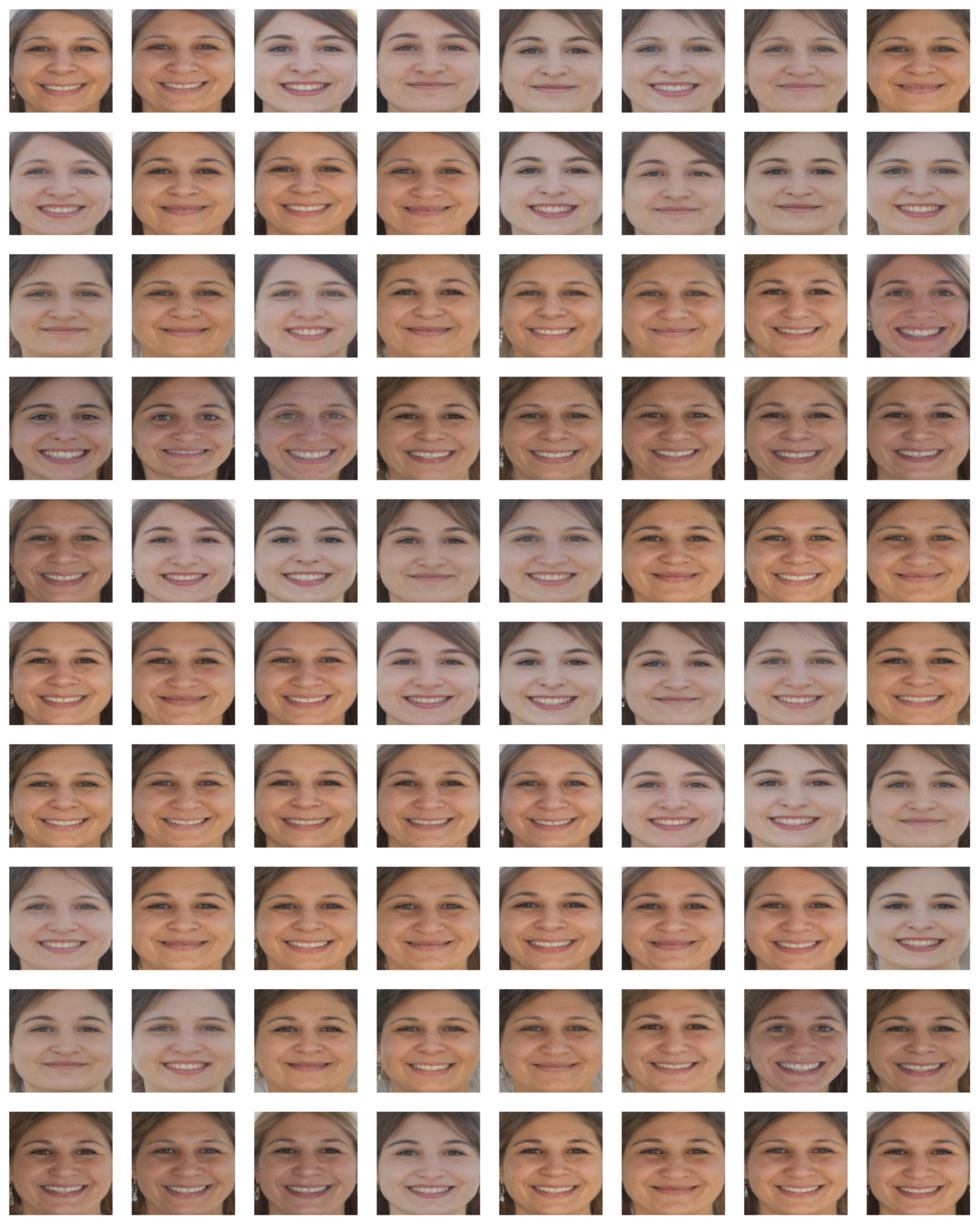}
      \caption{Bona fide subjects used for composition results where starting composition is HSEN-SM and ending composition is HSEM-SENM}
      \label{fig:BonafideSubjects6}
\end{figure*}

\begin{figure*}
     \centering
         \includegraphics[width=0.85\textwidth]{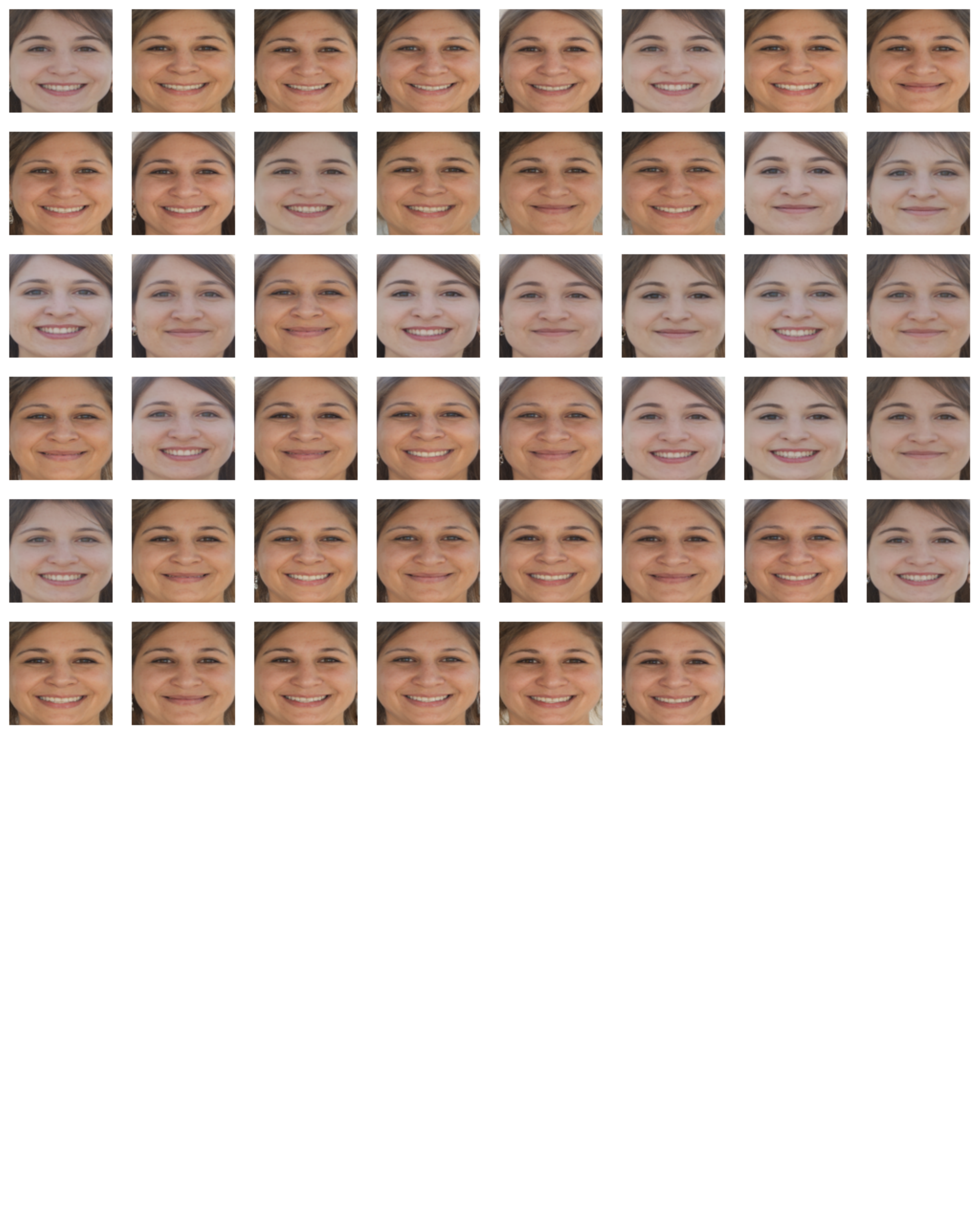}
     
       \caption{Bona fide subjects used for composition result where starting composition is HSEN-HENM and ending composition is HBSENM-HBSENM}
        \label{fig:BonafideSubjects7}
\end{figure*}


\EOD
\end{document}